\documentclass[a4paper,fleqn]{cas-dc}

\usepackage[numbers]{natbib}

\usepackage{siunitx}
\usepackage{fancyhdr}
\usepackage{pdfpages}

\def\tsc#1{\csdef{#1}{\textsc{\lowercase{#1}}\xspace}}
\tsc{WGM}
\tsc{QE}
\tsc{EP}
\tsc{PMS}
\tsc{BEC}
\tsc{DE}

\newcommand{\diff}{\mathop{}\!\mathrm{d}}

\ExplSyntaxOn
\cs_gset:Npn \__first_footerline:
{
	\group_begin:
	\small
	\sffamily
	\ifnum\theblind>0\relax
	\else 
	\__short_authors: :~
	\fi
	{ \rmfamily \itshape Composites~ Part~ A:~ Applied~ Science~ and~ Manufacturing~ (2026) }
	\group_end:
}
\ExplSyntaxOff


\begin{document}
\let\WriteBookmarks\relax
\def\floatpagepagefraction{1}
\def\textpagefraction{.001}
\shorttitle{Transient multimode heat transfer of an industrial automated tape laying process under rapidly changing conditions}
\shortauthors{B. Rameder et~al.}

\title [mode = title]{Transient multimode heat transfer of an industrial automated tape laying process under rapidly changing conditions}

\author[1]{Bernhard Rameder}[type=editor,
auid=000,bioid=1,
orcid=0000-0002-6792-6129]
\cormark[1]
\ead{bernhard.rameder@jku.at}
\ead[url]{https://www.jku.at/en/institute-of-robotics/}

\credit{Conceptualization, Methodology, Software, Validation, Formal analysis, Investigation, Data curation, Writing – original draft, Visualization, Project administration}

\affiliation[1]{organization={Institute of Robotics, Johannes Kepler University Linz},
	addressline={Altenberger Straße 69},
	postcode={4040},
	postcodesep={},
	city={Linz},
	country={Austria}}

\author[1]{Hubert Gattringer}[orcid=0000-0002-8846-9051]

\credit{Conceptualization, Resources, Writing -- review \& editing, Supervision, Project administration, Funding acquisition}

\author[1]{Andreas Müller}[orcid=0000-0001-5033-340X]

\credit{Conceptualization, Resources, Writing -- review \& editing, Supervision, Project administration, Funding acquisition}

\affiliation[2]{organization={FerRobotics Compliant Robot Technology GmbH},
	addressline={Altenberger Straße 66c}, 
	postcode={4040}, 
	postcodesep={}, 
	city={Linz},
	country={Austria}}

\author[2]{Ronald Naderer}

\credit{Resources, Writing -- review \& editing, Funding acquisition}

\cortext[cor1]{Corresponding author}

\begin{abstract}
This work presents a transient heat-transfer model of an industrial automated tape laying (ATL) process designed to overcome the limitations of conventional thermal models in composite manufacturing. The model solves the heat-conduction equation with coupled advection, conduction, convection, and radiation. A key innovation is the implementation of an analytical view factor approach that accounts for finite emitter and tape widths, thereby correcting systematic overestimations of radiative heat flux inherent in 1.5D simplifications. Furthermore, a local convection assessment incorporates mixed convection effects characterized by the Richardson number, ensuring accuracy across a wide range of process speeds. The ATL system is represented by two interacting subsystems: the moving tape substrate and the infrared heat sources. The tape is discretized using a two-node model that resolves the physical phase shift between the heated and monitored surfaces. Numerical stability under high dynamics is ensured by a monolithic solution strategy using a high-order implicit integration scheme. Model predictions were validated on an industrial ATL line, demonstrating an overall deviation of only \qty{1.08}{\percent} (NRMSE) under rapid velocity and current modulations. This framework provides a high-fidelity, physics-based foundation for thermal state estimation, supporting consistent in-situ consolidation and improved part quality.
\end{abstract}




\maketitle

\section{Introduction}

Current trends in modern manufacturing, driven by environmental protection and energy efficiency, require significant weight reduction without compromising component stability. Research indicates that the replacement of traditional steel with continuous fiber-reinforced polymer composites can enable a weight reduction of up to \qty{70}{\percent} \cite{deshpande2025comparative}. Furthermore, the aerospace, automotive and defense industries are increasingly seeking these high-performance parts to reduce fuel consumption and enhance system dynamics by lowering the overall mass. In the aerospace sector, it has been established that a \qty{1}{\percent} reduction in an aircraft's operating empty weight can translate into a reduction in fuel consumption of approximately \qty{1}{\percent} \cite{laureijs2017metal}, while in automotive applications, a \qty{10}{\percent} mass reduction can improve fuel economy by \qtyrange{6}{8}{\percent} \cite{joost2012reducing}.
To meet these demands, an increasing number of components are being manufactured from composite materials, as these offer extreme strength at a very low weight. Many of these components are manufactured using the automated tape laying (ATL) process, which builds the component up in layers of unidirectionally (UD) reinforced tapes. For these processes it is crucial to adhere to the material dependent bonding temperatures \cite{qureshi2014situ,Chen2021} and pressures \cite{chu2018placeability} at the nip point, which is the position where the actual UD tape layer is bonded onto the already deposited material on the used mold. An overview of the different process descriptions for heat transfer and intimate contact development, heat source devices and the importance of process parameters influencing the final product quality can be found in \cite{yassin2018processing}.
Measuring temperature directly at the nip point is often impractical due to space constraints. Consequently, sensors must be placed at a distance. To compensate, it is essential to use a thermodynamic model that predicts the nip point temperature based on the highly dynamic input variables, specifically the tape feed rate and the infrared heater supply current. Accurate results require the inclusion of all thermal transfer modes, namely advection, conduction, convection, and radiation.

While existing ATL models often rely on constant heat transfer coefficients, this study assesses convection locally by considering boundary layer development and the transition to mixed convection, as characterized by the Richardson number. This allows for superior accuracy across the wide range of velocities typical for industrial start-stop cycles. Considering these phenomena it is possible to accurately compute the transient temperature distribution of the tape substrate in the considered domain in presence of infrared heaters as heat source. Thereby, the heat conduction equation forms the basis of the process model. To account for the dominant spatial heat transfer caused by the moving material, an advection term is included in the equation.
Since the heat source and the temperature measurement are located on opposite sides of the tape, the heat flow through the material's thickness is explicitly integrated into the governing equations by incorporating a two-node model as described in \cite{reddy1987generalization,gilmore2002spacecraft}, based on a prescribed average temperature \cite{van2021modelling}.
Unlike conventional lumped-parameter methods, this approach accounts for thermal gradients, resolving the physical phase shift between opposing surfaces. This eliminates time delays and enables more precise temperature determination compared to empirical methods. Further improvements in model fidelity are achieved by discretizing the infrared heater model to capture dynamic thermal processes within the heater geometry, particularly during the inrush current phase. This leads to significant advancements in describing transient states. Additionally, the model was extended to accommodate the specific double-tube geometry of the used dual emitters.
By implementing sophisticated surface descriptions \cite{ehlert1993view,gross1981shapefactor}, the model corrects the systematic overestimation of radiative heat flux common in 1.5D simplifications \cite{SaRodrigues2022}. Unlike these simplified approaches, which assume infinite widths, the proposed model accounts for the finite dimensions of both the emitter and the tape. This enables the precise calculation of energy losses occurring at the edges, which is essential for a realistic irradiance mapping. Compared to computationally intensive optical ray tracing \cite{stokes2015optical}, this analytical approach provides a deterministic, noise-free irradiance distribution while reducing the numerical effort by several orders of magnitude.
A prominent example of such a laser-based ray-tracing/FEM framework was recently presented by Xu et al. \cite{xu2026laser} for CF/PEEK placement. While their formulation utilizes similar transient thermal core equations, it is fundamentally bounded to constant velocity regimes and static power settings, completely omitting the transient velocity fluctuations and time-varying power profiles characteristic of dynamic start-stop phases. In contrast, the presented framework explicitly incorporates these modulations while substituting the expensive iterative ray-tracing loops with the aforementioned analytical view-factor approach. By employing an implicit 2-stage Radau IIA scheme, this formulation allows for a robust adaptation to changing conditions without risking numerical instability. Consequently, the computational efficiency is substantially enhanced while maintaining a comparable level of predictive accuracy. The single-run simulation time drops from 20 minutes for a physical horizon of only a few seconds in the reference study to just 5.8 minutes for an extended process horizon of over 40 seconds.
This significant gain in efficiency is crucial for capturing transient process behavior and conducting extensive parameter studies within a reasonable timeframe. Furthermore, shadowing effects caused by the tape guiding plates were incorporated into the model. Collectively, these modifications enable highly accurate predictions of the transient thermal response.

The main contribution of this work lies in the synergistic combination of a transient infrared emitter model, capturing inrush current dynamics, with an advanced analytical view factor approach specifically tailored for double-tube geometries. Unlike existing literature, this model bridges the gap between high-fidelity optical ray tracing and simplified 1.5D thermal formulations. By integrating this heat source description into a computationally efficient two-node thermal laminate model, it becomes possible to predict surface-specific temperature gradients of the UD tape with high precision. Furthermore, the presented model employs a monolithic solution strategy using a high-order Radau IIA integration scheme \cite{Hairer1996}, which ensures numerical stability and eliminates time lags by solving the coupled equations simultaneously. This ensures robustness, even under the extreme parameter fluctuations typical of industrial start-stop cycles. \textcolor{blue}{Table~\protect\ref{TAB:ModelAdvancements}} summarizes these contributions and states the advancements beyond existing thermal ATL process models.
\begin{table*}[width=\textwidth,cols=4,pos=t]
	\caption{Comparison and advancements of the present model against existing ATL process models.}
	\label{TAB:ModelAdvancements}
	\begin{tabular*}{\textwidth}{@{} LLLL @{}} 
		\toprule
		\parbox[t]{2cm}{\textbf{Aspect}} &
		\parbox[t]{4.4cm}{\textbf{Existing ATL Models}} &
		\parbox[t]{4.4cm}{\textbf{Present Model (Novelty)}} &
		\parbox[t]{4.4cm}{\textbf{Advancement / Advantage}} \\
		\midrule
		\parbox[t]{2cm}{\textbf{Segmented IR Emitter Model}} & 
		\parbox[t]{4.4cm}{Spatially averaged heat source neglecting non-uniform transient filament dynamics.} & 
		\parbox[t]{4.4cm}{Discretized segment-specific emitter power evaluation capturing inrush current dynamics.} & 
		\parbox[t]{4.4cm}{Allows precise mapping and resolving of local transient temperature profiles, which directly affects radiation heat exchange.}  \\
		\midrule
		\parbox[t]{2cm}{\textbf{Finite-Geometry View Factor}} & 
		\parbox[t]{4.4cm}{Infinite parallel-plate and cylinder assumption or computationally expensive numerical integration.} & 
		\parbox[t]{4.4cm}{Analytical view factor formulation specifically tailored for finite geometries.} & 
		\parbox[t]{4.4cm}{Eliminates boundary errors and reduces computation time while maintaining high accuracy.} \\
		\midrule
		\parbox[t]{2cm}{\textbf{Two-Node Tape Model}} & 
		\parbox[t]{4.4cm}{Single isothermal node across the tape thickness (bulk temperature approximation).} & 
		\parbox[t]{4.4cm}{Computationally efficient two-node thermal laminate model resolving surface-specific temperatures.} & 
		\parbox[t]{4.4cm}{Predicts rapid, asymmetric surface-specific temperature gradients of the UD tape with high precision.} \\
		\midrule
		\parbox[t]{2cm}{\textbf{Local \& Mixed Convection Assessment}} & 
		\parbox[t]{4.4cm}{Averaged convection coefficients only for forced convection.} & 
		\parbox[t]{4.4cm}{Local evaluation and consideration of mixed convection (natural + forced).} & 
		\parbox[t]{4.4cm}{Improves local accuracy (nip- and measurement point) and captures heat loss changes even during start-stop cycles.} \\
		\midrule
		\parbox[t]{2cm}{\textbf{Monolithic Coupling}}  & 
		\parbox[t]{4.4cm}{Staggered, uncoupled, or sequentially time-lagged solvers for individual heat-transfer modes.} & 
		\parbox[t]{4.4cm}{Monolithic solution strategy using a high-order Radau IIA integration scheme.} & 
		\parbox[t]{4.4cm}{Eliminates time lags by solving coupled equations simultaneously. Ensures robustness under rapidly changing conditions.} \\
		\bottomrule
	\end{tabular*}
\end{table*}

The elaborated process model is validated on a self-developed industrial ATL line \cite{Rameder2024,rameder2025robot} by comparing the simulated temperature profile with the measured value in the defined measuring spot, where an excellent agreement with an overall deviation of only \qty{1.08}{\unit{\percent}} (NRMSE) could be observed.

While \textcolor{blue}{Section~\protect\ref{SEC:DesignATL}} addresses the design of the ATL platform to establish a foundation for accurate system modeling, \textcolor{blue}{Section~\protect\ref{SEC:ModelingFramework}} provides a detailed analysis of the governing heat transfer mechanisms for both the tape in the considered spatial domain and the infrared heaters.
Specifically, the tape model accounts for conduction through the material employing a two-node model for the tape strip, whereas a dedicated emitter model captures how inrush currents influence the dynamic behavior of the radiating infrared emitters. Both systems are coupled via an enclosure radiation model that accounts for geometric view factors. Before summarizing the research findings in \textcolor{blue}{Section~\protect\ref{SEC:Conclusion}}, the model is validated against experimental data obtained from measurements on the real system in \textcolor{blue}{Section~\protect\ref{SEC:Evaluation}}.

\section{Design and instrumentation of an experimental {ATL} platform}
\label{SEC:DesignATL}

\begin{figure}
	\centering
	\includegraphics[width=0.9\columnwidth]{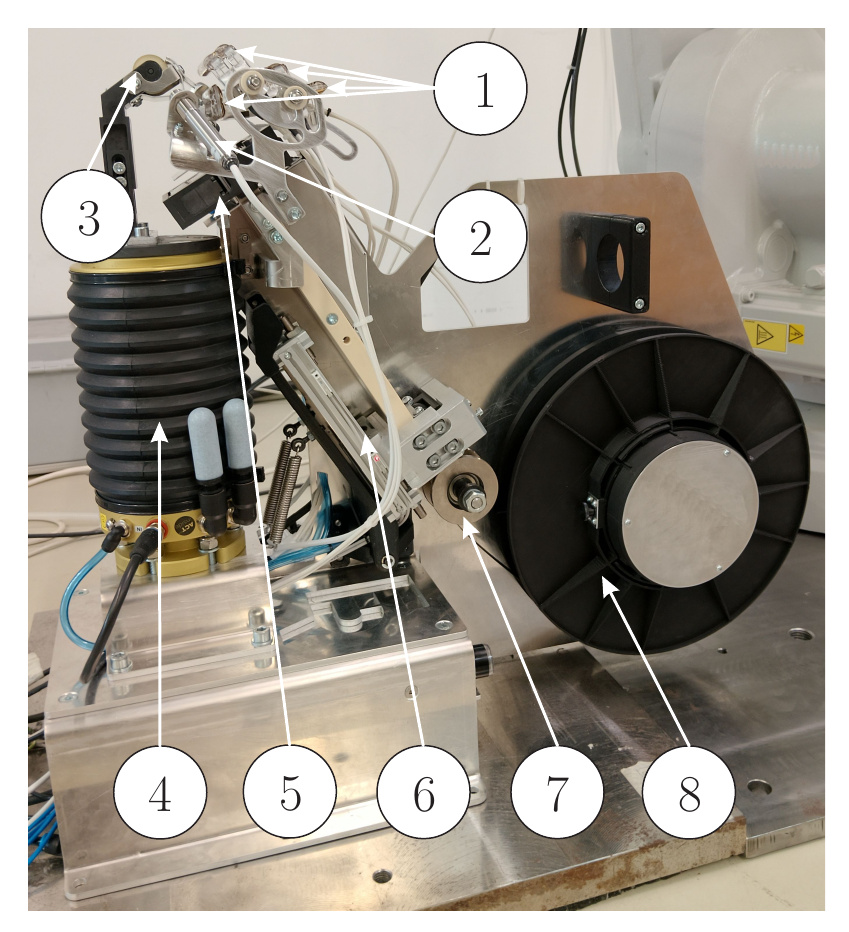}
	\caption{ATL prototype design. (1) Infrared heaters. (2) Temperature sensor. (3) Consolidation roller. (4) Active contact flange (ACF). (5) Tape cutter. (6) Tape feed. (7) Tape guide roller. (8) Tape storage spool.}
	\label{FIG:ATLSystem}
\end{figure}

\begin{figure}
	\centering
	\includegraphics[width=0.8\columnwidth]{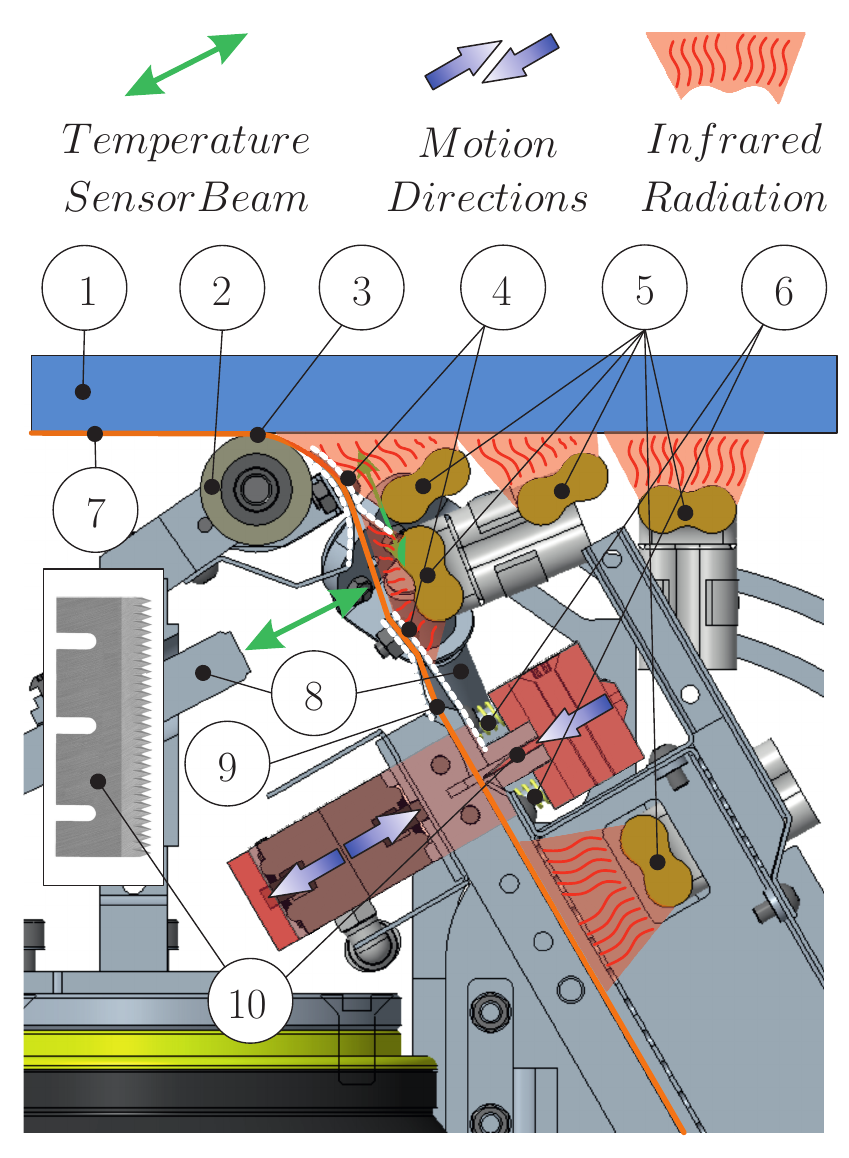}
	\caption{Heating and consolidation area of the ATL system. (1) Mold. (2) Consolidation roller. (3) Nip point. (4) Tape guide plates. (5) Infrared heaters. (6) Return springs. (7) Tape. (8) Temperature sensors. (9) Domain start. (10) Blade.}
	\label{FIG:ModelingScheme}
\end{figure}

A detailed explanation of the used setup is essential to understand the fundamental principles used for the proposed process model. Therefore, this section addresses the mechanical system setup of the self designed industrial automated tape laying device used in this study.
The prototype is shown in \textcolor{blue}{Fig.~\protect\ref{FIG:ATLSystem}}. The design is divided into modules responsible for heating, processing and feeding the tape, a consolidation unit and a tape storage spool. A detailed description of the specific design and functionality of the single subsystems is covered in \cite{Rameder2024}. 
A cross-sectional view of the device area used for modeling is shown in \textcolor{blue}{Fig.~\protect\ref{FIG:ModelingScheme}}. The considered spatial domain starts after the tape leaving the feeding section and ends in the nip point, where the tape gets tacked onto the mold.
The process model includes two Heraeus Noblelight Duo Gold infrared heaters to heat the material, with one positioned parallel to the tape strip at the beginning of the domain and the other nearly perpendicular near the end. To maintain focus on the tape's thermal history, additional setup components are excluded from the model, including the two emitters aligned with the mold used for synchronizing the substrate's temperature with the tape, as well as an optional yet inactive heater located before the tape cutting unit.
At the end of the process chain, a compaction roller mounted on a uniaxial force-controlled Active Contact Flange (ACF) by FerRobotics Compliant Robot Technology ensures precise, dynamic control of the required contact pressure at the nip point. At the begin and end of the domain, guiding sheet metal parts are installed to hold the tape in line during motion. The contactless infrared sensor Optris CS LT \cite{OptrisCSLT} captures the material temperature at a defined spot on the lower side of the tape, while a second sensor is aligned with the mold. The corresponding sensor beams are indicated by the green double-arrows in \textcolor{blue}{Fig.~\protect\ref{FIG:ModelingScheme}}.

\section{Modeling framework}
\label{SEC:ModelingFramework}

\begin{figure}
	\centering
	\includegraphics[width=1.00\columnwidth]{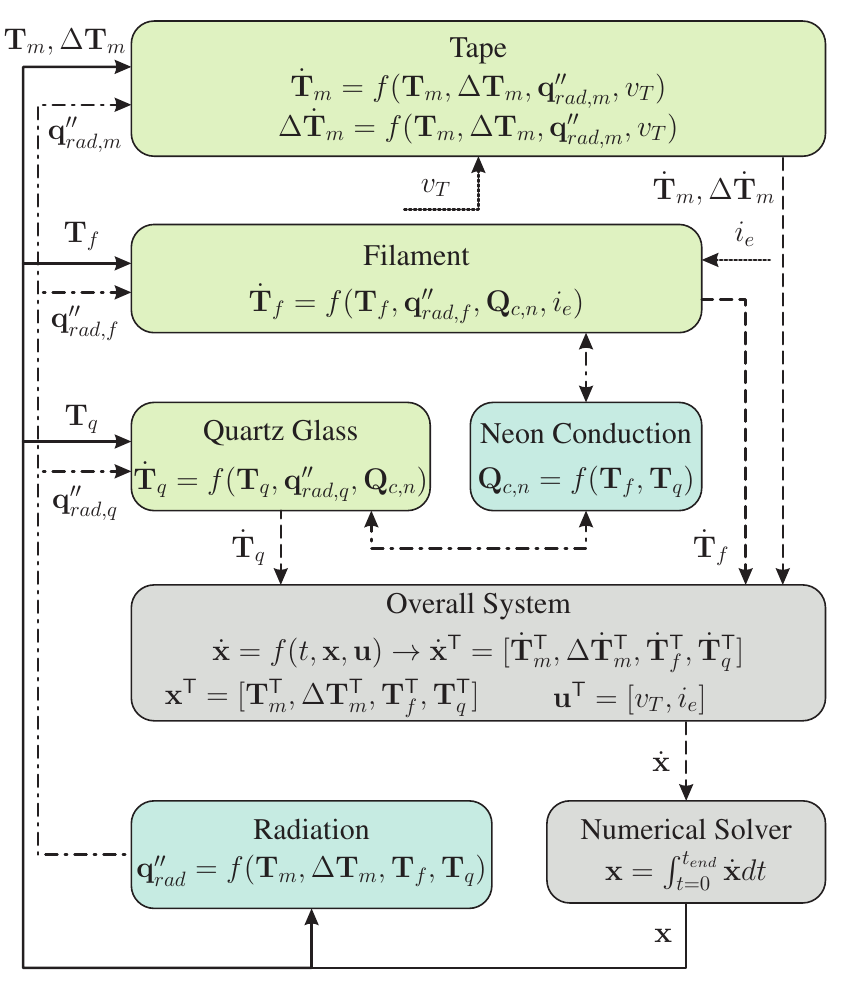}
	\caption{\textbf{Green boxes:} Governing heat transfer equations. \textbf{Blue boxes:} Interconnecting coupling heat transport phenomena. \textbf{Gray boxes:} Overall system and numerical solver.}
	\label{FIG:DependencyGraph}
\end{figure}

This section formulates the governing heat‑transfer problem on a fixed (Eulerian) spatial domain in which the moving tape traverses the heater field with velocity $v_{T}$. The tape’s transient temperature field is governed by the heat‑conduction equation, while energy exchange with the surroundings occurs at the boundaries through convection and thermal radiation. Because multiple surfaces (tape, heater tubes, guides, housing) mutually interact with each other, enclosure radiation is treated with a separate radiosity\slash view‑factor model that supplies the net radiative heat flux $\mathbf{q}_{rad}''$ to each surface patch. The heater’s electrical–thermal dynamics map the input current $i_{e}$ to filament temperature $T_{f}$, which in turn drives the radiative source seen by the tape. The tape temperature ($T_{m}, \Delta T_{m}$) is the primary state of interest, whereas the heater and radiation models are coupled submodels that provide boundary fluxes. Regarding the quartz glass envelope, its temperature $T_{q}$ depends on both neon conduction $Q_{c,n}$ and radiation. A dependency graph showing the interactions between these distinct subsystems is presented in \textcolor{blue}{Fig.~\protect\ref{FIG:DependencyGraph}}. Here, the numerical solver (\textcolor{blue}{Section~\protect\ref{SEC:NumericalImplementation}}) simultaneously solves the coupled equations, using the current temperatures to evaluate the occurring heat transfer phenomena.

\subsection{Assumptions and reference frame}
To limit the model complexity, some simplifications regarding system components and geometric details are assumed.

\subsubsection{Unidirectional carbon-fiber-reinforced tape}
\begin{figure}
	\centering
	\includegraphics[width=1\columnwidth]{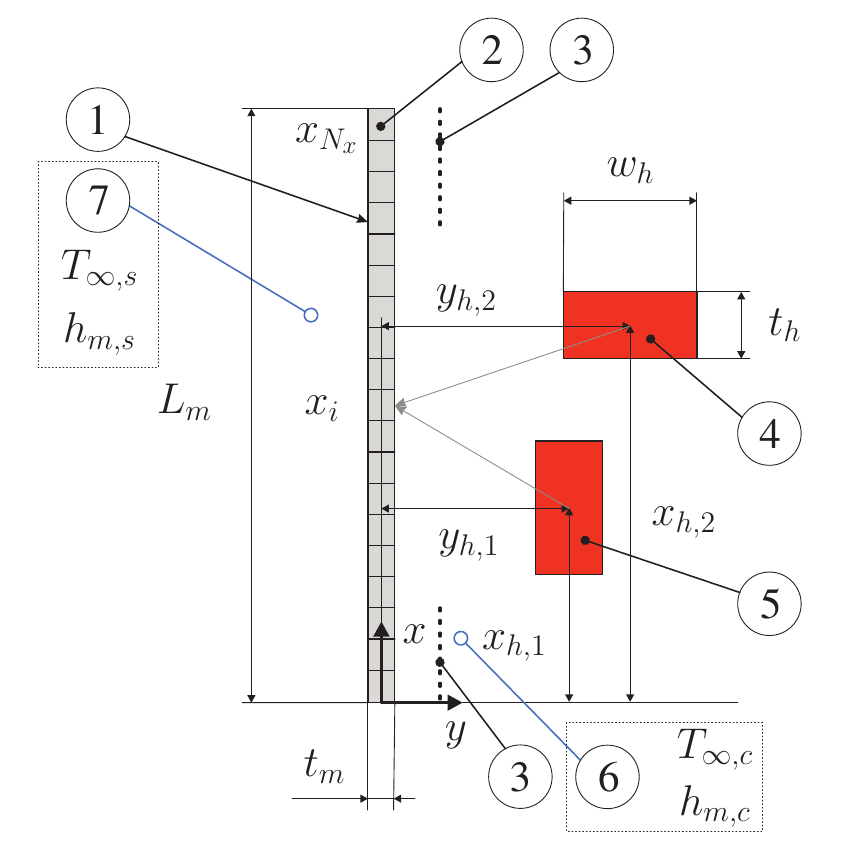}
	\caption{Process model scheme with two infrared heaters. (1) UD tape. (2) Nip point. (3) Tape guide plates (radiation shield). (4) Infrared heater perpendicular to tape. (5) Infrared heater parallel to tape. (6) Convection area inside radiation cavity. (7) Convection area surroundings.}
	\label{FIG:ProcessModelScheme}
\end{figure}

First, conduction between the tape and the guiding sheet metal parts were neglected once for the short flyby duration and on the other hand due to insufficient contact between the materials because of the constructively occurring gap distance between the metal sheets. Nevertheless, the guide plates were included in the radiation calculation as heat shield plates, which reduce the transferred radiant heat. Furthermore, the tape is assumed to move fully vertical to simplify the calculation of the different convection coefficients, although the tape in the real system has a certain feed angle and turns around the compaction roller close before the nip point. Thereby, the defined feed direction is upwards, because the system is spatially fixed mounted on the floor during the experiments, whereas the mold is handled with a robot above the device, which yields the coordinate origin depicted in \textcolor{blue}{Fig.~\protect\ref{FIG:ProcessModelScheme}}. The figure also includes some geometric parameters of the radiators like their center positions $x_{h,1}, y_{h,1}, x_{h,2}, y_{h,2}$, their height $t_{h}$ and their width $w_{h}$ as well as the tape's total length in the domain $L_{m}$ and its thickness $t_{m}$. The ambient temperatures $T_{\infty,s}$ and $T_{\infty,c}$ on the tape surface facing the surrounding or the radiation cavity respectively are later used to determine the convection heat by using the corresponding heat transfer coefficients $h_{m,s}$ and $h_{m,c}$. The tape itself is discretized along the motion direction in $N_{x}$ equally distributed parts and the nip point is located at the last segment $x_{N_{x}}$ at the end of the domain. To simplify the determination of the view factors between the heat sources and the tape segments, the first heat source ((5) in \textcolor{blue}{Fig.~\protect\ref{FIG:ProcessModelScheme}}) with respect to the tape origin is assumed to be fully parallel to the tape and the second one ((4) in \textcolor{blue}{Fig.~\protect\ref{FIG:ProcessModelScheme}}) is assumed to be perpendicular to the tape strip. For this cases the general view factor integral is solved, which is handled in more detail in following sections.

The material used is a unidirectional (UD) carbon-fiber-reinforced HDPE tape with a fiber volume fraction $\phi_{cf} = 0.33$. Owing to the aligned reinforcement, the effective thermal conductivity is orthotropic. Along the fiber direction, we adopt the Voigt model (rule of mixtures) \cite{korab2002thermal,bard2019influence}
\begin{equation}
	k_{m,\parallel} = \phi_{cf} k_{cf,\parallel} + (1 - \phi_{cf}) k_{HDPE},
\end{equation}
which provides a reliable upper bound for the thermal conductivity parallel to the fibers. This estimate utilizes the conductivities of the thermoplastic matrix $k_{HDPE}$ and the carbon fibers $k_{cf,\parallel}$ in the longitudinal direction. Transverse to the fibers, through the tape thickness $t_{m}$, a self-consistent formula presented and applied in \cite{rolfes1995transverse,bard2019influence} is used, which has shown good agreement for moderate fiber contents ($\phi_{cf} \leq 0.5$). This yields the effective thermal conductivity normal to the fiber alignment
\begin{equation}
	k_{m,\perp} = k_{HDPE} \frac{k_{cf,\perp} + k_{HDPE} + \phi_{cf} (k_{cf,\perp} - k_{HDPE})}{k_{cf,\perp} + k_{HDPE} - \phi_{cf} (k_{cf,\perp} - k_{HDPE})}
\end{equation}
as a function of the thermal conductivity of the used polymer matrix $k_{HDPE}$ and of the carbon fibers $k_{cf,\perp}$ in transversal direction, as well as the fiber volume content $\phi_{cf}$ of the used composite.

\begin{figure}
	\centering
	\includegraphics[width=1\columnwidth]{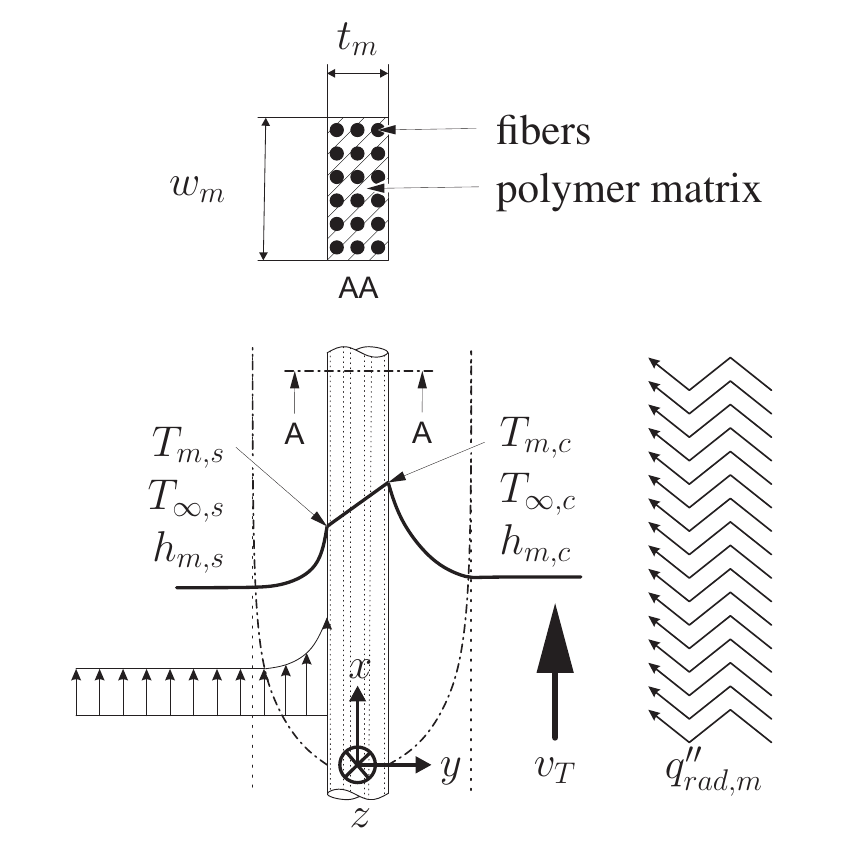}
	\caption{Tape architecture and heat transfer mechanisms.}
	\label{FIG:TapeHeatTransfer}
\end{figure}

The heat‑transfer mechanisms acting on the tape comprise convection on both exposed surfaces and thermal radiation incident on the heater‑facing surface. Convection is modeled as the superposition of natural and forced components and depends primarily on the local surface temperatures $T_{m,s}$ and $T_{m,c}$ and flow conditions. Radiative exchange between the heater assembly and the tape is computed spectrally based on Planck’s law with appropriate view factors. In practice, a gray‑body approximation with radiosity/view‑factor formulation is employed to obtain the net radiative heat flux $q_{rad}''$ on each surface element, which is counted positive when leaving the surface. Heat transport within the tape occurs by conduction across the thickness $t_{m}$ and both conduction and, predominantly, advection along the feed direction $x$ due to the tape motion with velocity $v_{T}$. Lateral temperature variations across the width $w_{m}$ are assumed small under the near‑uniform radiative exposure.
\textcolor{blue}{Fig.~\protect\ref{FIG:TapeHeatTransfer}} illustrates the dominant heat transfer mechanisms during processing, as well as it provides information about the UD architecture, highlighting the anisotropy that motivates the use of $k_{m,\parallel}$ and $k_{m,\perp}$ as defined above.

\subsubsection{Infrared heaters}
\begin{figure}
	\centering
	\includegraphics[width=.9\columnwidth]{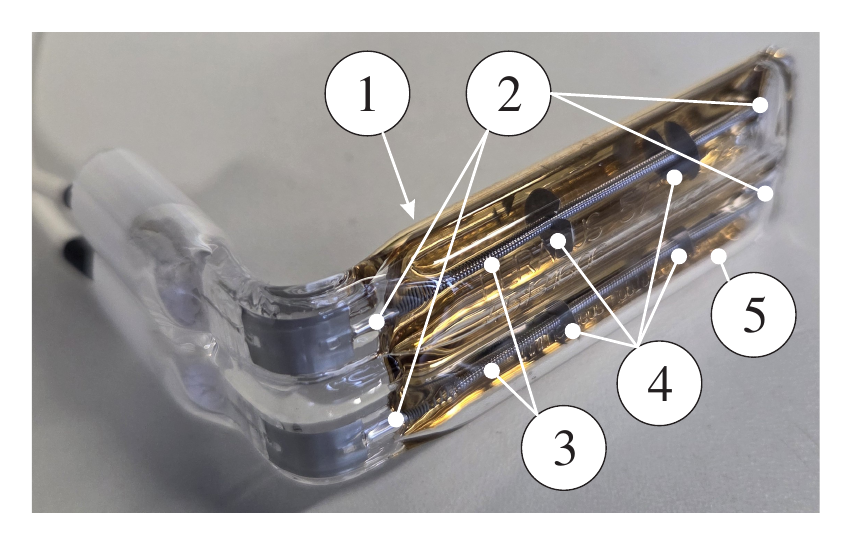}
	\caption{Heraeus Noblelight Duo Gold infrared heater. (1) Gold reflector. (2) Contact/Mounting wire. (3) Filament coil. (4) Guiding/Distance plates. (5) Quartz glass envelope.}
	\label{FIG:IRHeater}
\end{figure}

\begin{figure}
	\centering
	\includegraphics[width=1\columnwidth]{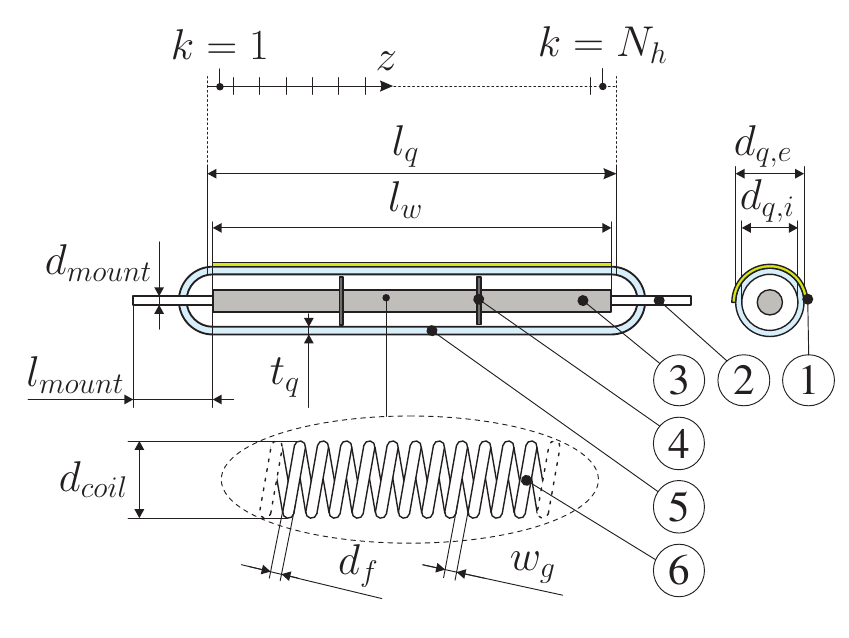}
	\caption{Infrared heater geometry considering a single tube. Discretized in $N_{h}$ segments with index $k$. (1) Gold reflector. (2) Contact/Mounting wire. (3) Filament coil. (4) Guiding/Distance plates. (5) Quartz glass envelope. (6) Filament wire in detailed view.}
	\label{FIG:IRHeaterGeometry}
\end{figure}

This subsection describes the discretized infrared (IR) heater assembly used in the ATL system. The source consists of a double‑tube geometry: two parallel quartz‑glass tubes, each housing a coiled tungsten filament, with a noble‑gas fill (assumed neon) to limit oxidation and tailor conductive-radiative heat transfer within the tube. Guiding plates center and support the filament coil to maintain a stable radiating geometry and uniform optical path to the tape. For modeling, each tube is segmented along its axial $z$-direction ($k = 1 \dots N_{h}$) to resolve spatial variations in filament temperature, radiative output, and heat losses. Radiative exchange with the tape and surrounding surfaces is computed via view factors under a gray‑body assumption, while convective losses to the ambient and conductive losses through mounts are included to capture the heater’s dynamic response to current modulation. A picture of the real double-tube heater is shown in \textcolor{blue}{Fig.~\protect\ref{FIG:IRHeater}}, whereas the simplified geometry and the discretization for one tube is depicted in \textcolor{blue}{Fig.~\protect\ref{FIG:IRHeaterGeometry}}. Here, $l_{q}$ and $l_{w}$ represent the lengths of the quartz glass tube and the coiled filament, respectively. The diameter of the filament wire is denoted by $d_{f}$, while $d_{coil}$ represents the coil diameter and $w_{g}$ the gap width between adjacent windings. The quartz glass tube has a wall thickness $t_{q}$ with internal and external diameter $d_{q,i}$ and $d_{q,e}$, respectively. The coil is attached to mounting wires with a diameter $d_{mount}$ and a length $l_{mount}$, which are hermetically sealed through the glass body.

\subsection{Governing heat-transfer equations}
The primary objective is to predict the transient temperature field of the tape as it moves through a spatially fixed yet time-varying heater field in a fixed (Eulerian) spatial domain. In this frame, the tape’s thermal history obeys the transient heat‑conduction equation with an advection term that accounts for material motion across the mesh, while heat exchange with the surroundings enters via boundary convection and thermal radiation. The governing equation for the tape temperature field $T_{m}$ is
\begin{equation}
	\rho_{m} c_{p,m} \left( \frac{\partial{T_{m}}}{\partial{t}} + \mathbf{v}_{T} \cdot \nabla T_{m} \right) = \nabla \cdot (k_{m} \nabla T_{m}) + {q}_{vol,m}''',
	\label{EQ:EnergyBalanceTape}
\end{equation}
with boundary heat-flux continuity on exposed surfaces
\begin{equation}
	-k_{m} (\nabla T_{m} \cdot \mathbf{n}) = q_{conv,m}'' + q_{rad,m}'',
	\label{EQ:BoundaryHeatFlux}
\end{equation}
where $\rho_{m}$, $c_{p,m}$ and $k_{m}$ are the density, specific heat capacity and thermal conductivity coefficient of the tape. $\mathbf{v}_{T}$ represents the tape velocity relative to the surrounding fluid at rest within the fixed domain, $q_{conv,m}''$ denotes the convective heat flux across the surface and $q_{rad,m}''$ is the net radiative flux acting on the tape supplied by the enclosure-radiation model that accounts for mutual exchange among heaters, tape, and surrounding surfaces. Vector $\mathbf{n}$ represents the surface normal of the respective segment. Internal volumetric heat generation terms ${q}_{vol,m}'''$ are later omitted, as there are no internal heat sources within the composite material. This formulation keeps the mesh stationary, isolates the advection due to motion in the PDE, and cleanly couples to the heater and radiation submodels through boundary fluxes.

To simplify the formulation, we exploit the tape’s small thickness $t_{m}$ and near‑uniform exposure across its width $w_{m}$. We assume lateral temperature variations are negligible and collapse the problem to a 1D field along the motion direction $x$ on a fixed (Eulerian) mesh. The tape cross‑section is homogenized in a first step, so boundary convection and radiation appear as an effective volumetric source
\begin{equation}
	q_{b,vol}''' = \frac{q_{conv,m}'' + q_{rad,m}''}{t_{m}}.
	\label{EQ:VolumetricBoundaryHeatFlux}
\end{equation}
The reduced model using the feed velocity $v_{T}$ in $x$-direction reads
\begin{equation}
	\frac{\partial{T_{m}}}{\partial{t}} + 
	v_{T} \frac{\partial{T_{m}}}{\partial{x}} =
	\frac{k_{m,\parallel}}{\rho_{m} c_{p,m}} \frac{\partial^2{T_{m}}}{\partial{x^2}} +
	\frac{q_{vol,m}''' - q_{b,vol}'''}{\rho_{m} c_{p,m}}.
\end{equation}
Due to the segmentation of the material in $N_{x}$ sections along its motion axis within the spatially fixed domain the spatial derivatives can be approximated using a differentiation scheme for the first and second derivative in the directions of interest, which leads to
\begin{multline}
	\frac{\partial{T_{m,i}}}{\partial{t}} = 
	-v_{T} \frac{T_{m,i} - T_{m,i-1}}{\Delta x} + \\
	\frac{k_{m,\parallel}}{\rho_{m} c_{p,m}} \frac{T_{m,i+1} - 2 T_{m,i} + T_{m,i-1}}{\Delta x^2} +
	\frac{q_{vol,m,i}''' - q_{b,vol,i}'''}{\rho_{m} c_{p,m}}.
	\label{EQ:HeatCondEquSpatialDiscrete}
\end{multline}
Index $i = 1 \hdots N_{x}$ represents the currently observed segment of the domain. Due to lack of internal heat sources, the heat production term $q_{vol,m}'''$ is set to zero in this case. However, the tapes heat flow is defined along its surfaces using the boundary heat-flux continuity defined in \textcolor{blue}{Eq.~\protect\ref{EQ:BoundaryHeatFlux}}, containing convection $q_{conv,m}''$ and radiation heat $q_{rad,m}''$ components summarized as effective volumetric heat source in \textcolor{blue}{Eq.~\protect\ref{EQ:VolumetricBoundaryHeatFlux}}.
Before defining the individual terms for the exposed surfaces, it is important to distinguish between the heated side of the tape and the side monitored by the contactless sensor. While thin tapes are typically assumed to heat up rapidly, leading to the neglect of temperature gradients across the thickness, the strongly asymmetric thermal boundary conditions, specifically the one-sided radiative heating, can still induce noteworthy temperature differences between opposite surfaces. To capture this phenomenon, the material is discretized into two layers along its thickness. This two-node approach allows for an internal conduction heat flow between the upper and lower halves of the tape, thereby accounting for the through-thickness temperature gradient that a standard lumped capacitance model would otherwise neglect.
While not within the scope of this work, compared to solving the standard heat conduction equation along axis $y$, this modeling approach offers significant advantages for future optimization cycles where computation time is critical, provided the tape is considered thermally thin. Furthermore, it simplifies the energy balancing significantly for complex movements through varying thermal zones.
To determine whether the material can be considered thermally thin \cite{Incropera2007}, the Biot number is calculated. The Biot number can be approximated using the characteristic length, which is the thickness of the tape $t_{m}$ in this case, the thermal conductivity $k_{m,\perp}$ of the composite material perpendicular to the fibers and a maximum expected convection coefficient $h_{m,max}$, which leads to a Biot number of 
\begin{equation}
	Bi = \frac{h_{m,max} t_{m}}{k_{m,\perp}} \approx 0.021
\end{equation}
when $h_{m,max} = \qty{50}{\watt\per\square\meter\per\kelvin}$ is assumed conservatively.
Although a value of $Bi \approx 0.021$ falls well below the standard threshold of \num{0.1}, which justifies a lumped capacitance formulation in the through-thickness direction ($y$) to determine the bulk temperature $T_{m}$, the severe one-sided radiative heat input violates the assumption of a completely uniform temperature field. Therefore, the formulation is extended to a coupled two-node model to accurately resolve the distinct surface temperatures at the opposite measuring and nip points. In this framework, \textcolor{blue}{Eq.~\protect\ref{EQ:HeatCondEquSpatialDiscrete}} describes the spatial and temporal evolution of the through-thickness mean temperature $T_{m}$, while the volumetric boundary heat source terms defined in \textcolor{blue}{Eq.~\protect\ref{EQ:VolumetricBoundaryHeatFlux}} utilize this thermally thin assumption to distribute the surface fluxes.
To apply this formulation, the temperatures of surfaces facing the infrared heat source inside the radiation cavity $T_{m,c}(y = t_{m}/2)$ and for surfaces facing the surroundings $T_{m,s}(y = -t_{m}/2)$ are defined. The temperature $T_{m}$ is defined as the arithmetic mean of these surface temperatures $T_{m} = (T_{m,c} + T_{m,s})/2$ and the difference reads $\Delta T_{m} = T_{m,c} - T_{m,s}$, resulting in a formulation for the surface temperatures
\begin{align}
	T_{m,c} = T_{m} + \frac{1}{2} \Delta T_{m}
	\label{EQ:TempCav} \\
	T_{m,s} = T_{m} - \frac{1}{2} \Delta T_{m}.
	\label{EQ:TempSurr}
\end{align}
This differentiation is essential for obtaining precise simulated surface temperatures at the measuring point and the nip point, which are located on opposite sides of the tape. Due to the system's design, these temperatures can vary significantly during dynamic transitions, leading to through-thickness temperature differences of up to $\Delta T_{m,max} = \qty{6}{\kelvin}$ in the evaluations presented in \textcolor{blue}{Fig.~\protect\ref{FIG:TapeTemperatures}} showing $T_{m,s}$ and $T_{m,c}$ for the measuring and nip point, respectively. $T_{m}$, on the other hand, indicates the temperature obtained considering only the lumped capacitance model. In this specific case the relative error is only approximately \qty{2}{\percent} of the corresponding mean temperature level of \qty{334}{\kelvin}. This value can increase depending on the prevailing process conditions, resulting in significant temperature differences that have to be accounted for in the computation to maintain the achieved model performance.

The conductive heat through the thickness is determined by Fourier's law $q_{cond,m}'' = k_{m,\perp}/t_{m} (T_{m,c} - T_{m,s})$ with thermal conductance per unit area $k_{m,\perp}/t_{m}$ in \unit{\watt\per\square\meter\per\kelvin}, where heat flows along the negative temperature gradient. The areal heat capacity of one of the two layers is defined as $C_{m}'' = (\rho_{m} c_{p,m} t_{m})/2$ with unit \unit{\joule\per\square\meter\per\kelvin}. Using these parameters, the energy balances for the two layers including advection can be established as
\begin{equation}
	C_{m}'' (\dot{T}_{m,c} + v_{T}\frac{\partial T_{m,c}}{\partial x}) = -q_{conv,c}'' - q_{rad,m}'' - \frac{k_{m,\perp}}{t_{m}} \Delta T_{m}
	\label{EQ:CondCav}
\end{equation}	
and
\begin{equation}
	C_{m}'' (\dot{T}_{m,s} + v_{T}\frac{\partial T_{m,s}}{\partial x}) = -q_{conv,s}'' + \frac{k_{m,\perp}}{t_{m}} \Delta T_{m},
	\label{EQ:CondSurr}
\end{equation}
where $q_{conv,c}'' = h_{m,c} (T_{m,c} - T_{\infty,c})$ and $q_{conv,s}'' = h_{m,s} (T_{m,s} - T_{\infty,s})$
are the convection heat terms of the surfaces facing the radiation cavity or the surroundings, respectively.
After subtracting \textcolor{blue}{Eq.~\protect\ref{EQ:CondSurr}} from \textcolor{blue}{Eq.~\protect\ref{EQ:CondCav}} and substituting \textcolor{blue}{Eq.~\protect\ref{EQ:TempCav}} and \textcolor{blue}{Eq.~\protect\ref{EQ:TempSurr}}, we obtain an ODE for the temperature difference between the surfaces facing the cavity and the surroundings
\begin{multline}
	\Delta \dot{T}_{m} = -v_{T} \frac{\partial \Delta T_{m}}{\partial x} + \frac{1}{C_{m}''} \Bigl[ - q_{rad,m}'' - h_{m,c} (T_{m} - T_{\infty,c}) + \\
	h_{m,s} (T_{m} - T_{\infty,s}) - \left(2 \frac{k_{m,\perp}}{t_{m}} + \frac{h_{m,c} + h_{m,s}}{2}\right) \Delta T_{m} \Bigr],
	\label{EQ:DeltaTm_ODE}
\end{multline}
which has to be solved to be finally able to determine the surface temperatures of interest in \textcolor{blue}{Eq.~\protect\ref{EQ:TempCav}} and \textcolor{blue}{Eq.~\protect\ref{EQ:TempSurr}}.
After spatial discretization, the equation for the temperature difference between the surfaces, analog to \textcolor{blue}{Eq.~\protect\ref{EQ:HeatCondEquSpatialDiscrete}}, is
\begin{multline}
	\Delta \dot{T}_{m,i} = -v_{T} \frac{\Delta T_{m,i} - \Delta T_{m,i-1}}{\Delta x} +
	\frac{1}{C_{m}''} \Bigl[ -q_{rad,m,i}'' - \\ 
	h_{m,c,i} (T_{m,i} - T_{\infty,c}) + h_{m,s,i} (T_{m,i} - T_{\infty,s}) - \\
	\left(2 \frac{k_{m,\perp}}{t_{m}} + \frac{h_{m,c,i} + h_{m,s,i}}{2}\right) \Delta T_{m,i} \Bigr].
	\label{EQ:DeltaTm_ODE_SpatialDiscrete}
\end{multline}

In order to calculate the convective heat transfer at the tape surfaces, it is necessary to define the heat transfer coefficients $h_{m,c}$ and $h_{m,s}$. Therefore, the Reynolds number $Re_{x}$ and Prandtl number $Pr$ have to be computed for evaluating the forced convection, whereas the Grashof number $Gr_{x}$ and the Rayleigh number $Ra_{x}$ are needed for natural convection. Additionally the Richardson number $Ri_{x}$ indicates, if forced $(Ri_{x} \ll 1)$ or natural $(Ri_{x} \gg 1)$ convection dominates, or a combined approach $(Ri_{x} \approx 1)$ for computing a combination of the different convection terms is more effective. All dimensionless numbers used to define the flow regime including the corresponding approximations for the occurring material properties are described in detail in \textcolor{blue}{Section~\protect\ref{SEC:S_DimensionlessNum}} of the supplementary material. When evaluating the Richardson number by using the expected maximum parameters listed in the supplementary \textcolor{blue}{Table~\protect\ref{TAB:S_MaxExpectedConvectionProperties}}, the flow regime is identified as combined natural and forced convection due to $Ri_{L_{m}}(L_{m},T_{s,max},v_{T,max}) \approx \qty{6.23}{}$.

With respect to our assumption of a vertically movement of the considered tape strip, the Nusselt number for forced convection of a vertically aligned thin plate in still air with surrounding laminar flow is described by Sakiadis \cite{sakiadis1961boundary,Sakiadis1961II,tsou1967flow} as
\begin{equation}
	Nu_{x,forced} \approx 0.447 Re_{x}^{1/2} Pr^{1/3}.
	\label{EQ:NuForcedConvection}
\end{equation}
The natural convection from a vertical plate follows the Churchill-Chu correlation \cite{churchill1975correlating}
\begin{equation}
	Nu_{x,natural} = 0.68 \frac{0.670 Ra^{1/4}}{\left[ 1 + (0.437/Pr)^{9/16} \right]^{4/9}}.
	\label{EQ:NuNaturalConvection}
\end{equation}
Considering the worst case scenario defined in \textcolor{blue}{Table~\protect\ref{TAB:S_MaxExpectedConvectionProperties}}, \textcolor{blue}{Eq.~\protect\ref{EQ:NuForcedConvection}} fulfills $Re_{L_{m}} \approx \num{658} \lesssim \num{5e5}$, whereas \textcolor{blue}{Eq.~\protect\ref{EQ:NuNaturalConvection}} satisfies $Ra_{L_{m}} \approx \num{2.19e6} \lesssim \num{e9}$ to remain laminar. The resulting mixed Nusselt number reads
\begin{equation}
	Nu_{x,mixed} = \left( Nu_{x,forced}^{n}(x) + Nu_{x,natural}^{n}(x) \right)^{1/n},
\end{equation}
where $n \approx 3$ is a good choice for most practical systems. Finally, the local convective heat transfer coefficient is defined as
\begin{equation}
h_{m}(x) = \frac{Nu_{x,mixed}(x) k_{air}(x)}{x}
\end{equation}
and is used for determining the convective heat losses $q_{conv,(c,s)}''(x) = h_{m,(c,s)}(x) (T_{m,(c,s)}(x) - T_{\infty,(c,s)})$ along the corresponding surfaces with position $x$ in the preceding equations. As a substitute for the generalized convection losses $q_{conv,m}''$ in equation \textcolor{blue}{Eq.~\protect\ref{EQ:VolumetricBoundaryHeatFlux}}, the sum of the convection terms $q_{conv,c}'' + q_{conv,s}''$ from both sides of the tape is used to calculate the ODE of the mean temperature $T_{m}$ in \textcolor{blue}{Eq.~\protect\ref{EQ:HeatCondEquSpatialDiscrete}}. A comprehensive sensitivity analysis comparing the fully coupled local mixed convection model against a simplified constant forced-convection framework is provided in \textcolor{blue}{Section~\protect\ref{SEC:S_FlowRegime}} of the supplementary material.

\subsection{Infrared heater source model}
To be able to compute the net radiation heat in the next section, which emanates from the used infrared emitters and irradiates the tape, this section gives a detailed insight into the electrical-thermal dynamics describing the influence of the input current $i_{e}$ on the filament temperature $T_{f}$ and how neon heat conduction, radiation, and convection affect the thermal behavior of the radiators. To accomplish this, the energy balances of the tungsten filament and the quartz glass body are established similarly yet with some modifications as it is done in \cite{SaRodrigues2022}.

\subsubsection{Tungsten filament}
In contrast to the consideration of a homogeneous structure as described in \cite{SaRodrigues2022}, here the entire radiator is discretized in $N_{h}$ segments along its longitudinal axis in order to be able to capture additional phenomena such as the rapid heating of certain areas of the heating coil when the inrush current occurs. This effect would be lost if the whole coil is seen as homogeneous part, because the electrical power due to the flow of the current through the electrical resistance has to heat the whole mass of the material at once, which slows down the actual dynamic of the infrared heaters. In contrast, segmentation causes individual areas to become slightly hotter, which in turn increases the electrical resistance due to the positive temperature coefficient of tungsten and generates even more energy at the corresponding point until an equilibrium is reached between heat dissipation and energy input. Thereafter, internal heat conduction combined with radiative exchange between adjacent windings lead to a uniform temperature distribution throughout the wire. This is what makes it possible to capture the high dynamics of the heating coil, which would otherwise be lost, if the coil is treated as a single material block.

The local conservation of energy for a spatial control volume $V_{k}^{f}$ with a closed boundary surface $A_{k}^{f}$ is governed by the following integral relation
\begin{equation}
	\int_{V_{k}^{f}} \rho_{f} c_{p,f} \frac{\partial T_{f}}{\partial t} \, dV = \int_{V_{k}^{f}} q_{vol,f}''' \, dV - \int_{A_{k}^{f}} \mathbf{q}'' \cdot \mathbf{n} \, dA,
\end{equation}
where $q_{vol,f}'''$ denotes the volumetric internal heat generation rate in \unit{\watt\per\cubic\meter}, $\mathbf{q}''$ represents the heat flux vector due to conduction, convection, and radiation in \unit{\watt\per\square\meter}, and $\mathbf{n}$ is the outward-pointing unit normal vector. Assuming a long, slender geometry, temperature gradients perpendicular to the longitudinal coil axis ($z$-direction) are considered negligible. The temperature field is thus assumed to be uniform across the wire cross-section within each coil segment.
To account for the helical winding, the actual control volume $V_{k}^{f}$ of a discrete segment $k$ with a spatial length of $\Delta z$ is defined by the total enclosed wire volume, yielding $V_{k}^{f} = a_{f} \Delta l_{f} = a_{f} \xi \Delta z$, where $\xi = dl_{f}/dz$ represents the geometric elongation factor, which is the unwound wire length $l_{f}$ per unit length of the coil axis $z$. This directly yields the differential volume relation $dV = a_{f} \xi dz$. Integrating the 3D energy balance over the spatial interval [$\text{lb} = z_{k} - \Delta z /2, \text{ub} = z_{k} + \Delta z /2$] transforms the relation into a one-dimensional continuum form
\begin{multline}
	\int_{\text{lb}}^{\text{ub}} \rho_{f} c_{p,f} a_{f} \xi \frac{\partial T_{f}}{\partial t} \, dz = \\
	\int_{\text{lb}}^{\text{ub}} \left( p_{el}' - q_{r,f}' - q_{c,n}' + q_{gap}' \right) dz +
	\left[ k_{f} a_{f} \frac{1}{\xi} \frac{\partial T_{f}}{\partial z} \right]_{\text{lb}}^{\text{ub}},
	\label{EQ:EnergyBalanceFilamentContinuum1D}
\end{multline}
where the boundary heat flux is split into axial conduction through the cross-sectional end faces of the wire, evaluated via Fourier's law, and radial heat losses across the outer lateral surface of the coil segment. The terms $p_{el}', q_{r,f}', q_{c,n}'$, and $q_{gap}'$ denote the line-source power and loss densities per unit length of the coil axis in \unit{\watt\per\meter}. Assuming spatially uniform material properties within each control volume and evaluating the integrals using segment-averaged values yields in the final discrete conservation equation for segment $k = 1 \dots N_{h}$
\begin{multline}
	m_{f,k} c_{p,f,k}(T_{f,k}) \frac{d T_{f,k}}{d t} = \\
	P_{el,k} - Q_{r,f,k} - Q_{c,n,k} + Q_{cond,f,k} + Q_{gap,k}
	\label{EQ:EnergyBalanceFilament}
\end{multline}
where $P_{el}$ is the term for electrical heat generation, $Q_{r,f}$ is the emitted radiation, $Q_{c,n}$ are the conduction losses to the neon gas, $Q_{cond,f}$ is the conductive heat transport along the coiled wire and $Q_{gap}$ is the combined conductive and radiative heat transfer in the gaps between the windings. All named heat terms are given in \unit{\watt}.
The other variables included are the filament mass $m_{f}$ in \unit{\kilogram}, the filament temperature $T_{f}$ in \unit{\kelvin} and the temperature-dependent specific heat of the tungsten filament $c_{p,f}(T_{f})$ in \unit{\joule\per\kilogram\per\kelvin}, as specified in \textcolor{blue}{Eq.~\protect\ref{EQ:S_SpecificHeatTungsten_R1}} and \textcolor{blue}{Eq.~\protect\ref{EQ:S_SpecificHeatTungsten_R2}} of the supplementary material according the suggestion in \cite{SaRodrigues2022}.
The heat transport terms governing the internal heater dynamics along the $z$-axis are $Q_{cond,f}$ and $Q_{gap}$. Additional heat loss to the thicker contact wires at the start ($k = 1$) and the end ($k = N_{h}$) of the filament coil is accounted for via $Q_{cond,mount}$, which introduces thermal gradients along the entire emitter geometry, causing varying heating rates and thereby capturing the internal heat distribution mechanisms. Here, the lead-in wires are assumed to share the same temperature as the quartz glass envelope at the boundaries, as they are embedded within the material. Furthermore, the thermal mass of the filament coil was adjusted at its boundaries and at the positions of the guiding plates to account for the mounting wires and distance holders shown in \textcolor{blue}{Fig.~\protect\ref{FIG:IRHeaterGeometry}}. Another phenomenon to consider is the heat transport across the inter-turn gaps of the coiled wire, which is particularly relevant during the initial transient immediately after switch-on. The net heat flux across this gap is the sum $Q_{gap} = Q_{cond,gap} + Q_{rad,gap}$ of gas conduction through the neon fill and surface-to-surface radiation between adjacent windings. This coupled treatment is crucial during early transient phases when temperature gradients and property variations are largest. A detailed derivation and the explicit expressions for these conductance and radiation terms are provided in \textcolor{blue}{Section~\protect\ref{SEC:S_HeatTransportEqu}} of the supplementary material.

The supply current input $i_{e}$ controls the electrical heating power of the system which is defined as
\begin{equation}
	P_{el} = i_{e}^{2} R,
	\label{EQ:ElectricPower}
\end{equation}
where the electrical resistance
\begin{equation}
	R = \frac{r_{t}(T_{f}) l_{f}}{a_{f}}
\end{equation}
is expressed as a function of the temperature-dependent resistivity $r_{t}(T_{f})$ of tungsten, the filament length $l_{f}$ and the cross section area $a_{f}$ of the filament wire. The resistivity was also adopted from \cite{SaRodrigues2022,Pettersson2000} and is given in \textcolor{blue}{Eq.~\protect\ref{EQ:S_ResistivityTungsten}} of the supplementary data.
The radiation heat loss of the filament $Q_{r,f} = q_{rad,f}'' A_{f}$ is the product of the net radiation heat flux per unit area $q_{rad,f}''$, which is handled later in the radiation model section, and the emitting surface area $A_{f}$ of the filament coil. As indicated in \textcolor{blue}{Eq.~\protect\ref{EQ:EnergyBalanceFilament}}, each heat term must be evaluated at the corresponding heater segment $k$.

\subsubsection{Neon gas infill}
Between tungsten filament coil and quartz glass tube, the cavity inside is filled with neon.The gas-filled cavity exhibits a significantly lower thermal mass than the surrounding solid components. Consequently, the transient effects within the neon infill are neglected, simplifying the energy balance to a quasi-stationary radial conduction problem with cylindrical solution \cite{Pettersson2000,Incropera2007}
\begin{equation}
	Q_{c,n} = 2 \pi l_{q} k_{n}(T_{n}) \frac{(T_{f} - T_{q})}{\ln{\frac{d_{q,i}}{d_{coil}}}},
\end{equation} 
where $T_{n} = (T_{f} + T_{q})/2$ stands for the neon mean temperature used to determine the temperature-dependent thermal conductivity of neon $k_{n}$, which is defined in \textcolor{blue}{Eq.~\protect\ref{EQ:S_ConductivityNeon}}.

The remaining sizes are the quartz glass tube length $l_{q}$ in \unit{\meter}, the quartz glass temperature $T_{q}$ in \unit{\kelvin}, the inner quartz glass diameter $d_{q,i}$ in \unit{\meter} and the filament coil diameter $d_{coil}$ in \unit{\meter}.

\subsubsection{Quartz glass envelope}
The next step is to establish the energy balance of each segment of the quartz glass body with index $k = 1 \dots N_{h}$. Analogous to the filament domain, the derivation originates form the general three-dimensional energy balance over the quartz volume $V_{k}^{q}$ enclosed by the surface $A_{k}^{q}$
\begin{equation}
	\int_{V_{k}^{q}} \rho_q c_{p,q} \frac{\partial T_q}{\partial t} \, dV = - \int_{A_{k}^{q}} \mathbf{q}'' \cdot \mathbf{n} \, dA.
\end{equation}
Hereby, a long,slender geometry and a constant temperature across the thickness $t_{q}$ of the glass tube are assumed. Following the same spatial integration and discretization procedure as detailed for the filament domain, this lead to \cite{SaRodrigues2022,Pettersson2000}
\begin{multline}
	m_{q,k} c_{p,q,k}(T_{q,k}) \frac{d T_{q,k}}{d t} = Q_{r,f,k} + Q_{c,n,k} - Q_{r,q,k} \\
	- \overline{h}_{q} A_{q,k} (T_{q,k} - T_{\infty,c}) + Q_{cond,q,k}
	\label{EQ:EnergyBalanceQuartz}
\end{multline}
where $m_{q}$ is the quartz glass mass in \unit{\kilogram} and $c_{p,q}$ the temperature-dependent specific heat of quartz in \unit{\joule\per\kilogram\per\kelvin}, which is defined in the supplementary \textcolor{blue}{Eq.~\protect\ref{EQ:S_SpecificHeatQuartz}} according the authors of \cite{SaRodrigues2022}. The heat terms given in \unit{\watt} on the right-hand side of the equation are the incoming radiation heat of the filament coil $Q_{r,f}$, the outgoing radiation heat $Q_{r,q}$ at the outer quartz glass surface, the neon conduction heat $Q_{c,n}$ and the axial quartz conduction heat $Q_{cond,q}$. The convective heat is defined by the average convection coefficient $\overline{h}_{q}$ in \unit{\watt\per\square\meter\per\kelvin}, the outer glass tube surface $A_{q}$ in \unit{\square\meter} and the difference between quartz glass temperature $T_{q}$ and the air temperature inside the radiation cavity $T_{\infty,c}$ in \unit{\kelvin}.
Consistent with the approach used for the tungsten coil, heat conduction along the glass cylinder is accounted for via the heat flow $Q_{cond,q}$. The underlying governing equations and detailed parameters are provided in \textcolor{blue}{Section~\protect\ref{SEC:S_HeatTransportEqu}} of the supplementary material.
The convection coefficient adopts the suggestions of \cite{SaRodrigues2022} for a cylinder with free convection by using the correlation
\begin{equation}
	\overline{h}_{q} = \frac{\overline{Nu}_{q} k_{air}(T_{film})}{d_{q,e}},
\end{equation}
where
\begin{equation}
	\overline{Nu}_{q} = \left\{ \num{0.6} + \frac{\num{0.387} Ra_{q}^{1/6}}{\left[ 1 + \left( \num{0.559}/Pr_{air}(T_{film}) \right)^{9/16} \right]^{8/27}} \right\}^{2}
\end{equation}
is the corresponding average Nusselt number, $k_{air}$ the thermal conductivity from \textcolor{blue}{Eq.~\protect\ref{EQ:S_ThermalConductivityAir}} and
\begin{equation}
	Ra_{q} = \frac{g \left( 1/T_{film} \right) \left( T_{q} - T_{\infty,c} \right) d_{q,e}^{3}}{\nu_{air} \left( T_{film} \right) \alpha_{air} \left( T_{film} \right)}
\end{equation}
the used Rayleigh number.

Due to its negligible thermal mass, the thin reflective gold layer applied to the rear of the double quartz glass tube is assumed to reach instantaneous thermal equilibrium with the glass substrate and therefore $T_{r} = T_{q}$ is used for further considerations.

\subsection{Enclosure radiation model}
This section formulates the enclosure radiation model that uses the time- and space-dependent temperatures of the tape surfaces and infrared emitters obtained in the preceding sections. Based on Planck's law \cite{modest2021radiative}, which describes spectral emission $E$, and combining it with appropriate view factors $F_{\theta-\psi}$, the irradiation $G$ of each element is computed. Together with the optical properties like emissivity $\epsilon$, transmissivity $\tau$ and reflectivity $r$, this yields the radiosity $J$ and, finally, the net radiative heat-flux density $q_{rad}''$ for every surface patch. These fluxes serve as boundary conditions (\textcolor{blue}{Eq.~\protect\ref{EQ:VolumetricBoundaryHeatFlux}}) in the tape heat-conduction model (\textcolor{blue}{Eq.~\protect\ref{EQ:HeatCondEquSpatialDiscrete}}) and are used for calculating the radiation heat in the energy balances of the tungsten filament defined in \textcolor{blue}{Eq.~\protect\ref{EQ:EnergyBalanceFilament}} and the quartz glass tube in \textcolor{blue}{Eq.~\protect\ref{EQ:EnergyBalanceQuartz}}.

\subsubsection{Modeling basis}
\begin{figure}
	\centering
	\includegraphics[width=0.7\columnwidth]{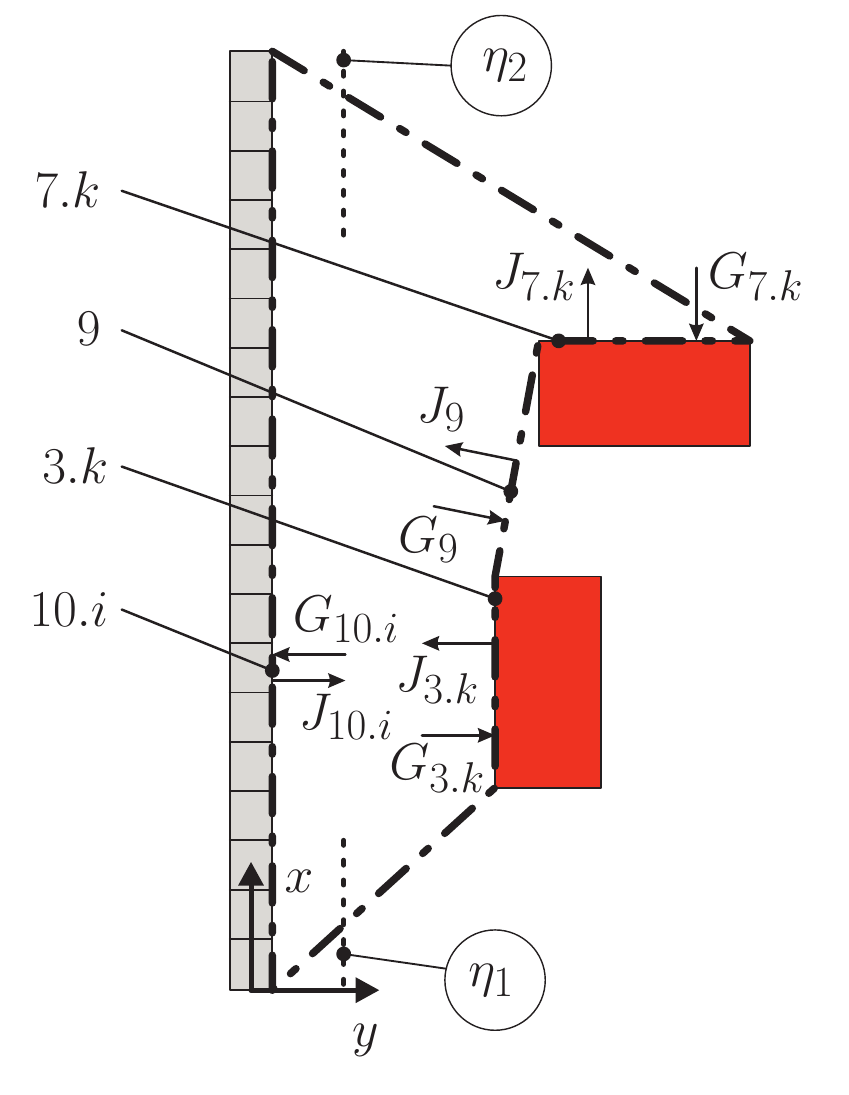}
	\caption{Surface assignment for outer radiation cavity. (3.k, 7.k) Segment $k$ of the simplified rectangular heater surface for emitter 1 or 2, respectively. (9) Surrounding surfaces. (10.i) UD tape surface of segment $i$. ($\eta_{1}, \eta_{2}$) Blockage factors accounting for geometrical obstruction due to tape guide plates.}
	\label{FIG:OuterRadiationCavity}
\end{figure}

\begin{figure}
	\centering
	\includegraphics[width=0.85\columnwidth]{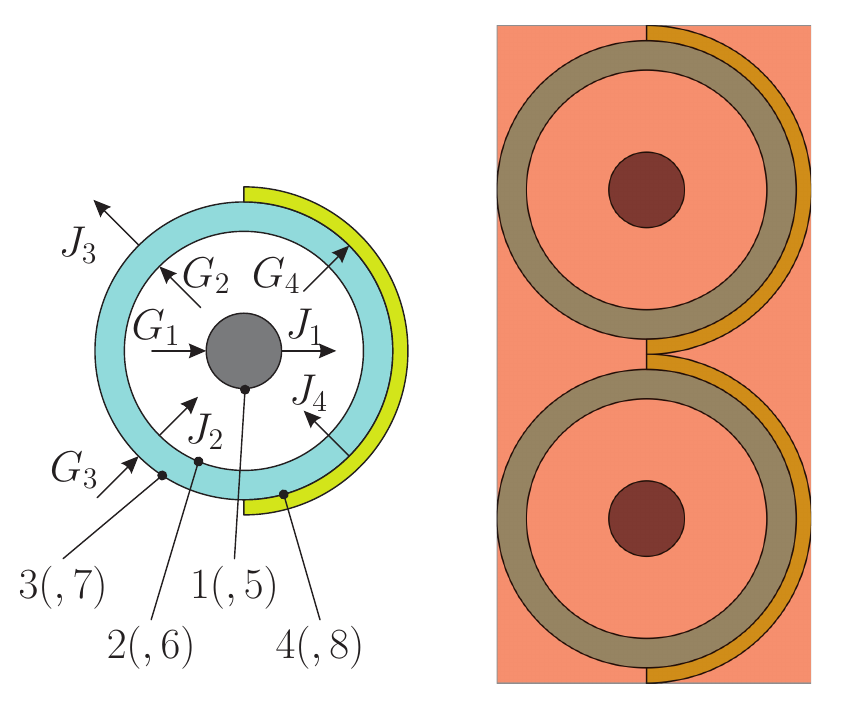}
	\caption{Left: Surface assignment for radiation within heater. (1,5) Filament coil surfaces. (2,6) Internal quartz glass surfaces. (3,7) External quartz glass surfaces. (4,8) Internal gold reflector surfaces. [(,\dots) indicates surface index of second heater] Right: Rectangular emitter simplification.}
	\label{FIG:TubeCrossSection}
\end{figure}

This subsection defines the enclosure used to represent the surfaces within the radiation cavity. The model incorporates the tape segments, the infrared heater surfaces, and the surroundings, the latter of which is modeled as a black body to account for dissipated radiative energy. The definition of the outer radiation cavity uses simplified rectangular radiation sources as a replacement for the double-tube geometry to simplify the view factor calculation. The small distances between the emitters and the tape justify this simplified approach. The outer and inner radiation cavities, along with the rectangular surrogate, are illustrated in \textcolor{blue}{Fig.~\protect\ref{FIG:OuterRadiationCavity}} and \textcolor{blue}{Fig.~\protect\ref{FIG:TubeCrossSection}}, respectively. Here, \textcolor{blue}{Fig.~\protect\ref{FIG:TubeCrossSection}} provides a detailed cross-section view of the infrared heater shown in \textcolor{blue}{Fig.~\protect\ref{FIG:IRHeaterGeometry}}. The outer radiation cavity is defined along the tape surface facing the cavity, the surroundings and the simplified rectangular heater surfaces as it is shown in \textcolor{blue}{Fig.~\protect\ref{FIG:OuterRadiationCavity}} with the dashed line. To reduce computation time, the twin-tube geometry was initially simplified to a single tube. Because the heater filaments of one radiator are connected in series and carry the same current, an identical temperature profile is assumed for the adjacent tube structure, significantly simplifying the radiative heat flux calculations.
To preserve an accurate energy balance within this symmetrical setup, a hybrid view factor approach is implemented. By defining effective view factors based on a rectangular surrogate geometry, the model seamlessly accounts for both the combined emission of the twin tubes and the split absorption of external irradiation without altering the underlying geometric solver matrix. All surfaces are modeled as gray bodies, with the exception of the surroundings, which are treated as a black body, and the quartz glass, which is considered semitransparent.

\subsubsection{Surface discretization and view factors}
In order to correctly determine the radiation fluxes between the surfaces involved, suitable view factors $F_{\theta-\psi}$ are introduced. Here, $F_{\theta-\psi}$ represents the fraction of radiation leaving surface $\theta$ that is directly intercepted by surface $\psi$. Following the numbering defined in \textcolor{blue}{Fig.~\protect\ref{FIG:OuterRadiationCavity}} and \textcolor{blue}{Fig.~\protect\ref{FIG:TubeCrossSection}} yields a predefined surface sequence $\theta = 1 \dots (2\cdot 4 \cdot N_{h} + N_{x} + 1)$ including two times the $N_{h}$ segments of all four surfaces of the heaters, the $N_{x}$ surfaces of the tape and the surroundings. This geometric relationship obeys the reciprocity theorem ($A_{\theta} F_{\theta-\psi} = A_{\psi} F_{\psi-\theta}$), ensuring energy conservation within the cavity.
\begin{figure}
	\centering
	\includegraphics[width=0.75\columnwidth]{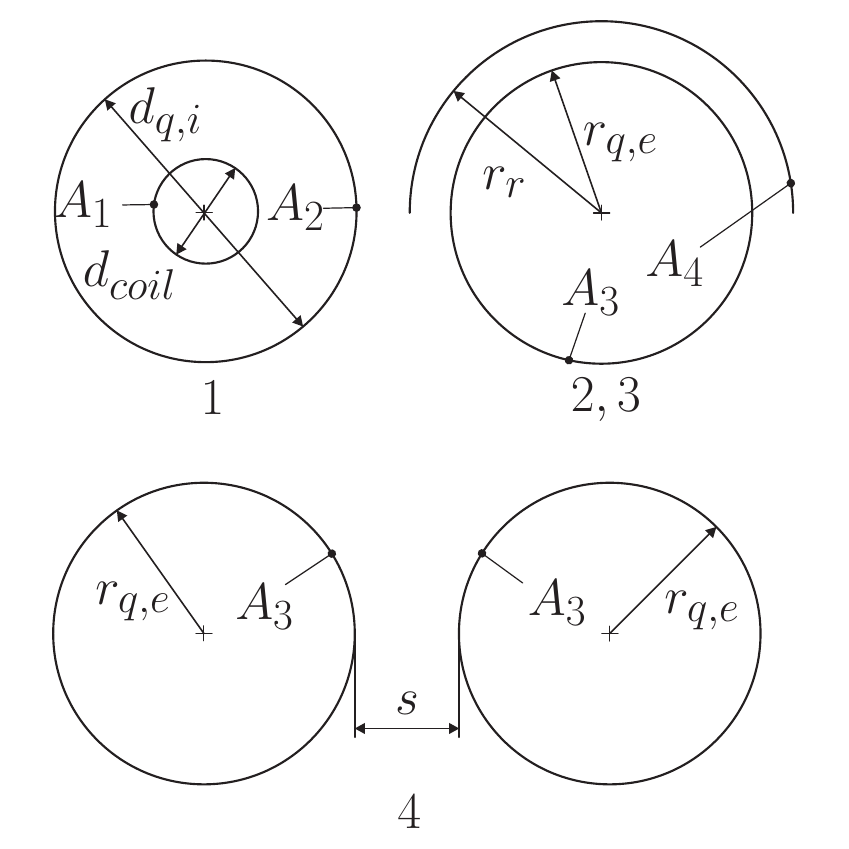}
	\caption{Geometry basement for view factor calculation of cylindrical surfaces. (1) Concentric cylinders of infinite length. (2) Infinitely long cylinder to interior of a concentric semi-cylinder. (3) Infinitely long semi-cylinder to itself with presence of a concentric, coaxial cylinder. (4) Infinitely long parallel cylinders of same diameter with distance $s$. ($d_{q,e} = 2 r_{q,e}$, $d_{r} = 2 r_{r}$)}
	\label{FIG:ViewFactorsCylinder}
\end{figure}
\begin{figure}
	\centering
	\includegraphics[width=0.75\columnwidth]{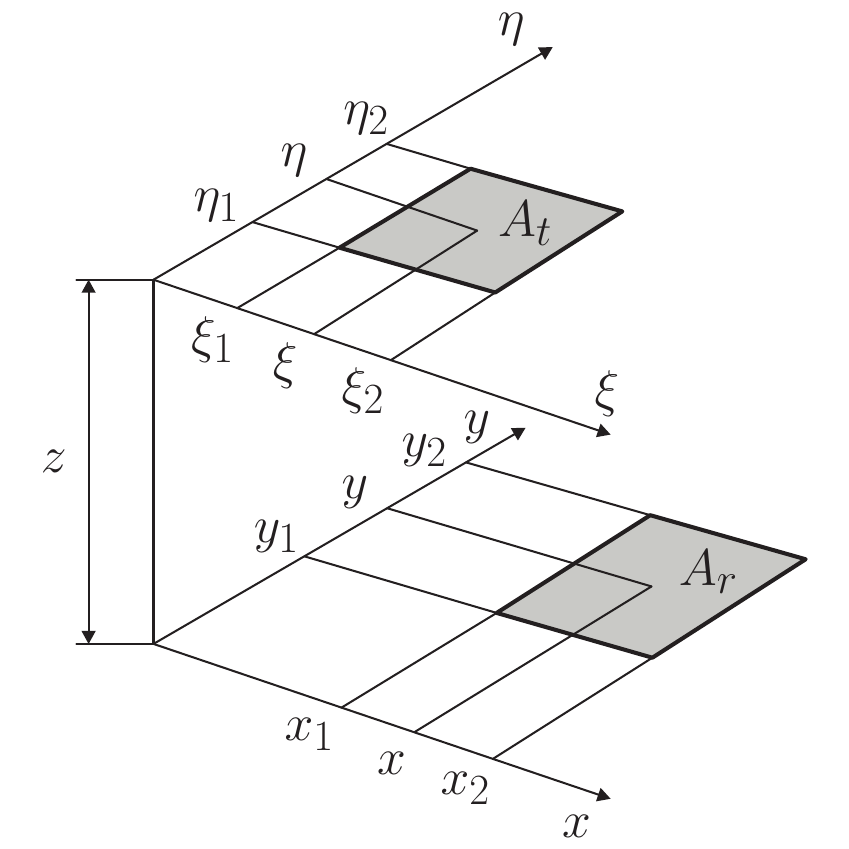}
	\caption{Geometry basement for view factor calculation of rectangle to rectangle in a parallel plane. $r$ represents a segment of the rectangular simplification of the double-tube geometry shown in \textcolor{blue}{Fig.~\protect\ref{FIG:OuterRadiationCavity}} and $t$ a segment of the tape.}
	\label{FIG:ViewFactorsParallelPlanes}
\end{figure}
\begin{figure}
	\centering
	\includegraphics[width=0.75\columnwidth]{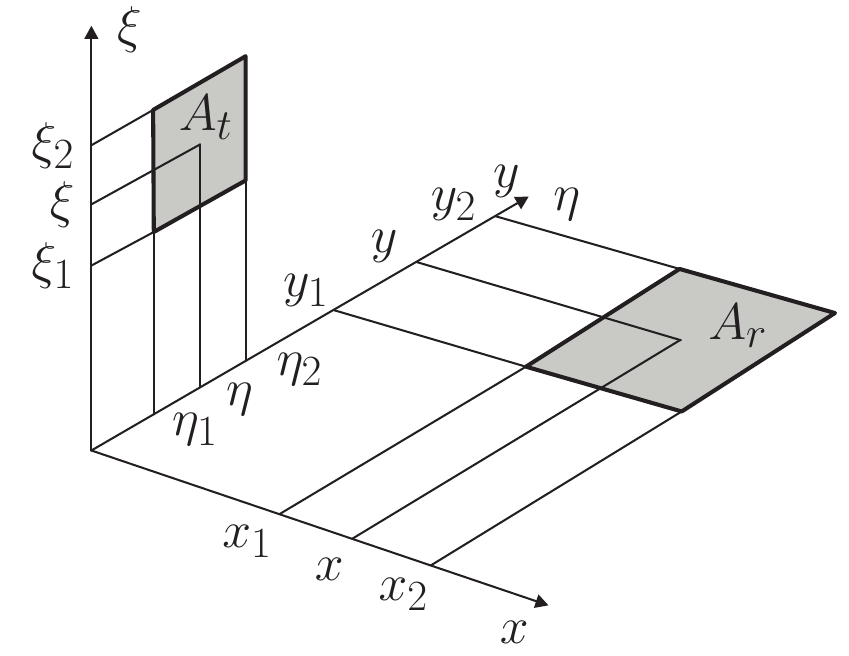}
	\caption{Geometry basement for view factor calculation of rectangle to rectangle in a perpendicular plane. $r$ represents a segment of the rectangular simplification of the double-tube geometry shown in \textcolor{blue}{Fig.~\protect\ref{FIG:OuterRadiationCavity}} and $t$ a segment of the tape.}
	\label{FIG:ViewFactorsPerpendicularPlanes}
\end{figure}
\textcolor{blue}{Fig.~\protect\ref{FIG:ViewFactorsCylinder}} shows the geometric basement for following definitions with respect to the cylindrical surfaces, whereas \textcolor{blue}{Fig.~\protect\ref{FIG:ViewFactorsParallelPlanes}} and \textcolor{blue}{Fig.~\protect\ref{FIG:ViewFactorsPerpendicularPlanes}} depicts the geometry for calculating the relations between parallel or perpendicular planes, respectively. Although the infrared heater is discretized along the tube axis, view factors between the filament surfaces and the internal quartz glass surfaces were determined for concentric cylinders of infinite length \cite{hottel1931radiant,howell2020thermal} using
\begin{align}
	F_{2-2} = 1 - \frac{d_{coil}}{d_{q,e} - 2 t_{q}}, \\
	F_{1-2} = 1, \\
	F_{2-1} = \frac{d_{coil}}{d_{q,e} - 2 t_{q}}.
\end{align}
The view factor between external quartz glass surface and gold reflector is obtained from the results presented in \cite{howell2020thermal} for an infinitely long cylinder to the interior of a concentric semi-cylinder and reads
\begin{equation}
	F_{3-4} = \frac{1}{2},
\end{equation}
whereas the reverse case is defined by
\begin{equation}
	F_{4-3} = 1,
	\label{EQ:F_43}
\end{equation}
assuming the whole gold reflector surface sees only the quartz glass, because it is directly applied on the tubes ($r_{r} = r_{q,e}$). Although only one of the tubes was used for modeling the emitters, the second one is taken into account in the radiation balance by applying the view factor
\begin{align}
	& F_{3-3} = \frac{1}{\pi}\left( \sqrt{X^{2} - 1} + \arcsin \left( \frac{1}{X} \right) - X  \right) \\
	& X = 1 + \frac{s}{2 d_{q,e}},
\end{align}
which considers two infinitely long parallel cylinders of same diameter with distance $s$ (here $s = 0$) as proposed in \cite{howell2020thermal,Sparrow1978,bopche2011local}. The general consideration of the interior correlation of an infinitely long semi-cylinder to itself with presence of a concentric, coaxial cylinder, as it occurs with the gold reflector layer, is given by the equation
\begin{align}
	& F_{4-4} = 1 - \frac{2}{\pi} \left[ (1 - R^{2})^{1/2} + R \arcsin(R) \right] \\
	& R = \frac{d_{q,e}}{d_{r}}
\end{align}
as stated in \cite{howell2020thermal}. However, due to the directly applied layer structure, where the external lamp diameter equals the one of the gold reflector ($d_{q,e} = d_{r}$), this view factor degenerates to $F_{4-4} = 0$ and strengthens the assumption made in \textcolor{blue}{Eq.~\protect\ref{EQ:F_43}}.
Other approaches, such as \cite{SaRodrigues2022}, define view factors between the heater and tape using correlations for infinitely long cylinders and rectangles \cite{feingold1970new,hamilton1952radiant} or by applying Hottel's cross-string method for 2D scenarios, as seen with their rectangular reflector and the tape pieces. In contrast, this work emphasizes methods that account for geometric depth differences to capture the interactions between the involved surfaces as accurately as possible. Without this distinction, the loss of radiation emitted by the heaters across the tape would be fully accounted for in the heat input of the material strip, leading to significant deviations from reality. Therefore, this approach uses a view factor definition that accounts for spatial expansions taking the extent of surfaces in depth into account when modeling the correlation between the emitter and tape segments.
As mentioned in the previous sections, a fully parallel alignment of the first heat source and the material strip is assumed. This configuration yields the view factor formulation for parallel rectangles $F_{r-t}$, which is detailed in \textcolor{blue}{Eq.~\protect\ref{EQ:S_F_PlanarPlanes}} and \textcolor{blue}{Eq.~\protect\ref{EQ:S_F_rt_parallel}} within \textcolor{blue}{Section~\protect\ref{SEC:S_ViewFactors}} of the supplementary material, including the corresponding literature references. By properly applying the geometric parameters, this relation is applicable in both directions ($F_{r-t},F_{t-r}$). For the second case, where the emitter is perpendicularly aligned with the tape, the formulations for perpendicular rectangles apply. Utilizing the framework of \textcolor{blue}{Eq.~\protect\ref{EQ:S_F_PlanarPlanes}}, the corresponding perpendicular view factor can be computed simply by substituting \textcolor{blue}{Eq.~\protect\ref{EQ:S_F_rt_parallel}} with \textcolor{blue}{Eq.~\protect\ref{EQ:S_F_rt_perpendicular}}.
To justify the necessity of this 3D approach, the proposed model was evaluated against a simplified 1.5D radiation framework. Assuming an infinitely long configuration of parallel plates \cite{wong1977handbook} completely neglects three-dimensional end-effects and radiation losses that occur because the heater's length extends perpendicularly beyond the tape's width. For the real geometric dimensions ($l_{q} \approx 2.5 w_{m}$), this 1.5D simplification overestimates the real radiation exchange area ($A_{\theta} F_{\theta - \psi}$) by \qty{159}{\percent}. This comparison demonstrates that a 1.5D approach introduces unacceptable errors, proving that a 3D formulation is strictly required to capture these lateral losses. Additionally, the effect on the view factor caused by the cylindrical shape of the emitter was evaluated by comparing the deviation when using an infinitely long parallel plate model and an infinitely long cylinder-to-plate configuration \cite{feingold1970new}.
Although flattening the curved emitter profile to a 1.5D plate overestimates the radiation exchange area by another \qty{29}{\percent}, the real double-tube geometry is non-ideal, representing a transitional shape between parallel tubes and a flat rectangle (see \textcolor{blue}{Fig.~\protect\ref{FIG:IRHeater}}). Therefore, the rectangular simplification remains appropriate for describing the actual system while maintaining geometric proximity.

Summarizing all view factors in accordance with the strictly defined sequence $\theta$ of surface assignments yields the view factor matrix $\mathbf{F}$. This matrix accounts for all relations between the involved surfaces, including the surroundings. As the determination of the relations to each surrounding segment is complex, the view factors interacting with these areas are initially distributed equally to fulfill the geometric view factor summation rule $\sum_{\psi} F_{\theta-\psi} = 1$ for the unscaled domain.
Additionally, the view factors must be adjusted according to the rectangular surrogate assumption of the twin tubes. This is achieved by modifying the corresponding view factors via a mathematical projection based on the view factor reciprocity property ($A_{\theta} F_{\theta-\psi} = A_{\psi} F_{\psi-\theta}$). While the outer glass tube surfaces ($A_{3} = A_{q}$) interact analytically with the adjacent tube ($F_{3-3}$) and the gold reflector ($F_{3-4}$), their radiative coupling to the surroundings and the tape segments $i$ utilizes effective view factors ($F_{\theta-\psi}^{*}$) derived from a dedicated rectangular surrogate geometry ($r$).

The scaling factors arise directly from the geometric relationship between the gold-uncovered surface area of a single tube ($A_{3}/2 = A_{q}/2$) and the total area of the rectangular simplification representing the uncovered area of both tubes ($A_{r} = 2 A_{q} /2 = A_{q}$). To maintain thermodynamic consistency, the effective view factors are established by balancing the total energy exchange with the surrogate geometry
\begin{align}
	& \quad \frac{A_{3}}{2} F_{3-\psi}^{*} = A_{r} F_{r-\psi} \rightarrow F_{3-\psi}^{*} = 2 F_{r-\psi} \\
	& \quad A_{\psi} F_{\psi-3}^{*} = \frac{1}{2} A_{\psi} F_{\psi-r} \rightarrow F_{\psi-3}^{*} = \frac{1}{2} F_{\psi-r}.
\end{align}
Due to this deliberate scaling, the direct numerical sum of the tube row exceeds unity ($F_{3-3} + F_{3-4} + \sum_{\psi \neq 3,4} F_{3-\psi}^{*} > 1$).
However, because the enhanced emission ($F_{3-\psi}^{*}$) is mathematically counterbalanced by the split absorption of environmental irradiation ($F_{\psi-3}^{*}$), strict energy conservation is inherently preserved within the solver.

As illustrated in \textcolor{blue}{Fig.~\protect\ref{FIG:ModelingScheme}}, the tape is partially shadowed by the sheet metal guides. To account for this geometric obstruction, the radiation exchange is modeled using effective view factors $F_{\theta-\psi,eff}$ in the affected areas. These are derived by adjusting the theoretical view factors $F_{\theta-\psi}$ with a blockage factor $\eta_{b}$ (where $0 \leq \eta_{b} \leq 1$), such that $F_{\theta-\psi,eff} = \eta_{b} \cdot F_{\theta-\psi}$. This factor represents the proportion of the radiant flux that reaches the tape through the guiding geometry. While the fundamental view factor identities remain theoretically valid, this is approach allows for simplified calculation of the heat balance without the need to define the shielding guides as separate thermal surfaces.

To populate the view factor matrix, the full 3D view factors between all participating surfaces (considering $N_{x}$ tape segments and $N_{h}$ heater segments) must be evaluated, although this comprehensive computation was restricted to the outer radiation cavity defined in \textcolor{blue}{Fig.~\protect\ref{FIG:OuterRadiationCavity}}.

\subsubsection{Spectral emission and gray approximation}
This section establishes the spectral basis for radiative emission and the conditions under which a gray approximation is adequate. The fundamental equation describing the spectral emission of a black-body, defining the hemispherical emissive power in \unit{\watt\per\cubic\meter} at temperature $T$, is Planck's law \cite{modest2021radiative} for a certain wavelength
\begin{equation}
	E_{b,\lambda} = \frac{2 \pi h c_{0}^{2}}{\lambda^{5}} \frac{1}{\exp{\left( \frac{h c_{0}}{\lambda k_{b} T} \right) - 1}},
	\label{EQ:SpectralEmissivePower}
\end{equation}
or for the whole spectrum with unit \unit{\watt\per\square\meter} yielding in Stefan-Boltzmann law
\begin{equation}
	E_{b} = \int_{0}^{\infty} E_{b,\lambda}(T) \diff{\lambda} = \sigma T^{4},
\end{equation} 
respectively. Occurring quantities are
\begin{itemize}
	\item $\lambda \hdots$ wavelength in \unit{\meter},
	\item $T \hdots$ absolute black-body temperature in \unit{\kelvin},
	\item $h \hdots$ Planck constant \qty{6.626e-34}{\joule.\second},
	\item $c_{0} \hdots$ speed of light in vacuum \qty{2.998e8}{\meter\per\second},
	\item $k_{b} \hdots$ Boltzmann constant \qty{1.381e-23}{\joule\per\kelvin},
	\item $\sigma \hdots$ Stefan-Boltzmann const. \qty{5.670e-8}{\watt\per\square\meter\per\kelvin^{4}}.
\end{itemize}
If only a specific spectral range is of interest, the integration borders are adopted to the corresponding lower and upper wavelengths $\lambda_{1}$ and $\lambda_{2}$, which leads to the black-body band radiation
\begin{equation}
	E_{b,\lambda_{1-2}} = \int_{\lambda_{1}}^{\lambda_{2}} E_{b,\lambda}(T) \diff{\lambda}
\end{equation}
given in \unit{\watt\per\square\meter}. Since this integral cannot be solved analytically, an infinite series approximation
\begin{multline}
	E_{b,\lambda_{1-2}} = \sigma T^{4} \Bigg[ \frac{15}{\pi^{4}} \Bigg( \sum_{n = 1}^{\infty} \frac{e^{-n \xi_{2}}}{n} \Bigg( \xi_{2}^{3} + \frac{3 \xi_{2}^{2}}{n} + \frac{6 \xi_{2}}{n^{2}} + \frac{6}{n^{3}} \Bigg) \\
	- \sum_{n = 1}^{\infty} \frac{e^{-n \xi_{1}}}{n} \Bigg( \xi_{1}^{3} + \frac{3 \xi_{1}^{2}}{n} + \frac{6 \xi_{1}}{n^{2}} + \frac{6}{n^{3}} \Bigg) \Bigg) \Bigg],
\end{multline}
evaluated at its first 10 terms with
\begin{align}
	\xi_{1} = \frac{h c_{0}}{k_{b} \lambda_{1} T} \\
	\xi_{2} = \frac{h c_{0}}{k_{b} \lambda_{2} T}
\end{align}
was used as proposed by \cite{SaRodrigues2022} based on the work of \cite{howell2020thermal,pettersson1999heat}. This approach maintains sufficient accuracy while significantly reducing calculation time compared to standard numerical integration. For real surfaces, the spectral emissivity $\epsilon_{\lambda}(\lambda,T)$ has to be integrated into the spectral emissive power calculation
\begin{equation}
	E_{\lambda}(T) = \epsilon_{\lambda}(\lambda,T) E_{b,\lambda}(T)
\end{equation}
leading to
\begin{equation}
	E_{\lambda_{1-2}}(T) = \int_{\lambda_{1}}^{\lambda_{2}}\epsilon_{\lambda}(\lambda,T) E_{b,\lambda}(T) \diff{\lambda}
\end{equation}
for the real band radiative power. However, if it is assumed that the gray approximation can be used, where $\epsilon$ is seen to be constant across the whole wavelength spectrum, the equation can be simplified to
\begin{equation}
	E_{\lambda_{1-2}}(T) = \epsilon(T) \int_{\lambda_{1}}^{\lambda_{2}} E_{b,\lambda}(T) \diff{\lambda}.
\end{equation}
This approximation is appropriate when $\epsilon_{\lambda}(\lambda,T)$ vary slowly over the dominant emission band given by $E_{b,\lambda_{1-2}}(T)$ for all interacting surfaces, for weak spectral filtering, which means that intervening media and windows do not introduce sharp spectral cutoffs within the dominant bands as well as the induced error stays within a tolerable range.

\subsubsection{Optical properties}
Having established the spectral basis and the conditions for gray approximation, now the optical inputs are specified to compute the radiative exchange. \textcolor{blue}{Section~\protect\ref{SEC:S_OpticalProperties}} in the supplementary material compiles the emissivity $\epsilon$, reflectivity $r$ and transmissivity $\tau$ of all participating surfaces, namely the tape, the emitters, the surroundings and any semitransparent components like the quartz glass, including potentially occurring temperature dependencies. These properties satisfy the energy balance
\begin{equation}
	\epsilon(T) + r(T) + \tau(T) = 1
	\label{EQ:EnergyBalanceOptProp}
\end{equation}
and determine how Planck-weighted emission $E_{\lambda_{1-2}}$ and enclosure view factors $F_{\theta-\psi}$ translate into irradiation $G$, radiosity $J$ and net heat flux $q_{rad}''$, which are addressed in the next section. For opaque surfaces \textcolor{blue}{Eq.~\protect\ref{EQ:EnergyBalanceOptProp}} reduces to
\begin{equation}
	\epsilon(T) \approx 1 - r(T),
	\label{EQ:EnergyBalanceOptPropOpaque}
\end{equation}
because of $\tau(T) \approx 0$. Where detailed spectral data are unavailable, temperature-dependent values, calibrated against measurements, are used. Since the IR emitter operates in the short-wave infrared (SWIR) spectrum, the optical properties were specifically characterized for this wavelength range to precisely model the energy absorption. To be able to accurately determine the occurring heat-up rates and the effectiveness of the used gold reflector layer, both the spectral emissivity and transmissivity in the SWIR band are critical.

Given the emissivity and transmissivity of the surfaces, the reflectivity is determined using \textcolor{blue}{Eq.~\protect\ref{EQ:EnergyBalanceOptProp}} for semitransparent media and \textcolor{blue}{Eq.~\protect\ref{EQ:EnergyBalanceOptPropOpaque}} for opaque media.
All optical properties are consolidated into either a diagonal emissivity $\boldsymbol{\epsilon}$ and reflectivity matrix $\boldsymbol{r}$ or a coupling matrix $\boldsymbol{\tau}$ that accounts for thermal interaction between semitransparent surfaces via transmission coefficients, following the previously defined surface sequence.

\subsubsection{Radiosity, irradiation and net radiative heat flux}
Building on the diffuse-gray approximation established in the previous section, this chapter formulates radiative exchange in the enclosure using radiosity $J$ and irradiation $G$. Under the gray assumption, each surface element is characterized by a temperature $T_{\theta}$, a total hemispherical emissivity $\epsilon_{\theta}$, a transmissivity factor $\tau_{\theta}$ for transparent surfaces, the corresponding reflectivity $r_{\theta}$, and purely geometric view factors $F_{\theta-\psi}$. Spectral variation is subsumed into $\epsilon_{\theta}$, so the exchange can be written compactly without wavelength dependence. The key quantities and relations describing the radiation inside the defined cavity are the radiosity
\begin{equation}
	J_{\lambda_{1-2},\theta} = \epsilon_{\theta} E_{b,\lambda_{1-2},\theta} + r_{\theta} G_{\lambda_{1-2},\theta} + \tau_{\psi-\theta} G_{\lambda_{1-2},\psi},
\end{equation}
which is the hemispherical radiant flux leaving surface $\theta$ per unit area, as well as the irradiation
\begin{equation}
	G_{\lambda_{1-2},\theta} = \sum_{\psi=1}^{N} F_{\theta-\psi} J_{\lambda_{1-2},\psi},
	\label{EQ:Irradiation}
\end{equation}
that defines the hemispherical radiant flux incident on surface $\theta$ per unit area. \textcolor{blue}{Eq.~\protect\ref{EQ:Irradiation}} is obtained by substituting the view factor reciprocity property
\begin{equation}
	A_{\theta} F_{\theta-\psi} = A_{\psi} F_{\psi-\theta}
\end{equation}
into the irradiation balance
\begin{equation}
	A_{\theta} G_{\lambda_{1-2},\theta} = \sum_{\psi = 1}^{N} A_{\psi} J_{\lambda_{1-2},\psi} F_{\psi-\theta}
\end{equation}
of a specific surface, where $A_{\theta}$ and $A_{\psi}$ are the corresponding surfaces areas.
The matrix form incorporates all surfaces of the individual radiosity and irradiation components as follows:
\begin{align}
	& \quad \mathbf{J}_{\lambda_{1-2}} = \boldsymbol{\epsilon} \mathbf{E}_{\lambda_{1-2}} + \mathbf{r} \mathbf{G}_{\lambda_{1-2}} + \boldsymbol{\tau} \mathbf{G}_{\lambda_{1-2}}
	\label{EQ:RadiosityMatrix} \\
	& \quad \mathbf{G}_{\lambda_{1-2}} = \mathbf{F} \mathbf{J}_{\lambda_{1-2}}.
	\label{EQ:IrradiationMatrix}
\end{align}
To be able to compute the irradiation, \textcolor{blue}{Eq.~\protect\ref{EQ:RadiosityMatrix}} is inserted into \textcolor{blue}{Eq.~\protect\ref{EQ:IrradiationMatrix}}, which yields the equation system
\begin{equation}
	\mathbf{G}_{\lambda_{1-2}} = \left( \mathbf{I} - \mathbf{F} \left( \mathbf{r} + \boldsymbol{\tau} \right) \right)^{-1} \mathbf{F} \boldsymbol{\epsilon} \mathbf{E}_{\lambda_{1-2}}
\end{equation}
providing the irradiation vector $\mathbf{G}_{\lambda_{1-2}}$. With known $\mathbf{G}_{\lambda_{1-2}}$, the radiosity defined in \textcolor{blue}{Eq.~\protect\ref{EQ:RadiosityMatrix}} can be solved. Finally, the net radiative heat-flux density then yields
\begin{equation}
	\mathbf{q}_{rad}'' = \mathbf{J}_{\lambda_{1-2}} - \mathbf{G}_{\lambda_{1-2}}
	\label{EQ:NetRadiationMatrix}
\end{equation}
and is counted positive when leaving the corresponding surface. Depending on the predefined surface sequence mentioned in previous sections, it is possible to extract the surface specific radiation heat flow from $\mathbf{q}_{rad}''$ to obtain $\mathbf{q}_{rad,m}''$, $\mathbf{q}_{rad,q}''$ and $\mathbf{q}_{rad,f}''$.

\subsection{Coupling and numerical implementation}
\label{SEC:NumericalImplementation}
This section addresses the numerical time integration of the coupled thermo-radiative problem. Physically, the combination of transient heat conduction (\textcolor{blue}{Eq.~\protect\ref{EQ:HeatCondEquSpatialDiscrete}}, \textcolor{blue}{Eq.~\protect\ref{EQ:DeltaTm_ODE_SpatialDiscrete}}, \textcolor{blue}{Eq.~\protect\ref{EQ:EnergyBalanceFilament}}, \textcolor{blue}{Eq.~\protect\ref{EQ:EnergyBalanceQuartz}}) and geometric radiation exchange (\textcolor{blue}{Eq.~\protect\ref{EQ:RadiosityMatrix}}, \textcolor{blue}{Eq.~\protect\ref{EQ:IrradiationMatrix}}, \textcolor{blue}{Eq.~\protect\ref{EQ:NetRadiationMatrix}}) naturally forms a semi-discrete differential-algebraic equation (DAE) system. While the spatial discretization of the energy balances yields differential equations for the temperature field, the radiosity and irradiation relations introduce purely algebraic constraints. Because these radiative quantities contain no time derivatives, they must remain in instantaneous equilibrium with the thermal state at any given point.

Rather than solving this coupled formulation as a DAE, the system is reduced to a purely ordinary differential equation (ODE) system by analytically eliminating the algebraic variables beforehand. By sequentially resolving the radiation exchange, the net radiative heat flux is expressed as a direct, non-linear function of the thermal state vector $\mathbf{x}$. This reduces the global state vector to include only the nodal temperatures and their associated thermal differences, which is then integrated in time using a stiffly accurate Radau IIA implicit Runge-Kutta method.
This significantly decreases the size of the linear systems to be solved at each implicit stage. To handle the resulting non-linear complexity and ensure fast convergence of the implicit solver, CasADi \cite{Andersson2019} is utilized. It provides exact analytic Jacobians of the entire nested evaluation chain via algorithmic differentiation, ensuring both maximum computational efficiency and high numerical stability.

\subsubsection{Time integration scheme}
The system's transient behavior is computed using a self-implemented Radau IIA method, an implicit Runge-Kutta scheme of order $2 s - 1$, where $s$ is the number of stages. For this simulation, a 2-stage configuration ($s = 2$) of order 3 was chosen due to its excellent stability properties for stiff differential equations.
The state vector
\begin{equation}
	\mathbf{x}^{\mathsf{T}} = 
	\begin{bmatrix}
	\mathbf{T}_{m}^{\mathsf{T}} & \Delta \mathbf{T}_{m}^{\mathsf{T}} & \mathbf{T}_{f}^{\mathsf{T}} & \mathbf{T}_{q}^{\mathsf{T}}
	\end{bmatrix},
\end{equation}
containing all integrated variables, evolves according to the system of ordinary differential equations (ODE) $\dot{\mathbf{x}} = f(t,\mathbf{x},\mathbf{u})$. The state derivative vector $\dot{\mathbf{x}}$ is constructed by assembling the individual balance equations defined in \textcolor{blue}{Eq.~\protect\ref{EQ:HeatCondEquSpatialDiscrete}}, \textcolor{blue}{Eq.~\protect\ref{EQ:DeltaTm_ODE_SpatialDiscrete}}, \textcolor{blue}{Eq.~\protect\ref{EQ:EnergyBalanceFilament}} and \textcolor{blue}{Eq.~\protect\ref{EQ:EnergyBalanceQuartz}} of all thermal nodes. For each thermal node $n$, the derivative is defined by the net heat flow rates $\dot{x}_{n} = \frac{1}{C_{n}} \sum Q_{j}(t,\mathbf{x},\mathbf{u})$, with $C_{n} = m_{n} c_{p,n}$ being the heat capacity of node $n$ and $Q_{j}$ the different heat flow terms. For both the initial ($t = \qty{0}{\second}$) and boundary conditions ($x = \qty{0}{\meter}$), the temperature of the incoming tape $T_{m}$, the filament temperature $T_{f}$ and the quartz glass temperature $T_{q}$ are set to a measured steady-state value of approximately \qty{40}{\celsius} observed after several cycles. Since the system was allowed to thermally equilibrate before measurement, the tape is assumed to be homogeneously preheated. Consequently, the through-thickness temperature difference $\Delta T_{m}$ is initially set to zero. External inputs, such as heater supply current or tape velocity, are represented by the time-dependent vector
\begin{equation}
	\mathbf{u}^{\mathsf{T}}(t) = 
	\begin{bmatrix}
		v_{T} & i_{e}
	\end{bmatrix}.
\end{equation}
These inputs are evaluated at specific intermediate time points within the stage calculations of the Radau scheme to ensure high temporal accuracy. Specifically, the implicit 2-stage Radau IIA method utilizes a constant time-step size. At each time step, the implicit stage equations are solved via a CasADi-based Newton-Raphson algorithm. Convergence is determined by a strict absolute residual tolerance of \num{5e-4}, while the maximum number of iterations is preset to 5. A relative tolerance is omitted due to prior problem-specific state normalization.

\subsubsection{Model coupling and radiation integration}
The thermal model is coupled with the radiative network equations using a monolithic approach. This coupling is required to ensure numerical robustness against the $T^{4}$ non-linearities inherent in radiative heat transfer. Rather than employing a partitioned scheme, the radiative coupling is directly embedded within the system function $f(t,\mathbf{x},\mathbf{u})$ through the term $Q_{rad}$. This term is evaluated using the current state of the entire thermal vector $\mathbf{x}$ to fully account for mutual inter-reflections and view-factors.

Mathematically, the radiation calculation forms a consecutive, nested part of the ODE system function. In each iteration of the implicit solver's nonlinear solver loop, these relations are evaluated to update the net radiative heat flow rates based on the current stages. This tight numerical integration ensures that the L-stability of the Radau IIA scheme is fully preserved. Consequently, the solver can utilize larger time steps while maintaining physical consistency and strict convergence, even in the presence of high thermal gradients.

\section{Experimental setup and evaluation}
\label{SEC:Evaluation}
\begin{figure}
	\centering
	\includegraphics[width=0.8\columnwidth]{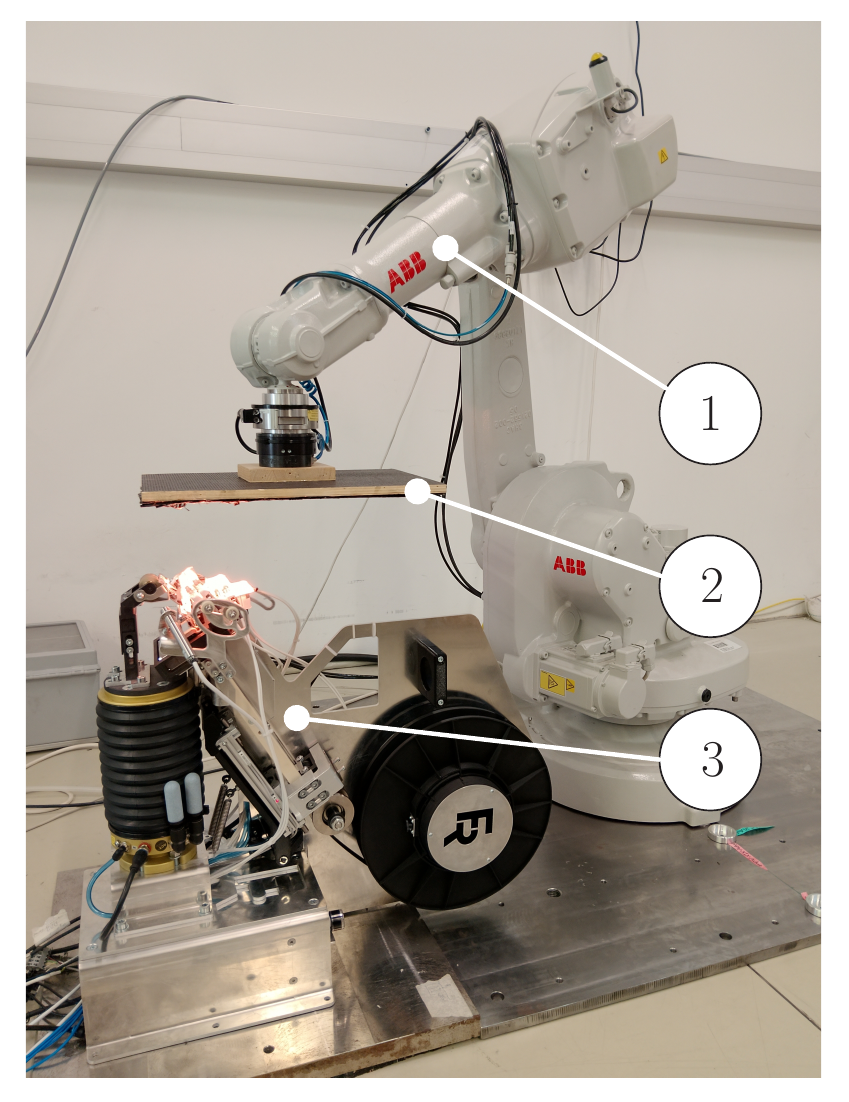}
	\caption{Experimental setup with plane test object as mold. (1) Handling robot. (2) Test mold. (3) Automated tape laying system (ATL).}
	\label{FIG:ExperimentalSetup}
\end{figure}

This section documents the experimental setup and the evaluation procedure used to validate the thermo-radiative model. The test rig, shown in \textcolor{blue}{Fig.~\protect\ref{FIG:ExperimentalSetup}}, comprises an ATL unit mounted spatially fixed on the floor and an industrial robot for handling the mold and placing the tape onto it during the process. The path and trajectory planning for this configuration between spatially fixed TCP and handling robot is addressed in \cite{rameder2025robot}.

\begin{figure}
	\centering
	\includegraphics[width=0.95\columnwidth]{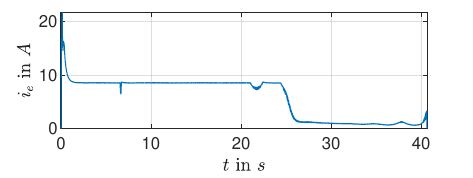}
	\caption{Supply current measurement $i_{e}$ during test trajectory.}
	\label{FIG:SupplyCurrentMeas}
\end{figure}

\begin{figure}
	\centering
	\includegraphics[width=0.95\columnwidth]{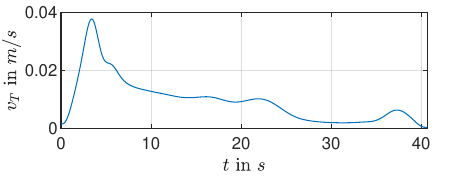}
	\caption{Tape velocity $v_{T}$ during test trajectory.}
	\label{FIG:TapeVelocityMeas}
\end{figure}

\begin{figure}
	\centering
	\includegraphics[width=0.9\columnwidth]{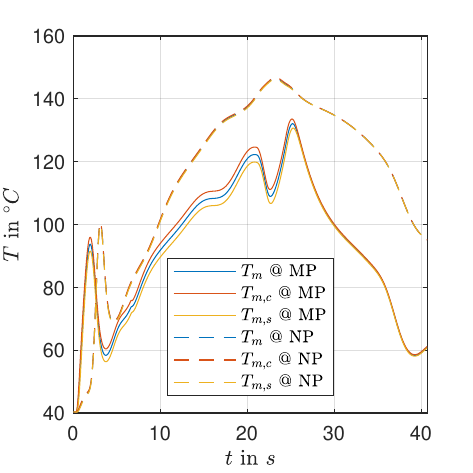}
	\caption{Temperature simulation at measuring point (MP) and nip point (NP) on both sides of tape during test trajectory.}
	\label{FIG:TapeTemperatures}
\end{figure}

To create a reference for validating the model, a trajectory with dynamic transitions was performed, while simultaneously tracking the supply current $i_{e}$ applied to the infrared heaters and the temperature $T_{m,s,meas}$ on a specific measuring point on the tape side facing the surroundings. This trajectory was generated using a preliminary thermal model in an optimization problem to achieve the minimum-time path by utilizing the maximum available heating power. Consequently, it serves as an ideal test case for evaluating the model presented here, as it provides the dynamic inputs to validate the model's performance. The corresponding measurements are shown in \textcolor{blue}{Fig.~\protect\ref{FIG:SupplyCurrentMeas}} and \textcolor{blue}{Fig.~\protect\ref{FIG:TapeVelocityMeas}}, where both include dynamic transitions especially at the beginning of the trajectory.
The measurement of current $i_{e}$ serves as input for the thermodynamic model especially in \textcolor{blue}{Eq.~\protect\ref{EQ:ElectricPower}}, where the electric power, that forces the internal heat generation in the filament coil, is calculated, whereas $v_{T}$ is used for determining the advection term in \textcolor{blue}{Eq.~\protect\ref{EQ:HeatCondEquSpatialDiscrete}} and \textcolor{blue}{Eq.~\protect\ref{EQ:DeltaTm_ODE_SpatialDiscrete}} as well as for forced convection phenomena. Since the measurement captures the total supply current for two twin-tube heaters in parallel, it is halved in the model to represent the current per individual branch. The result of the simulation concerning $T_{m,c}$, $T_{m,s}$ and $T_{m}$ using the named input properties leads to the temporal course evaluated at the measuring point (MP) and the nip point (NP) depicted in \textcolor{blue}{Fig.~\protect\ref{FIG:TapeTemperatures}}. The effect of the heat flow through the tape thickness modeled in \textcolor{blue}{Eq.~\protect\ref{EQ:DeltaTm_ODE_SpatialDiscrete}} can clearly be seen in this figure, where two different effects can be observed. First, the temperature difference between the opposing surfaces varies because of the heating and acceleration processes. At the begin of the considered time horizon, the materials temperature is consistent all over its distinction. Since the emitters heat up, leading to increased radiation heat impacting the tape, the gap between the different sides widens. The decreasing velocity after the initial phase further amplifies this reaction since the exposure of the tape beneath the heaters increases. After a certain time, the supply current is lowered to a relatively small value, which leads to the temperature gap being closed again in the final phase of the process. Besides the temporal effect, the position in the domain also significantly affects the difference between the two surface temperatures. Positions close to the start or end of the domain show hardly any differences in the surface temperatures of the opposing surfaces, which is mainly caused by the shielding plates in this areas as well as the specific alignment of the infrared heaters leading to weaker exposure at the named regions. In the center of the domain, where the parallel infrared heater can fully interact with the material in absence of any shielding surfaces, the gap between the side facing the radiation cavity and the one facing the surroundings increases noteworthy, which can be seen at the temperatures evaluated at the measuring point in \textcolor{blue}{Fig.~\protect\ref{FIG:TapeTemperatures}}. However, the temperatures in the nip point does not really drift apart due to the aforementioned effects.

\begin{figure}
	\centering
	\includegraphics[width=1\columnwidth]{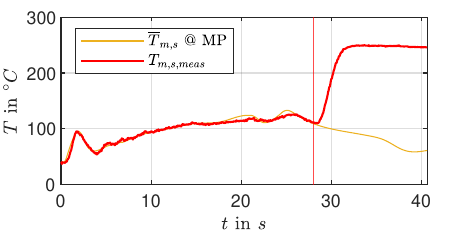}
	\caption{Simulated $\overline{T}_{m,s}$ and measured $T_{m,s,meas}$ temperature at measuring point on tape side facing the surroundings during test trajectory.}
	\label{FIG:ComparisonTempSimMeas}
\end{figure}

\begin{figure}
	\centering
	\includegraphics[width=1\columnwidth]{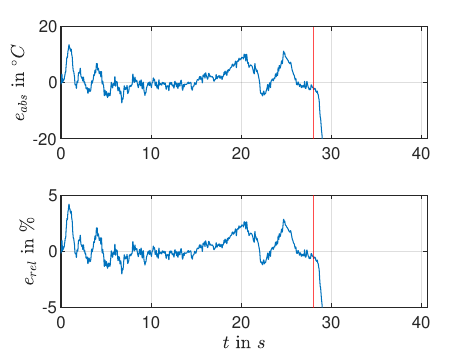}
	\caption{Absolute and relative error between simulated and measured temperature at specific measuring point.}
	\label{FIG:SimError}
\end{figure}

\begin{figure}
	\centering
	\includegraphics[width=1\columnwidth]{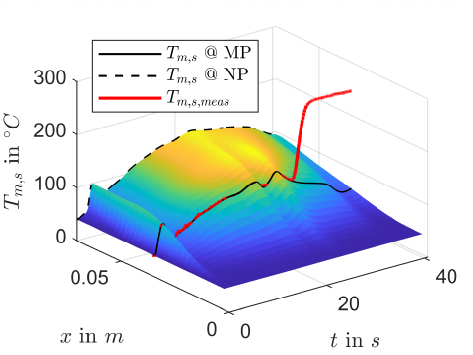}
	\caption{Spatio-temporal temperature distribution along the tape domain with highlighted profiles $T_{m,s}$ at measuring point (MP) and nip point (NP), including experimental measurement data $T_{m,s,meas}$ at the MP for validation.}
	\label{FIG:TapeTempSurf}
\end{figure}

\textcolor{blue}{Fig.~\protect\ref{FIG:ComparisonTempSimMeas}} compares the simulated tape temperature $T_{m,s}$ on the surroundings side with the corresponding reference temperature $T_{m,s,meas}$, which is represented by the red line. As the infrared temperature sensor spot captures more than one segment, depending on the size of the spatial discretization steps, the mean value of the temperatures $\overline{T}_{m,s}$ of the affected segments was used as comparative value. The high degree of filtering used to mitigate noise from the robot's drive system leads to a damping of the signal. This is most critical in high-gradient thermal zones, where the sensor output fails to represent the dynamic temperature spikes accurately, which can also be seen at the deviation to the measurement $T_{m,s,meas}$ in the first \qty{5}{\unit{\second}} or between second 18 to 28, shown by the absolute ($e_{abs} = \overline{T}_{m,s} - T_{m,s,meas}$) and relative error ($e_{rel} = (\overline{T}_{m,s} - T_{m,s,meas})/T_{m,s,meas}$) in \textcolor{blue}{Fig.~\protect\ref{FIG:SimError}}. The vertical red line in \textcolor{blue}{Fig.~\protect\ref{FIG:ComparisonTempSimMeas}} and \textcolor{blue}{Fig.~\protect\ref{FIG:SimError}} indicates the time when the tape expires the sensor area after cutting at the end of the process. The error immediately increases because now the sensor is pointing at one of the infrared heaters in the background, which can be comprehended by looking at \textcolor{blue}{Fig.~\protect\ref{FIG:ModelingScheme}}. The total simulation result of the temperature field of the tape's surface that is aligned to the sensor, including the reference measurement, is depicted in \textcolor{blue}{Fig.~\protect\ref{FIG:TapeTempSurf}}.
As can be seen, with this approach, quite exact predictions of the temperature at the points of interest are possible, while simultaneously capturing the dynamic thermal transitions of the tape at high heating rates of the emitters during the inrush current phase. The maximum absolute and relative errors with respect to the $N_{t}$ temporal measurements as well as the root mean squared error (RMSE)
\begin{equation}
	RMSE = \sqrt{\frac{1}{N_{t}} \sum_{j = 1}^{N_{t}} (\overline{T}_{m,s} - T_{m,s,meas})^2}
\end{equation}
and normalized root mean squared error (NRMSE)
\begin{equation}
	NRMSE = \frac{RMSE}{\overline{T}_{m,s,meas}} \cdot 100
\end{equation}
with $\overline{T}_{m,s,meas}$ being the temporal arithmetic mean value of the measured data are summarized in \textcolor{blue}{Table~\protect\ref{TAB:ErrorEvaluation}}. The results demonstrate that the model is highly consistent with the measurements at the monitoring point, especially in capturing the transient response during significant variations in tape velocity or strongly fluctuating current modulations. Furthermore, the pronounced nonlinearities within the radiation model were successfully replicated without compromising the stability of the system. Nevertheless, to reach this high-fidelity representation, the remaining uncertain parameters required systematic tuning to bridge the gap between numerical simulation and physical reality. \textcolor{blue}{Section~\protect\ref{SEC:S_ParameterTuning}} in the supplementary material addresses the tuning process to reach the accuracy presented in this paper. However, it should be noted that the reported model performance is evaluated using the same dynamic trajectory and measurement point employed for tuning. Therefore, a comprehensive validation using independent experimental datasets under varying operating conditions is the subject of future work.

\begin{table}[width=.9\linewidth,cols=4,pos=h]
	\caption{Error evaluation compared to measured temperature on tape at surface facing the surroundings.}
	\label{TAB:ErrorEvaluation}
	\begin{tabular*}{\tblwidth}{@{} LLLL@{} }
		\toprule
		Metric & Value & Unit \\
		\midrule
		$max(e_{abs})$ & 13.52 & \unit{\celsius} \\
		$max(e_{rel})$ & 4.19 & \unit{\percent} \\
		$RMSE$ & 4.02 & \unit{\celsius} \\
		$NRMSE$ & 1.08 & \unit{\percent} \\
		\bottomrule
	\end{tabular*}
\end{table}

\begin{figure}
	\centering
	\includegraphics[width=1\columnwidth]{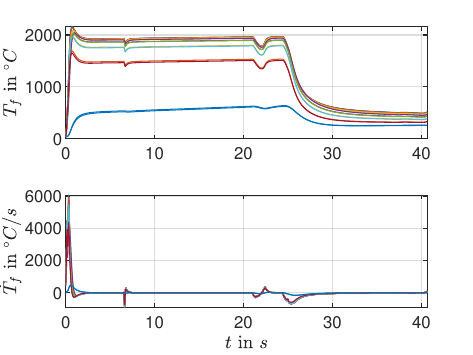}
	\caption{Filament temperature simulation $T_{f,k}$ and its derivative $\dot{T}_{f,k}$ at each segment $k$.}
	\label{FIG:FilamentTemp}
\end{figure}

\begin{figure}
	\centering
	\includegraphics[width=1\columnwidth]{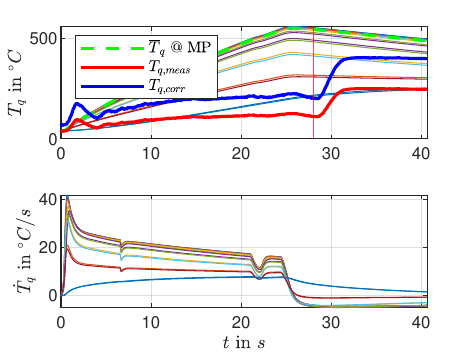}
	\caption{Quartz glass temperature simulation $T_{q,k}$ and its derivative $\dot{T}_{q,k}$ at each segment $k$. MP = Monitoring Point.}
	\label{FIG:QuartzTemp}
\end{figure}

The simulation results of the filament temperature $T_{f,k}$ and the quartz glass temperature $T_{q,k}$, as well as their derivatives $\dot{T}_{f,k}$ and $\dot{T}_{q,k}$ evaluated at all $N_{h}$ discretized segments, are shown in \textcolor{blue}{Fig.~\protect\ref{FIG:FilamentTemp}} and \textcolor{blue}{Fig.~\protect\ref{FIG:QuartzTemp}}. It can be seen that the inrush current significantly influences the heat generation in the different segments, which yields high heating rates especially in the area of the free filament coil in the heater center. The model addresses the early heat concentration in these segments because there are fewer heat losses to the contact wires or guiding plates than at the segments directly neighboring them. Furthermore, the electric resistance of tungsten increases with temperature, again leading to higher heat generation according to \textcolor{blue}{Eq.~\protect\ref{EQ:ElectricPower}}. This leads to a high heating rate within the first seconds, allowing for the reproduction of the measured heat peak at early stages that appears in \textcolor{blue}{Fig.~\protect\ref{FIG:TapeTemperatures}} and \textcolor{blue}{Fig.~\protect\ref{FIG:ComparisonTempSimMeas}}. Without this segmentation, the entire mass of the filament coil would be treated as a single thermal entity, significantly increasing thermal inertia and artificially  slowing down the initial heating rate. Consequently, a simplified, non-segmented emitter model would fail to reproduce the internal dynamics, resulting in the loss of the characteristic temperature peak observed within the first \qty{5}{\unit{\second}}.

Another interesting observation can be made in \textcolor{blue}{Fig.~\protect\ref{FIG:QuartzTemp}} by comparing the measured temperature with the quartz glass temperature simulation when the tape expires the measuring point after cutting. Behind the red vertical line, the sensor now points on one of the heaters in the background and therefore shows $T_{q,meas}$. Corrected for the emissivity of the quartz glass against the used sensor emissivity, similar to \textcolor{blue}{Eq.~\protect\ref{EQ:S_EmissivityCorrection}}, the measured and corrected temperature $T_{q,corr}$ is qualitative in good agreement with the segments of the quartz glass envelope that are detected by the sensor spot, which further validates the used heat source model. As the heaters are not symmetrically aligned with the tape strip, the measurement correlates not with the temperature at the center segment, but with those captured by the area of the sensor spot in close neighborhood. The mean value of the temperatures from the affected segments are represented by $\overline{T}_{q}$ and used as comparative value. The discrepancy between the measured and simulated temperature is primarily due to the spectral characteristics of the measurement. Although quartz glass is generally opaque in the \qty{8}{\mu\meter} - \qty{14}{\mu\meter} sensor range, the intense thermal radiation from the internal filament still influences the measurement through significant temperature gradients within the glass and secondary heating effects. Furthermore, the simulation does not account for ambient reflections on the glass surface or the angle-dependent emissivity of the quartz tube. A precise determination of a combined effective emissivity was omitted due to the extreme complexity of modeling these overlapping thermal and optical effects.

\begin{figure*}
	\centering
	\includegraphics[width=0.9\textwidth]{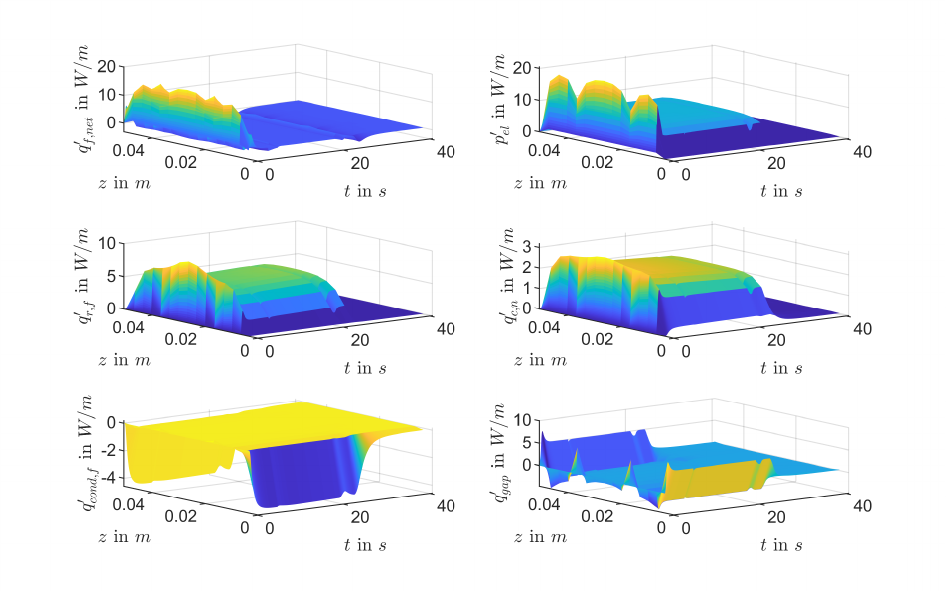}
	\caption{Spatial distribution of the line-source power and heat transfer rates along the axial position $z$ of the infrared heater at different time steps according to \textcolor{blue}{Eq.~\protect\ref{EQ:EnergyBalanceFilamentContinuum1D}}. The first subplot displays the sum of all terms, which is the net linear heat rate $q_{f,net}'$.}
	\label{FIG:ThermalHeatFlow}
\end{figure*}

Additionally, the different line-source power and heat transfer rates inside the filament coil described in \textcolor{blue}{Eq.~\protect\ref{EQ:EnergyBalanceFilamentContinuum1D}}, as well as the total line-source net heat rate $q_{net,f}'$ are presented in \textcolor{blue}{Fig.~\protect\ref{FIG:ThermalHeatFlow}}. Again, the concentration of the generated heat can be assigned to specific segments during the inrush current phase. Thereafter, in a quasi stationary state of the heater, the heat terms equalize increasingly.

The computational efficiency was evaluated on a workstation equipped with an Intel Core i7-13620H CPU and 16 GB of RAM. Utilizing the CasADi framework (v3.7.2), the aforementioned simulation required a total execution time of \qty{347}{\unit{\second}} using $N_{x} = 30$ spatial discretization steps for the tape in the respective domain and $N_{h} = 15$ spatial steps for the infrared heaters. To verify spatial grid independence, two refined configurations were evaluated:
\begin{itemize}
	\item Refined tape grid ($N_{x} = 60, N_{h} = 15$): The temperature at the measurement point differs by only \qty{0.17}{\percent} NRMSE (\qty{0.61}{\kelvin} RMSE), confirming full grid independence regarding the tape discretization.
	\item Refined heater grid ($N_{x} = 30, N_{h} = 31$): This configuration yields a difference of \qty{1.86}{\percent} NRMSE (\qty{6.93}{\kelvin} RMSE).
\end{itemize}
Although doubling the heater resolution introduces an absolute shift in the values, the overall variation remains under \qty{2}{\percent} NRMSE and does not alter the qualitative thermal trends. However, this refinement increases the simulation time by approximately \qty{148}{\percent} to \qty{862}{\unit{\second}}. Therefore, the baseline configuration ($N_{x} = 30, N_{h} = 15$) was maintained as a trade-off between sufficient macro-level accuracy and computational efficiency.

\section{Conclusions}
\label{SEC:Conclusion}
In this paper a thermo-radiative model for automated tape laying (ATL) was established and validated. Through experimental tests with synchronized measurements of end-effector velocity, ATL supply current, and surface temperature, it is demonstrated that simulations driven by real-world data accurately reproduce transient thermal responses. The model captures both the sharp heating onset at the start of motion and the subsequent transition to steady-state conditions. Within the margins of sensor noise, material property variability, and modeling simplifications, the simulated temperatures closely align with actual surface measurements. Future integration of sensing in the post-nip region and on the mold surface would further validate model fidelity across the entire bond interface.

The key modeling contributions include heater segmentation, which captures internal, highly dynamic thermal behavior and significantly improves the description of transient states, thereby reducing defect risk and improving part quality. The redefined view factor modeling prevents the overestimation of net incident radiative flux on the tape surface.
Moreover, a local convection assessment is implemented, accounting for boundary layer development and the transition to mixed convection as characterized by the Richardson number. The introduced two-layer tape model resolves phase-shifted temperatures on opposing surfaces, enabling an accurate, side-specific comparison to measured data. The high fidelity of the integrated process model is demonstrated by a maximum deviation of only \qty{1.08}{\percent} (NRMSE) compared to industrial ATL measurements.

Possible limitations may occur due to the assumption of a diffuse-gray enclosure and lumped optical properties that underrepresent band-specific effects during significant temperature variations. Real-time execution was not targeted, because current runtimes reflect high-fidelity coupling rather than control-loop constraints like cycle times. Uncertainties remain in emissivity and convection coefficients due to limited measurement equipment for accurately determining corresponding values.

Future investigations will leverage this validated model to enable trajectory optimization and AI-assisted process design. The use in constrained trajectory optimization enables the computation of robot paths and speed profiles that exploit the available heating power while respecting constraints like robot kinematics and process limits. Thereby, joint, velocity and acceleration bounds and the reachable workspace are considered as well as maximum allowable surface temperatures, desired target nip-point temperature bands and thermal gradients to avoid defects. Another aspect would be the AI-assisted planning and learning, where surrogate models get trained on simulation or experimental data for fast predictions in optimization and control. Additionally, physics-informed machine learning can be used to retain extrapolation by encoding energy balance and the nonlinear radiation structure. To further improve the process quality, a parameter estimation can be introduced to minimize remaining deviations to a desired behavior. Extended sensing like at post-nip or mold-surface locations can be used to validate the model at different stages.

\printcredits


\section*{Data availability}
The authors confirm that the data supporting the findings of this study are entirely available within the article and its supplementary materials.

\section*{Acknowledgments}
This work has been supported by the ”LCM - K2 Center for Symbiotic Mechatronics” within the framework of the Austrian COMET-K2 program.

\bibliographystyle{elsarticle-num}

\bibliography{Bib_ATLProcessModel}

\clearpage
\pagestyle{empty} 

\makeatletter
\define@key{Gin}{artifact}[]{}
\makeatother

\newcounter{SavedMainPages}
\setcounter{SavedMainPages}{\value{page}}
\addtocounter{SavedMainPages}{-1} 

\makeatletter
\renewcommand{\thepage}{\arabic{SavedMainPages}}
\makeatother



\section*{Supplementary material}

# Supplementary material: Transient multimode heat transfer of an industrial automated tape laying process under rapidly changing conditions

Bernhard Rameder[a,*], Hubert Gattringer[a], Andreas Müller[a] and Ronald Naderer[b]

[a] *Institute of Robotics, Johannes Kepler University Linz, Altenberger Straße 69, 4040 Linz, Austria*
[b] *FerRobotics Compliant Robot Technology GmbH, Altenberger Straße 66c, 4040 Linz, Austria*



ABSTRACT

This supplementary document provides comprehensive thermal and optical property datasets, extreme value analyses, detailed formulations for internal heat transport mechanisms, and calibration protocols supporting the numerical modeling and experimental investigations presented in the main manuscript "Transient multimode heat transfer of an industrial automated tape laying process under rapidly changing conditions". Specifically, this material details the material-specific characteristic curves essential for accurate thermal tracking of the automated tape laying (ATL) process. To define the operational boundaries, extreme value thresholds characterizing the underlying fluid flow regimes are systematically evaluated and mapped according to the appropriate dimensionless indicators. In addition, the explicit governing equations, local thermal conductances, and radiative network relations for the emitter components are provided. Furthermore, the temperature-dependent optical properties used in the enclosure radiation model are documented, alongside a rigorous description of the associated experimental measurement methods and sensor setups utilized for data acquisition. Thereafter, analytical view factor formulations are presented, forming the mathematical foundation for the radiative boundary conditions. Finally, the tuning of uncertain system parameters is elaborated, providing the foundational calibration steps required to achieve the predictive accuracy demonstrated in the primary research article.

## S1. Dimensionless numbers

To ensure a realistic representation of the thermal behavior within the presented model, accurate determination of the convective heat transfer coefficients ($h_{m,c}, h_{m,s}$) at the model boundaries is of high importance. Since these coefficients strongly depend on the local flow morphology and the prevailing transport mechanisms, they are determined using analytical or empirical Nusselt number correlations ($Nu$) stated in the main manuscript. To select the thermodynamically appropriate Nusselt correlations for the specific application at hand, the flow regime must be uniquely characterized beforehand. Methodologically, this classification is achieved through a coupled system of dimensionless numbers.

The Reynolds number

$$Re_x = \frac{Ux}{\nu} \tag{S1}$$

primarily serves to identify the hydrodynamic flow regime (laminar vs. turbulent) under the influence of forced convection. The fluid-specific Prandtl number

$$Pr = \frac{\nu}{\alpha} = \frac{\nu\rho c_p}{k} = \frac{\mu c_p}{k} \tag{S2}$$

links momentum transport to thermal transport, thereby describing the ratio of velocity boundary layer thickness to thermal boundary layer thickness. If temperature-induced density gradients occur alongside the forced flow, the ratio

*Corresponding author
bernhard.rameder@jku.at (B. Rameder)
https://www.jku.at/en/institute-of-robotics/ (B. Rameder)
ORCID(s): 0000-0002-6792-6129 (B. Rameder); 0000-0002-8846-9051 (H. Gattringer); 0000-0001-5033-340X (A. Müller)

of free to forced convection is mathematically evaluated via the Richardson number

$$Ri_x = \frac{Gr_x}{Re_x^2}, \tag{S3}$$

while the intensity of this buoyancy-driven, free convection is quantified by the Grashof number

$$Gr_x = \frac{g\beta(T_s - T_\infty)x^3}{\nu^2}, \tag{S4}$$

which represents the ratio of buoyancy forces to viscous forces within the fluid. Finally, the resulting convective heat transport capability is scaled by the Rayleigh number

$$Ra_x = Gr_x Pr. \tag{S5}$$

The systematic calculation and comparison of these five dimensionless numbers enables a well-founded definition of the flow characteristics. This builds the mathematical foundation required to select the appropriate Nusselt equations, thereby ensuring an accurate transfer of heat phenomena into the numerical simulation.

The approximations of the material properties

$$\rho_{air}(T_{film}) = \frac{p}{R_s \cdot T_{film}}, \tag{S6}$$

$$\nu_{air}(T_{film}) \approx \nu_0 \left(\frac{T_{film}}{T_0}\right)^{1.7}, \tag{S7}$$

$$\mu_{air}(T_{film}) = \nu_{air} \cdot \rho_{air}, \tag{S8}$$

$$\alpha_{air}(T_{film}) = \frac{k_{air}}{\rho_{air} \cdot c_{p,air}} \text{ and} \tag{S9}$$

$$\beta_{air}(T_{film}) \approx \frac{1}{T_{film}}, \tag{S10}$$

evaluated at film temperature

$$T_{film} = \frac{T_s + T_\infty}{2}, \tag{S11}$$

are the density $\rho_{air}$, kinematic $\nu_{air}$ [S1] and dynamic viscosity $\mu_{air}$, thermal diffusivity $\alpha_{air}$ and coefficient of thermal expansion $\beta_{air}$ of the surrounding air. $U$ is the relative velocity between the translating material and the surrounding fluid, which is $\upsilon_T$ in this approach, and $g$ stands for gravity. Eq. S6 is the ideal gas equation and holds the specific gas constant for air $R_s = 287.1\,\mathrm{J\,kg^{-1}\,K^{-1}}$ as well as the standard pressure $p = 101\,325\,\mathrm{Pa}$. With respect to the expected temperature range of the air around the tape, the specific heat $c_{p,air} = 1006\,\mathrm{J\,kg^{-1}\,K^{-1}}$ can be seen as constant in a range between 273.15 K and 473.15 K (0 °C and 200 °C), whereas the thermal conductivity can be approximated with

$$k_{air}(T_{film}) = k_{air,0} \left(\frac{T}{T_0}\right)^{3/2} \frac{T_0 + S_\lambda}{T + S_\lambda} \tag{S12}$$

following the suggestion in [S2], where $S_\lambda \approx 194\,\mathrm{K}$ can be used for air. $T_s$ and $T_\infty$ are the general definitions of the actual surface segment temperature at the considered material side and the corresponding ambient temperature, respectively. The kinematic viscosity $\nu_0 = 1.51 \times 10^{-5}\,\mathrm{m^2\,s^{-1}}$ and the thermal conductivity $k_{air,0} = 0.026\,\mathrm{W\,m^{-1}\,K^{-1}}$ are defined at room temperature $T_0 = 293.15\,\mathrm{K}$ (20 °C).

## S2. Fluid flow regime analysis and extreme values

To ensure the thermodynamic consistency of the boundaries applied in the primary thermal model, an analysis of the fluid flow regimes surrounding the ATL process zones is required. In this context, identifying the convective

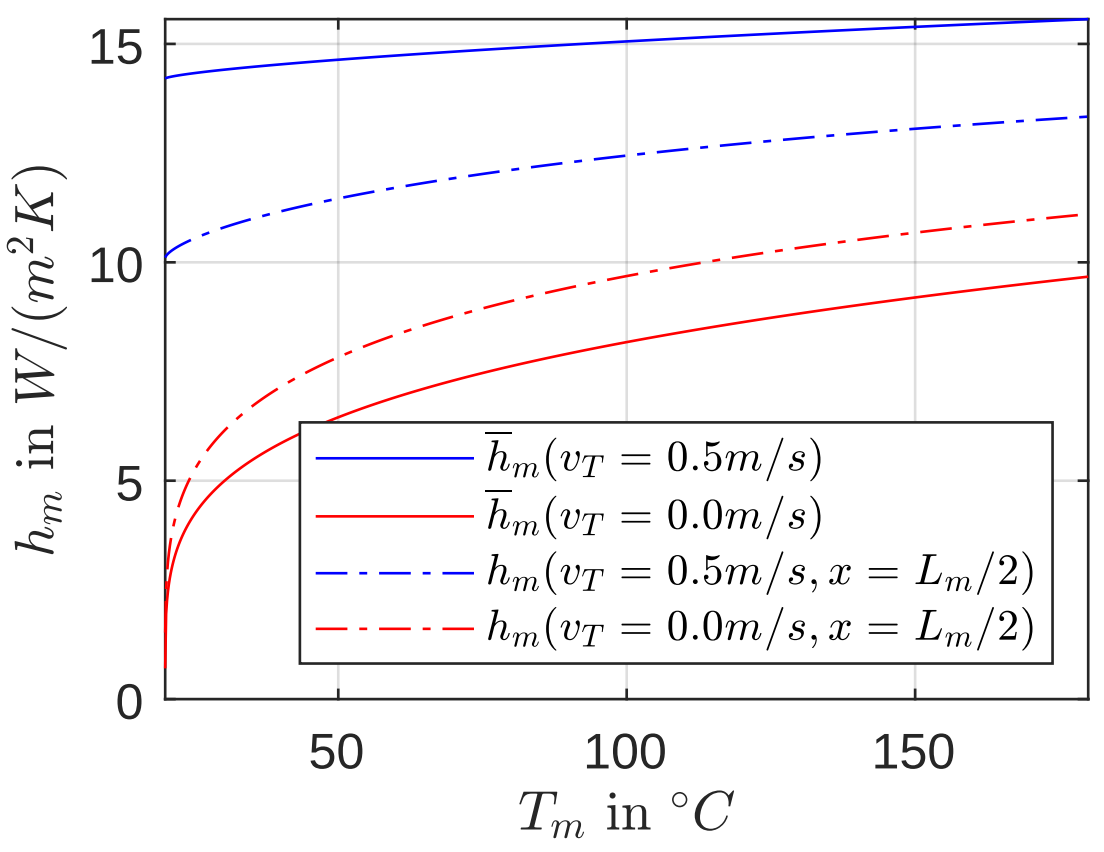


**Figure S1:** Temperature dependency of the averaged (full lines) and local (dash-dotted lines) mixed-convection coefficient $h_m$ at different velocities evaluated at the center point $x = L_m/2$ of the spatial domain.

heat transfer behavior is essential to justify the use of specific Nusselt number correlations under transient processing conditions. Table S1 compiles the geometric parameter ($x_{max}$), fluid properties ($T_\infty$) and assumed extreme operational values ($v_{T,max}$, $T_{s,max}$) used to compute the worst-case flow regimes. These values serve to characterize the convective boundaries and systematically verify that the flow remains within predictable regimes. Specifically, the data are used to confirm the Reynolds number ($Re$) for forced convection does not exceed the critical threshold required to maintain laminar stability, and that the Rayleigh number ($Ra$) for natural convection remains below the limit for the onset of turbulent buoyant flow. Consequently, the upper bounds for the corresponding Nusselt numbers ($Nu$) are established, guaranteeing conservative yet physically sound heat transfer coefficients throughout large temperature ranges.

**Table S1**
Maximum parameters for flow regime identification.

| Parameter | Max. value | Unit |
|---|---|---|
| $v_{T,max}$ | 0.2 | $\mathrm{m\,s^{-1}}$ |
| $T_{s,max}$ | 423.15 (150) | K (°C) |
| $T_\infty$ | 293.15 (20) | K (°C) |
| $x_{max} = L_m$ | 0.07 | m |

To demonstrate the necessity of the implemented mixed-convection model over a simplified constant convection coefficient approach, a parametric analysis was conducted across three distinct physical domains.

*Temperature dependency:* Evaluating fluid properties at a single fixed temperature is insufficient. As shown in Fig. S1, natural convection ($v_T = 0\,\mathrm{m\,s^{-1}}$) couples non-linearly with temperature. The model accurately captures the competitive interplay where rising temperatures enhance buoyancy forces while simultaneously increasing air viscosity. This creates a self-regulating thermal damping effect that a rigid, constant convection coefficient model completely ignores.

*Position dependency:* Assuming a spatially uniform coefficient misrepresents boundary layer physics. As shown in Fig. S2, the local coefficient $h_m$ spikes drastically at the leading edge ($x \rightarrow 0$) because the thermal boundary layer thickness is zero, and then decays steeply along the path. For a low-conductivity material like HDPE, heat cannot easily redistribute internally. Relying on an integrated average ($\bar{h}_m$) would mask this severe spatial thermal shock, which is important for predicting localized polymer degradation and entry-zone warping.

*Velocity dependency:* A pure forced-convection assumption fails to capture the fluid regime transitions. As shown in Fig. S3, the non-zero baseline at $v_T = 0\,\mathrm{m\,s^{-1}}$ proves that buoyancy-driven natural convection dominates at low processing speeds, while forced Sakiadis flow takes over at higher velocities. Omitting natural convection or utilizing

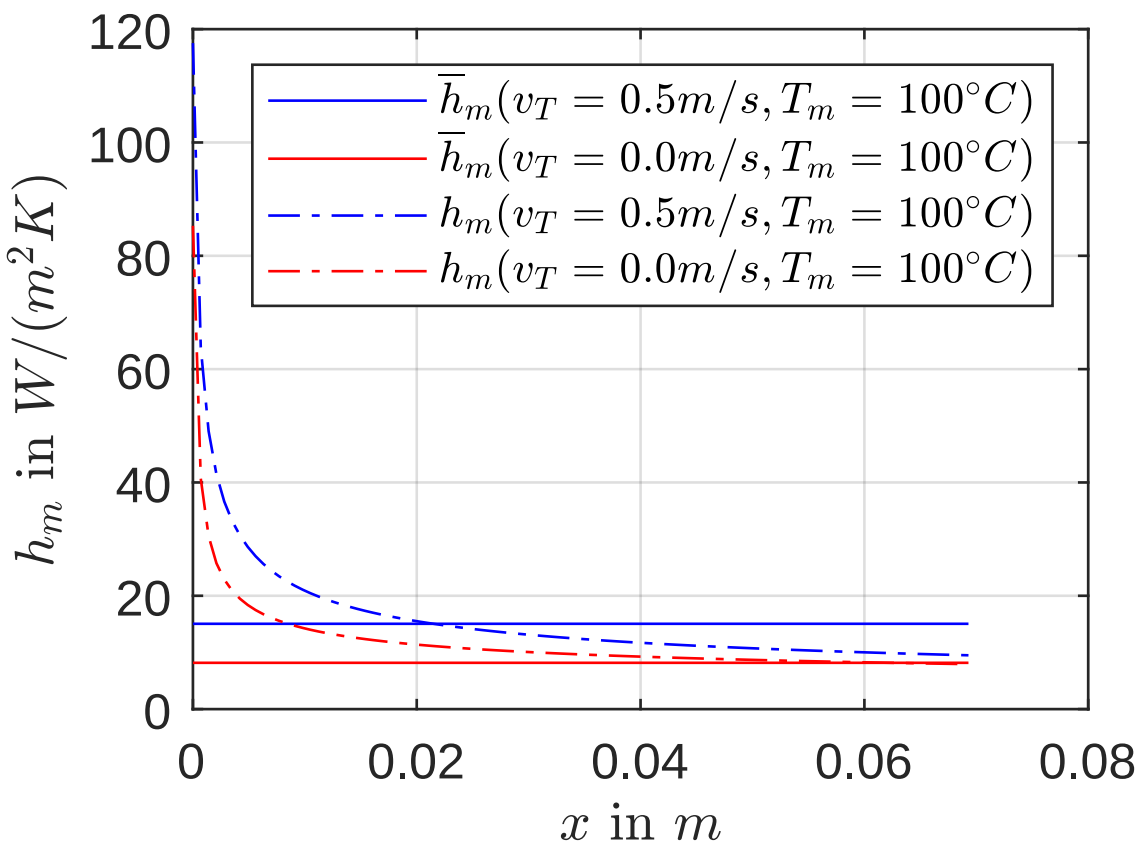


**Figure S2:** Position dependency of the averaged (full lines) and local (dash-dotted lines) mixed-convection coefficient $h_m$ at different velocities evaluated at an exemplary temperature $T_m = 100°C$.

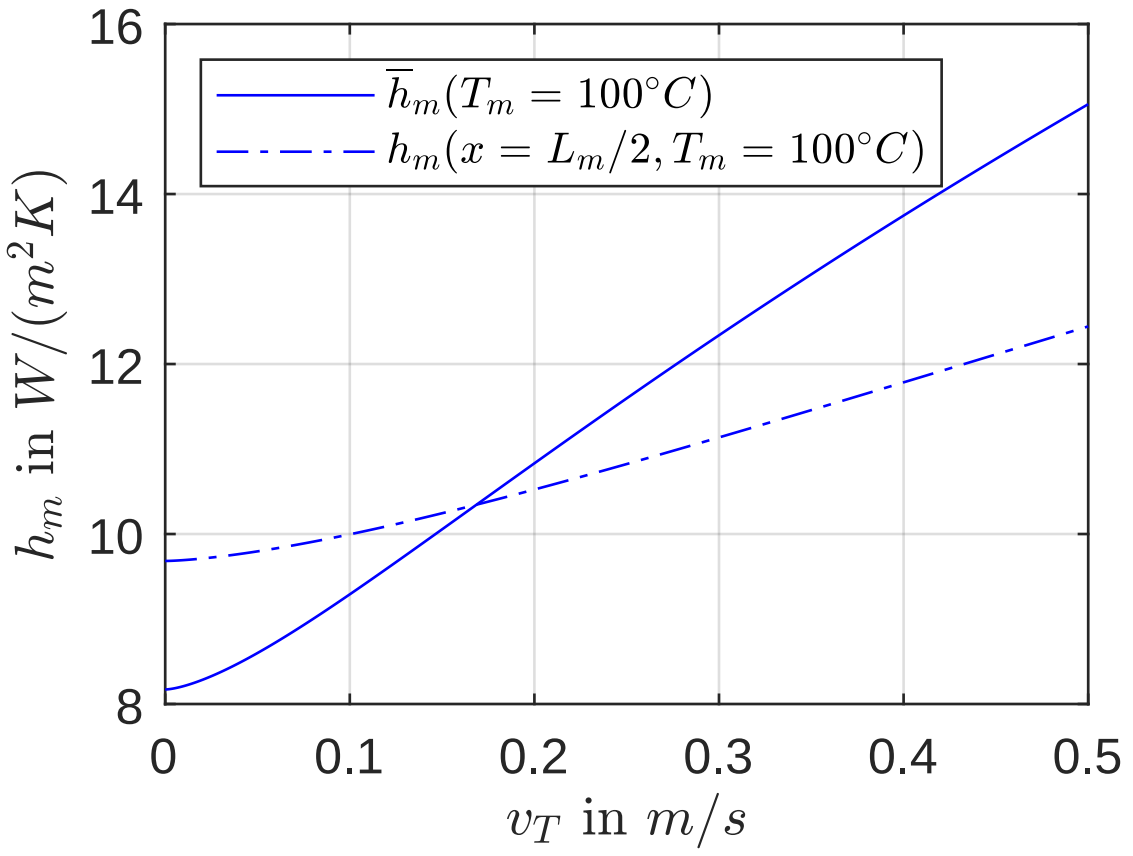


**Figure S3:** Velocity dependency of the averaged (full line) and local (dash-dotted line) mixed-convection coefficient $h_m$ at the center point $x = L_m/2$ of the spatial domain for an exemplary temperature $T_m = 100°C$.

a constant forced-convection coefficient would completely neglect thermal losses during stationary process phases, standstill, or machine ramp-up.

To analyze the long-term coupling behavior of the models, a transient simulation was performed comparing the fully resolved local mixed-convection model against a simplified constant forced-convection framework (evaluated at a reference operating point of $v_{T,op} = 0.2\,\mathrm{m\,s^{-1}}$ and $T_{m,op} = 150\,°\mathrm{C}$), as illustrated in Fig. S4. An initial tape temperature of approximately 100 °C was deliberately chosen for this analysis to actively uncover and amplify the boundary layer phenomena, which were suppressed under the initial 40 °C baseline setup due to the following physical mechanisms.

*Inlet temperature effect:* Although the local Sakiadis convection coefficient mathematically spikes at the leading edge ($x \to 0$), the resulting heat flux ($q''_{conv,m,(c,s)} = h_{m,(c,s)}(T_{m,(c,s)} - T_{\infty,(c,s)})$) remains near zero if the material enters near ambient temperature (20 °C). Raising the inlet temperature, like it is the case for tape preheating, removes this thermal masking and reveals the true impact of the entry-zone cooling.

*Temperature and material impact:* Convective losses scale heavily with the temperature difference to the surroundings. While this effect is subtle for low-temperature applications, it becomes highly critical for high-temperature polymers such as PEEK. The high-temperature simulation clearly demonstrates that the choice of the convection model leads to substantial deviations in the predicted core temperatures over time.

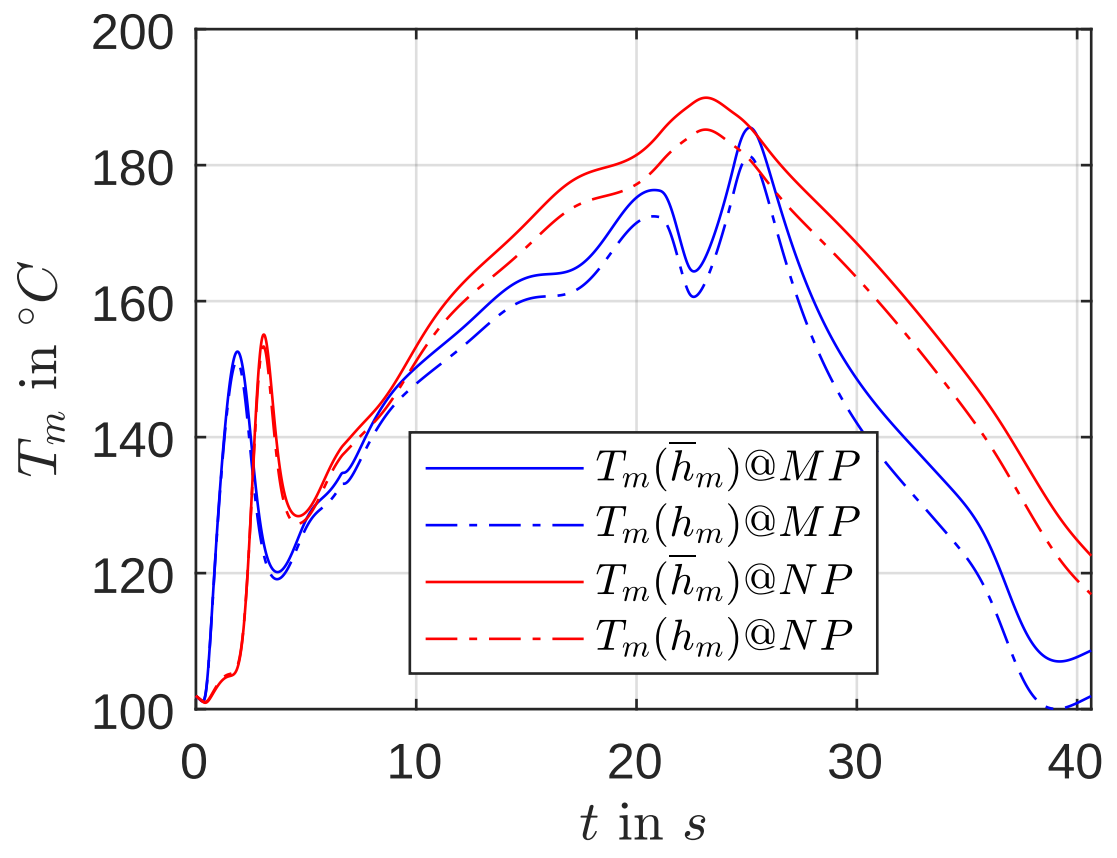


**Figure S4:** Transient convection assessment analysis comparing the fully coupled local mixed model against a constant forced-convection framework. The curves are evaluated at the measurement point (MP) and the nip point (NP).

*Regime transition during deceleration and cooling:* As the process velocity drops towards the end of the trajectory and the tape undergoes transient cooling caused by a decreased heater current, the natural convection component becomes increasingly dominant. This shifts the local Richardson number, leading to an amplified divergence between the local mixed model and the static, forced-only baseline.

Due to the highly coupled nature of the transient process, where velocity, local position, and surface temperatures change simultaneously, the individual contributions of each isolated factor cannot be extracted from this final temperature plot alone. This comprehensively justifies the necessity of the isolated parameter studies provided above (Fig. S1, Fig. S2, Fig. S3). The transient plot in Fig. S4 proves that utilizing a simplified, constant forced-convection coefficient results in a significant underestimation of local heat losses. Consequently, the simplified model predicts falsely elevated tape temperatures (solid lines), whereas the local mixed-convection model (dash-dotted lines) correctly captures the cumulative thermal losses, validating its necessity for accurate process window predictions.

## S3. Detailed formulations for filament and envelope heat transport mechanisms

This section provides the detailed mathematical formulations, auxiliary heat-balance terms, and geometric parameters for the heat transport mechanisms within the emitter. Consistent with the simplified representations in the main text, the conductive and radiative transport equations for both the tungsten filament coil and the quartz glass envelope are explicitly defined below.

To determine the conductive heat $Q_{cond,f}$ along the coiled wire, the thermal conductance

$$G_f = \frac{k_f a_f}{l_f} \tag{S13}$$

of each segment is first calculated, where $k_f$ is the thermal conductivity of tungsten, $a_f$ is the cross-sectional area of the wire, and $l_f$ is the unwound length in the considered segment. This allows for the computation of the conductive heat flow

$$Q_{cond,f} = G_f \Delta T_f \tag{S14}$$

between adjacent filament coil segments. Additional heat loss terms to the thicker contact wires at the start ($k = 1$) and the end ($k = N_h$) of the filament coil lead to

$$Q_{cond,mount} = G_{mount}(T_{f,k} - T_{q,k}) \tag{S15}$$

with the corresponding thermal conductance

$$G_{mount} = \frac{k_f a_{mount}}{l_{mount}}. \tag{S16}$$

Again, $a_{mount}$ and $l_{mount}$ represent the mounting wire's cross-sectional area and length, respectively.

The net heat flux across the gap is the sum of gas conduction through the neon fill and surface-to-surface radiation between adjacent windings

$$Q_{gap} = Q_{cond,gap} + Q_{rad,gap}. \tag{S17}$$

Here,

$$Q_{cond,gap} = G_{cond,gap} \Delta T_f \tag{S18}$$

with thermal conductance

$$G_{cond,gap} = \frac{k_n(T_{gap}) a_g}{w_g} \tag{S19}$$

represents the conductive transport through the neon in the gap, and

$$Q_{rad,gap} = \sigma \epsilon_{gap} a_g (T_{f,k}^4 - T_{f,k+1}^4) \tag{S20}$$

captures the radiative heat exchange between facing filament surfaces. The temperature-dependent thermal conductivity $k_n$ according to Eq. S29 used in Eq. S19 is evaluated at the corresponding approximated mean gap temperature

$$T_{gap} = \frac{T_{f,k} + T_{f,k+1}}{2} \tag{S21}$$

and the effective segment exchange area $a_g = d_f \cdot l_f / N_h$ for a given gap width $w_g$. The radiative heat exchange between adjacent coil windings is modeled by considering them as gray, diffuse surfaces in an open enclosure. To account for both the geometric configuration and the surface properties, the net radiation exchange is determined using the network resistance model [S3]. The effective emissivity, which characterizes the radiant interaction between two windings, is defined as

$$\epsilon_{gap} = \frac{1}{\frac{1}{\epsilon_{f,k}} + \frac{1}{F_{k,k+1}} \left( \frac{1}{\epsilon_{f,k+1}} - 1 \right)}, \tag{S22}$$

where

$$F_{k,k+1} = \sqrt{1 + \left( \frac{w_g}{d_f} \right)^2} - \frac{w_g}{d_f} \tag{S23}$$

represents the geometric view factor between the windings [S4], and $\epsilon_{f,k}, \epsilon_{f,k+1}$ denote the respective surface emissivities. This formulation, based on the Oppenheim surface and space resistance network, ensures that multiple reflections between the curved surfaces are accurately captured, even when the view factor is less than unity due to the spacing of the coils.

Consistent with the approach used for the tungsten coil, heat conduction along the glass cylinder is modeled using the thermal conductance

$$G_q = \frac{k_q a_q}{l_q} \tag{S24}$$

and the resulting heat flow

$$Q_{cond,q} = G_q \Delta T_q \tag{S25}$$

along the corresponding axis, with $k_q$ being the thermal conductivity of quartz glass, $a_q$ the cross-sectional area of the hollow cylinder, and $l_q$ the corresponding segment length.

## S4. Material properties and optical characterization

The accuracy of the multimode thermal modeling framework relies heavily on the precision of the underlying thermo-physical material properties. Because the ATL process operates across broad thermal windows, assuming constant material parameters would lead to significant modeling errors. This section summarizes the core material datasets and experimental methodologies used to feed the numerical solver. Specifically, it encompasses both the temperature-dependent bulk properties of the system components and the specialized optical characteristics governing the radiative heat exchange.

### S4.1. Temperature-dependent material parameters

To capture the transient behavior of the radiating infrared emitters and the surrounding domain, the temperature-dependent variations of the constituent materials must be accounted for.

For the heating elements, the specific heat capacity $c_{p,f}$ and the electrical resistivity $r_t$ of tungsten are mapped over the full operational temperature range to precisely model the electrical inrush currents and thermal inertia. According the suggestion in [S5], obtaining the corresponding data from [S6, S3, S7], the temperature dependent specific heat of tungsten used in the energy balance of the tungsten filament in the main document for the range of $298\,\mathrm{K} < T_f < 1900\,\mathrm{K}$ is given with

$$c_{p,f}(T_f) = \frac{1}{183.84\times 10^{-3}}\left[23.9593{+}2.639\,68\frac{T_f}{10^3}{+}1.257\,75\left(\frac{T_f}{10^3}\right)^2{-}0.253\,64\left(\frac{T_f}{10^3}\right)^3{-}(0.048\,407)/\left(\frac{T_f}{10^3}\right)^2\right] \tag{S26}$$

and for temperatures between $1900\,\mathrm{K} < T_f < 3680\,\mathrm{K}$ by

$$c_{p,f}(T_f) = \frac{1}{183.84\times 10^{-3}}\left[-22.5764{+}90.2798\frac{T_f}{10^3}{-}44.2715\left(\frac{T_f}{10^3}\right)^2{+}7.176\,63\left(\frac{T_f}{10^3}\right)^3{-}(24.0974)/\left(\frac{T_f}{10^3}\right)^2\right]. \tag{S27}$$

The resistivity was also adopted from [S5, S6] and reads

$$r_t(T_f) = -5.0134\times 10^{-9} + 7.307\,075\,9\times 10^{-11} T_f^{1.1796583} \tag{S28}$$

for a temperature range $300\,\mathrm{K} \le T_f \le 3655\,\mathrm{K}$. Furthermore, to evaluate the heat dissipation within the emitter envelope, the temperature-dependent thermal conductivity

$$k_n(T_n) = -3.601\,802\,1\times 10^{-3} + 1.429\,062\,3\times 10^{-3} T_n^{0.63362378} \tag{S29}$$

of the neon fill gas in $\mathrm{W\,m^{-1}\,K^{-1}}$, as suggested in [S6] for the range $100\,\mathrm{K} \le T_n \le 2500\,\mathrm{K}$ is implemented. Lastly, the thermal behavior of the protective emitter casing is defined by the specific heat capacity

$$c_{p,q}(T_q) = \frac{\left[55.98 + 15.4\times 10^{-3} T_q - \frac{14.4\times 10^5}{T_q^2}\right]}{60.085\times 10^{-3}}. \tag{S30}$$

of the fused quartz glass, according to the authors of [S5] using the literature [S3, S8, S9].

### S4.2. Determination of optical properties

In addition to the thermal bulk properties, the radiative heat transfer within the enclosure is highly sensitive to the surface characteristics of the utilized materials. This subsection details the consolidation of these optical properties. While specific material datasets were determined via in-house experimental measurement methods and data acquisition protocols to capture temperature-dependent behavior, properties for other components were sourced from established literature including empirical or analytical methods like the Lorentz-Drude model. This hybrid approach ensures that the diffuse-gray enclosure model is backed by a combination of empirical measurements, literature data, and analytical fits, thereby minimizing uncertainties during high-temperature process phases.

*Tungsten:* The temperature-dependent emissivity of tungsten $\epsilon_f(T_f)$ is determined following the approach of Minissale et al. [S10], who validated a Fresnel-based model and two temperature-dependent Lorentz-Drude (TDLD) formulations against experimental data. While the Fresnel model effectively captures spectral reflectivity at discrete temperatures, the TDLD models provide a superior physical description by accounting for the temperature sensitivity of both intra- and interband electronic transitions across a broad range of 300 K to 3500 K. In this work, the emissivity values derived from the TDLD model within a spectral range of 0.6 µm to 2.8 µm were stored in a look-up table, allowing for efficient computation at the specific temperatures of interest.

*Gold:* The emissivity of gold used for the reflector surface $\epsilon_r(T_r)$ is suggested in different literature like [S11] to be approximately $\epsilon_r = 0.02$ for a temperature range between 373 K and 673 K in the near infrared range and handled in more detail for different wavelengths and temperatures in [S12, S13, S14, S15].

*Quartz glass:* To determine the emissivity $\epsilon_q(T_q)$ of the quartz glass of type "Heraeus Covantics HLQ® 200 series" used for the Heraeus Noblelight Duo Gold infrared heaters, the total transmissivity $\tau_{q,tot}$ was first obtained from the material's characteristic curves given in [S16]. Since specific data like the refractive index $n_\lambda \approx 1.45$ for the primary material was unavailable, the quartz material "Heraeus Covantics Infrasil® 300 series" addressed in [S17, S18] was used as a reference due to its similar optical properties.

First, the surface reflectivity $R$ is calculated using the Fresnel equation for normal incidence

$$R = \left(\frac{n_\lambda - 1}{n_\lambda + 1}\right)^2 . \tag{S31}$$

To account for the internal wavelength-dependent transmission $\tau_\lambda$ excluding surface losses, the total transmissivity $\tau_{\lambda,tot}$ is corrected as follows

$$\tau_\lambda = \frac{\tau_{\lambda,tot}}{1 - 2R}, \tag{S32}$$

which is a common first-order approximation to remove surface reflection losses from the total measured transmission. The spectral emissivity $\epsilon_\lambda$ is then calculated by considering multiple reflections between the surfaces. This leads to the standard analytical expression

$$\epsilon_\lambda = \frac{(1 - R)(1 - \tau_\lambda)}{1 - R\tau_\lambda} \tag{S33}$$

for hemispherical or normal emissivity of a plane-parallel plate, which serves as approximation for the tube walls by accounting for finite internal reflections [S19]. Finally, the effective emissivity $\epsilon_{eff}$ is determined. For practical calculation, the integral over the spectral range

$$\epsilon_{eff}(T) = \frac{\int_{\lambda_1}^{\lambda_2} \epsilon_\lambda(\lambda) E_{b,\lambda}(\lambda, T)\,\mathrm{d}\lambda}{\int_{\lambda_1}^{\lambda_2} E_{b,\lambda}(\lambda, T)\,\mathrm{d}\lambda} \tag{S34}$$

is simplified into a discrete sum over $N$ wavelength intervals

$$\epsilon_{eff}(T) = \frac{\sum_{i=1}^{N} \left(\epsilon_{\lambda,i}(\lambda) E_{b,\lambda}(\lambda_i, T)\Delta\lambda_i\right)}{\sum_{i=1}^{N} \left(E_{b,\lambda}(\lambda_i, T)\Delta\lambda_i\right)}, \tag{S35}$$

weighted by the spectral emissive power of a blackbody $E_{b,\lambda}$ defined in the main manuscript and evaluated at a temperature expected in the mean operating point. The effective transmissivity $\tau_{eff}$ for the defined spectral range is computed equally by replacing $\epsilon_{\lambda,i}(\lambda)$ with $\tau_{\lambda,i}(\lambda)$ in Eq. S35.

*Carbon-fiber reinforced HDPE tape:* Because no reliable literature data for the tape's total hemispherical emissivity $\epsilon_m(T_{m,c})$ (HDPE matrix with carbon-fiber reinforcement) were available, $\epsilon_m(T_{m,c})$ was determined experimentally. Two procedures that both require heating the tape to a steady, uniform temperature and measuring that temperature with

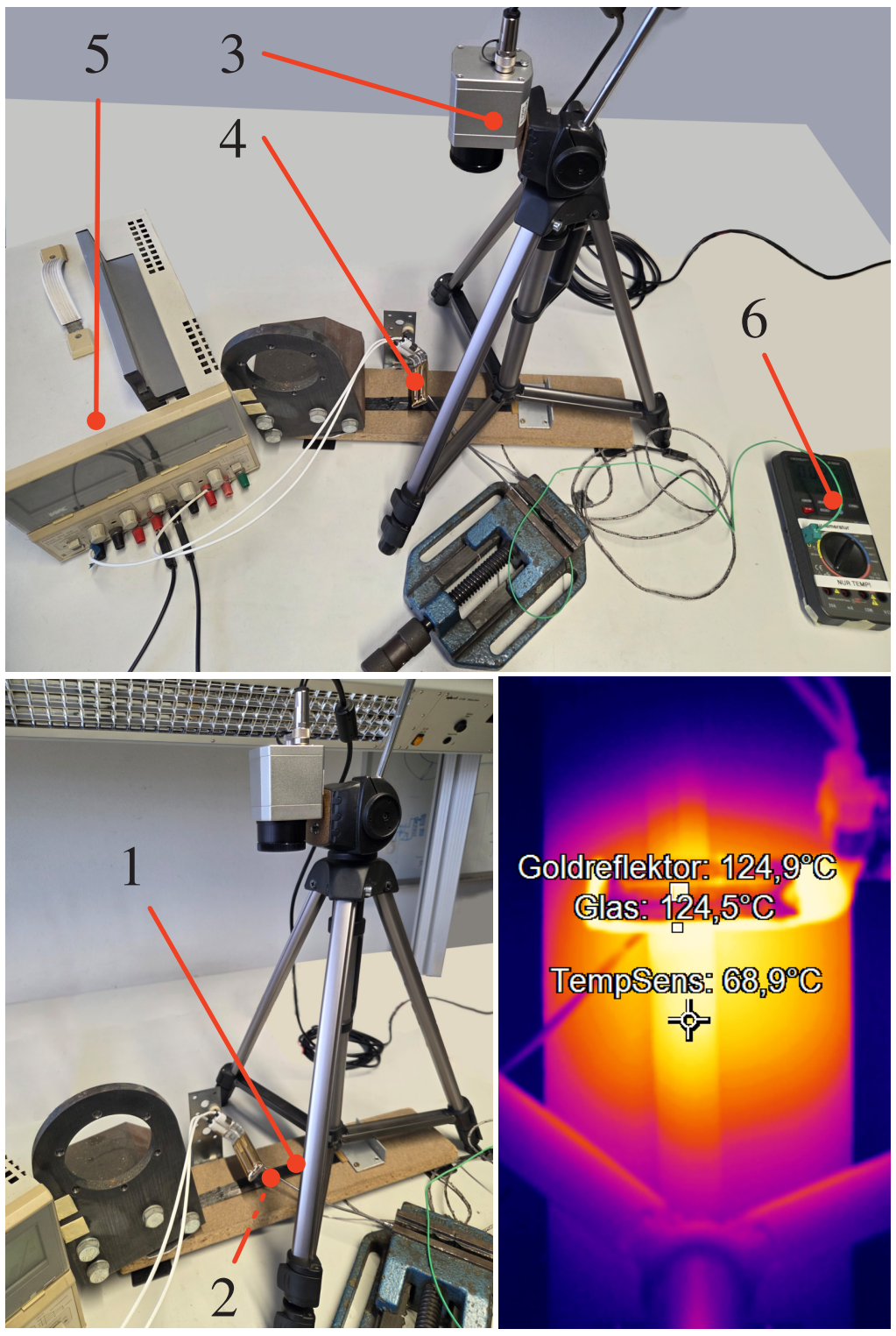


**Figure S5:** Setup for measuring material emissivity $\epsilon_m(T_{m,c})$. (1) Tape strip. (2) Contact thermometer. (3) Thermal infrared camera. (4) Infrared heater. (5) Power supply. (6) Evaluation device contact thermometer.

a contact thermometer as a reference are presented. A thermal infrared camera positioned above the sample provides the non-contact reading. The experimental setup is shown in Fig. S5.

Method A is based on the camera emissivity $\epsilon_{cam}$ adjustment. At each setpoint, the camera's emissivity setting is adjusted until the camera-reported temperature at the measurement spot matches the contact thermometer value. The resulting camera setting is taken as the total hemispherical emissivity $\epsilon_m(T_{m,c})$ for that temperature. This method benefits from the camera's internal corrections (e.g. optics, atmospheric path, reflected background), provided the background temperature and distance are correctly configured.

Method B evaluates $\epsilon_m(T_{m,c})$ by back-calculation at fixed camera settings. Hereby, the camera emissivity is fixed at $\epsilon_{cam}$ and the apparent temperature $T_{app}$ is recorded. Assuming diffuse-gray behavior and accounting for reflected ambient radiation at temperature $T_{amb}$, the measured radiance to the physical radiance of the surface is equated

$$\epsilon_m \sigma T^4 + (1-\epsilon_m)\sigma T_{amb}^4 = \epsilon_{cam}\sigma T_{app}^4 + (1-\epsilon_{cam})\sigma T_{amb}^4, \tag{S36}$$

which yields

$$\epsilon_m(T) = \epsilon_{cam} \frac{T_{app}^4 - T_{amb}^4}{T^4 - T_{amb}^4}. \tag{S37}$$

Here, $T$ is the contact-measured surface temperature, and $\sigma$ is the Stefan-Boltzmann constant.

Both methods produced consistent $\epsilon_m(T_{m,c})$ values across the investigated temperature range, which can be seen in Fig. S6. However, finally Method A was adopted for subsequent analyses because the camera's internal radiometric corrections improve robustness to reflected and transmitted radiation yielding measured values that show qualitative agreement with those reported for carbon fiber reinforced polycarbonate (CF/PC) in [S20], which serves as a close proxy for CF-HDPE. Emissivity values below the specified temperature range cannot be determined without significant error. Consequently, these values are assumed to remain constant and equal to the value measured at the lowest recorded temperature. For the experimental setup, the surroundings were modeled as a black body with $\epsilon_s = 1$.

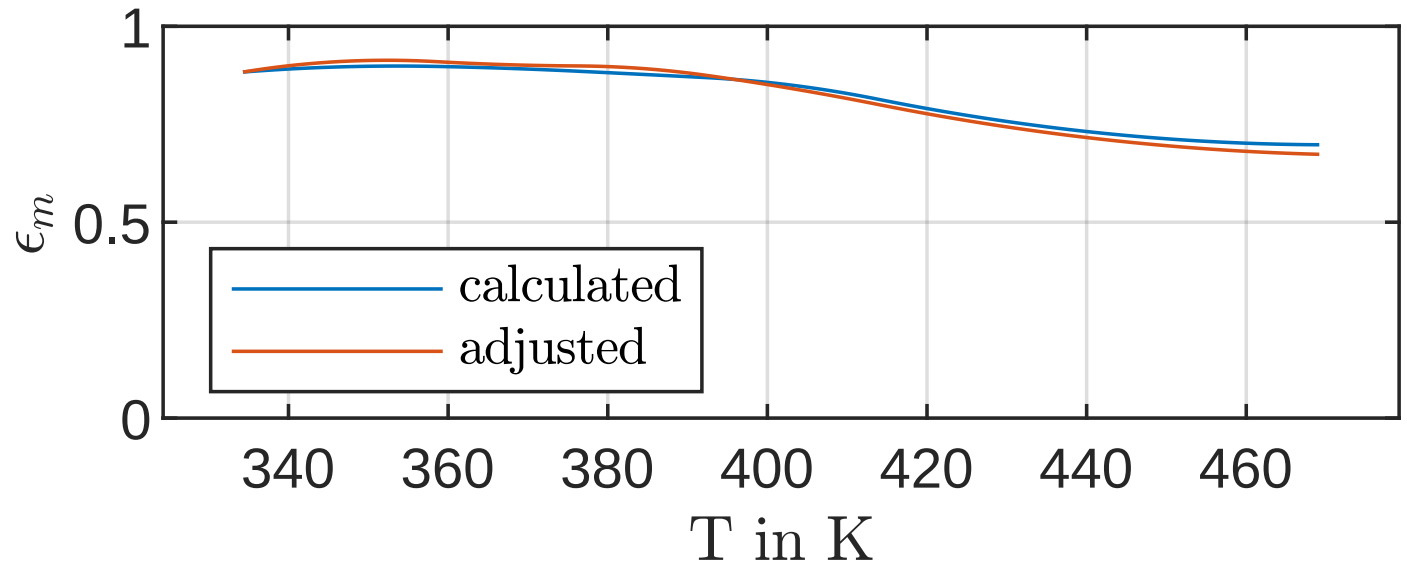


**Figure S6:** Determined tape material emissivity $\epsilon_m(T_{m,c})$.

## S5. Analytical formulations and references for view factor calculation

To account for radiative heat transfer within the numerical model, the precise determination of view factors is required. Due to the geometric complexity of the setup, the underlying mathematical derivations result in extensive analytical expressions. To maintain the conciseness and readability of the main manuscript, these comprehensive formulations, along with their respective geometric definitions and original literature references, are compiled in this section. The equations presented herein form the mathematical basis for the radiative boundary conditions utilized in the simulation results in the main text.

Assuming a fully parallel alignment of the first heat source and the material strip yields the view factor formulation

$$F_{r-t} = \frac{1}{(x_2 - x_1)(y_2 - y_1)} \sum_{i=1}^{2} \sum_{j=1}^{2} \sum_{k=1}^{2} \sum_{l=1}^{2} (-1)^{i+j+k+l} G_{\|}(x_i, y_j, \eta_k, \xi_l) \tag{S38}$$

with

$$G_{\|}(x, y, \eta, \xi) = \frac{1}{2\pi} \Bigg[ (y-\eta)\sqrt{(x-\xi)^2 + z^2} \arctan\left( \frac{y-\eta}{\sqrt{(x-\xi)^2+z^2}} \right) + (x-\xi)\sqrt{(y-\eta)^2 + z^2} \arctan\left( \frac{x-\xi}{\sqrt{(y-\eta)^2+z^2}} \right) - \frac{z^2}{2} \ln\left( (x-\xi)^2 + (y-\eta)^2 + z^2 \right) \Bigg] \tag{S39}$$

defined in [S21, S22, S23, S24, S25, S26] for rectangle to rectangle in a parallel plane. The second case, where the emitter is perpendicularly aligned with the tape, the equations proposed in [S21, S22, S24, S25, S27] for the use on rectangle to rectangle in a perpendicular plane can be applied. Taking advantage of Eq. S38, the corresponding view factor can be computed by exchanging $G_{\|}$ with

$$G_{\perp}(x, y, \eta, \xi) = \frac{1}{2\pi} \left[ (y-\eta)\sqrt{(x^2+\xi^2)} \arctan(K) - \frac{1}{4} \left( (x^2+\xi^2) \ln(1+K^2) - (y-\eta)^2 \ln(1+\frac{1}{K^2}) \right) \right], \tag{S40}$$

where $K = (y-\eta)/\sqrt{(x^2+\xi^2)}$. The geometric setup for calculating the first case with parallel rectangles is shown in Fig. S7, whereas the case with perpendicular rectangles is depicted in Fig. S8.

## S6. System parameter tuning and calibration

To bridge the gap between numerical simulation and physical reality, the integrated process model must be precisely aligned with the experimental setup. While the fundamental governing equations are derived from first-principles thermodynamics, certain system properties exhibit inherent variability or are inaccessible to direct measurement. This section establishes the framework for parameter consolidation and details the subsequent optimization steps required to achieve the predictive capability demonstrated in the primary research article.

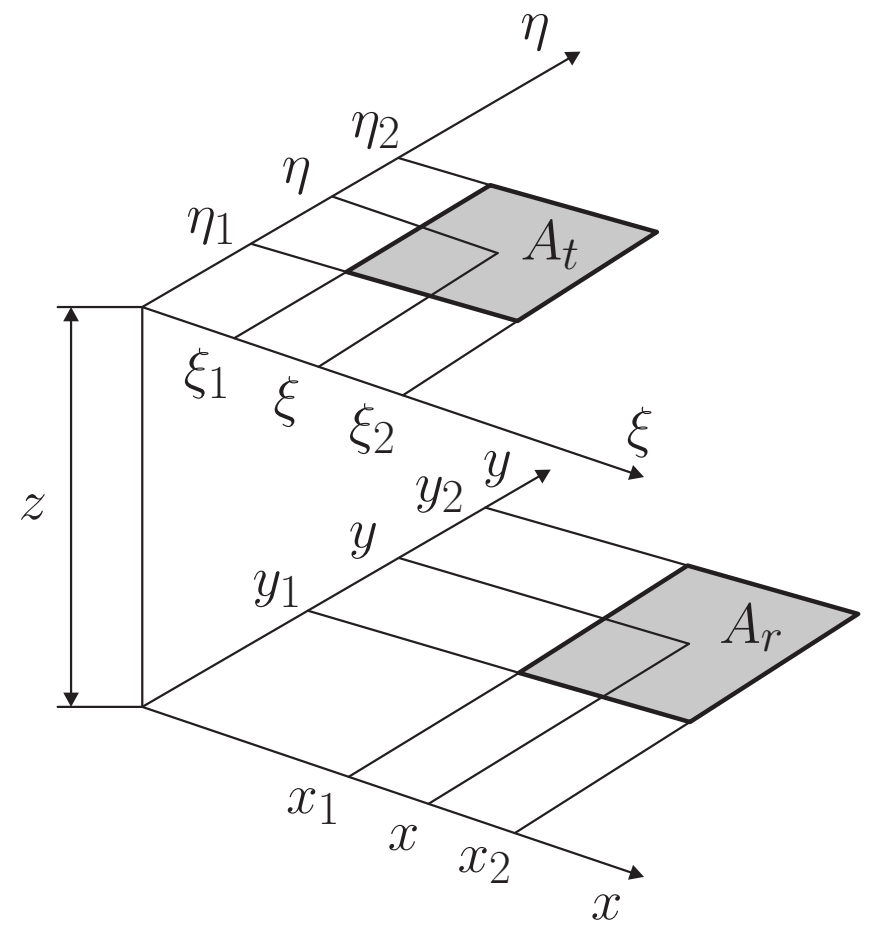


**Figure S7:** Geometry basement for view factor calculation of rectangle to rectangle in a parallel plane. $r$ represents a segment of the rectangular simplification of the heater double-tube geometry and $t$ a segment of the tape.

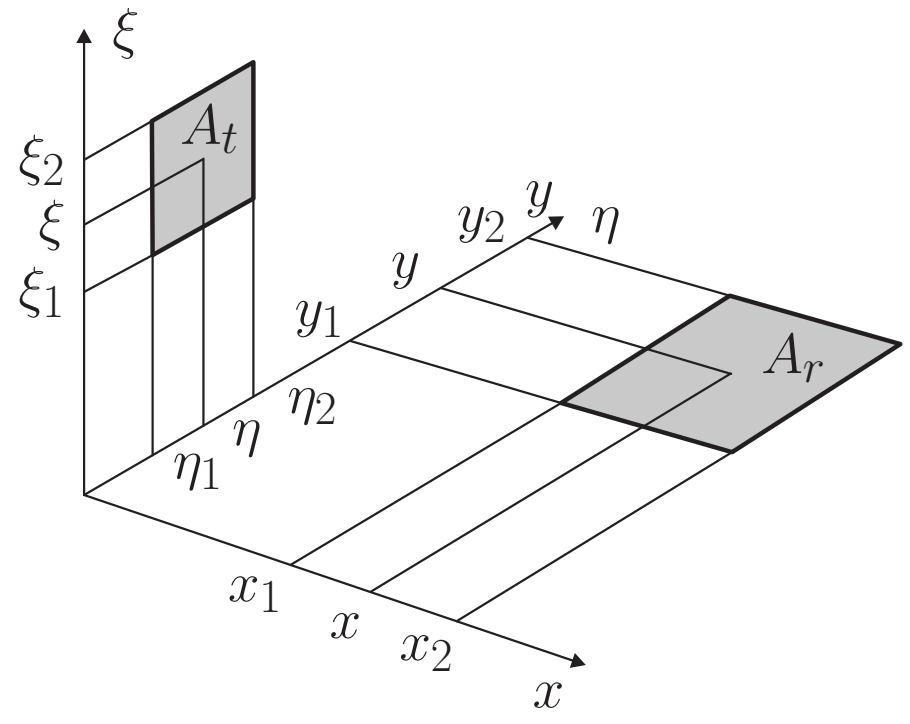


**Figure S8:** Geometry basement for view factor calculation of rectangle to rectangle in a perpendicular plane. $r$ represents a segment of the rectangular simplification of the heater double-tube geometry and $t$ a segment of the tape.

## S6.1. System parameters

A comprehensive overview of the geometrical, thermal and spectral parameters utilized across the multimode heat transfer framework is compiled in the following tables. Specifically, Table S2 includes material-specific parameters of the tape, whereas Table S3 holds the parameters assigned to the infrared heater systems used. These datasets serve as the baseline for the numerical solver, comprising both deterministic physical constants and fixed equipment specifications. Importantly, this inventory also isolates the subsets of uncertain parameters (bold printed nominal values in Table S3), such as geometric details inside the quartz glass envelope like wire or coil diameters lack sufficient direct measurements and are therefore subjected to iterative adjustment in the subsequent subsection. The origin of the quantities is specified in the 'Source' column of the respective parameter tables.

Due to the lack of direct manufacturer specifications for the specific heat capacity of the composite tape $c_{p,m}$, this parameter was analytically determined using the linear rule of mixtures based on the respective mass fractions of the constituent phases

$$c_{p,m} = c_{p,cf} \cdot \phi_{mass,cf} + c_{p,HDPE} \cdot \phi_{mass,HDPE}, \tag{S41}$$

where the mass fractions are derived from their corresponding volume fractions as follows

$$\phi_{mass,cf} = \frac{\rho_{cf}\phi_{cf}}{\rho_m}, \tag{S42}$$

$$\phi_{mass,HDPE} = \frac{\rho_{HDPE}(1-\phi_{cf})}{\rho_m}. \tag{S43}$$

**Table S2**
Material-specific system parameters.

| Parameter | Value | Unit | Source |
|---|---|---|---|
| $c_{p,cf}$ | 750 | $\mathrm{J\,kg^{-1}\,K^{-1}}$ | literature |
| $c_{p,HDPE}$ | 2000 | $\mathrm{J\,kg^{-1}\,K^{-1}}$ | literature |
| $k_{cf,\parallel}$ | 17 | $\mathrm{W\,m^{-1}\,K^{-1}}$ | literature |
| $k_{cf,\perp}$ | 1.5 | $\mathrm{W\,m^{-1}\,K^{-1}}$ | literature |
| $k_{HDPE}$ | 0.44 | $\mathrm{W\,m^{-1}\,K^{-1}}$ | literature |
| $L_m$ | $70 \times 10^{-3}$ | m | geometry |
| $t_m$ | $0.32 \times 10^{-3}$ | m | data sheet |
| $w_m$ | $20 \times 10^{-3}$ | m | data sheet |
| $\phi_{cf}$ | 0.33 | - | data sheet |
| $\rho_{cf}$ | 1750 | $\mathrm{kg/m^3}$ | literature |
| $\rho_{HDPE}$ | 950 | $\mathrm{kg/m^3}$ | literature |
| $\rho_m$ | 1200 | $\mathrm{kg/m^3}$ | data sheet |

The unwound length of the filament $l_f$ results from

$$l_f = \frac{l_w}{p_{coil}} \sqrt{\left(\pi(d_{coil} - d_f)\right)^2 + p_{coil}^2} \qquad \text{(S44)}$$

using the filament coil pitch

$$p_{coil} = d_f + w_g. \qquad \text{(S45)}$$

**Table S3**
Heater-specific system parameters. Nominal values indicate untuned parameters, whereas actual values represent the values used for calculations. Only bold values are assumed to be uncertain and were subsequently tuned.

| Parameter | Actual value | Nominal value | Unit | Source |
|---|---|---|---|---|
| $d_{coil}$ | $\mathbf{1.1 \times 10^{-3}}$ | $\mathbf{1.3 \times 10^{-3}}$ | m | estimated → tuned |
| $d_f$ | $\mathbf{0.21 \times 10^{-3}}$ | $\mathbf{0.2 \times 10^{-3}}$ | m | estimated → tuned |
| $d_{mount}$ | $\mathbf{0.8 \times 10^{-3}}$ | $\mathbf{1 \times 10^{-3}}$ | m | estimated → tuned |
| $d_{q,e}$ | $10 \times 10^{-3}$ | $10 \times 10^{-3}$ | m | measured |
| $k_f$ | 160 | 160 | $\mathrm{W\,m^{-1}\,K^{-1}}$ | literature |
| $k_q$ | 1.39 | 1.39 | $\mathrm{W\,m^{-1}\,K^{-1}}$ | literature |
| $l_{mount}$ | $\mathbf{8 \times 10^{-3}}$ | $\mathbf{10 \times 10^{-3}}$ | m | estimated → tuned |
| $l_q$ | $51 \times 10^{-3}$ | $51 \times 10^{-3}$ | m | measured |
| $l_w$ | $50 \times 10^{-3}$ | $50 \times 10^{-3}$ | m | measured |
| $t_q$ | $1.2 \times 10^{-3}$ | $1.2 \times 10^{-3}$ | m | assumed |
| $w_g$ | $\mathbf{0.16 \times 10^{-3}}$ | $\mathbf{0.2 \times 10^{-3}}$ | m | estimated → tuned |
| $\lambda_1$ | $\mathbf{0.5 \times 10^{-6}}$ | $\mathbf{0.8 \times 10^{-6}}$ | m | literature → tuned |
| $\lambda_2$ | $\mathbf{6 \times 10^{-6}}$ | $\mathbf{8 \times 10^{-6}}$ | m | literature → tuned |
| $\rho_f$ | 19 250 | 19 250 | $\mathrm{kg/m^3}$ | literature |
| $\rho_q$ | 2201 | 2201 | $\mathrm{kg/m^3}$ | literature |

## S6.2. Tuning and calibration

This subsection details the systematic tuning process applied to the previously isolated uncertain system parameters. To rigorously quantify the necessity and efficiency of this calibration, the baseline response of the uncalibrated model, using purely estimated or comparable values (nominal values Table S3), is evaluated as a reference case. The temperature profiles of both the nip point and the monitoring point for the untuned model are depicted in Fig. S9,

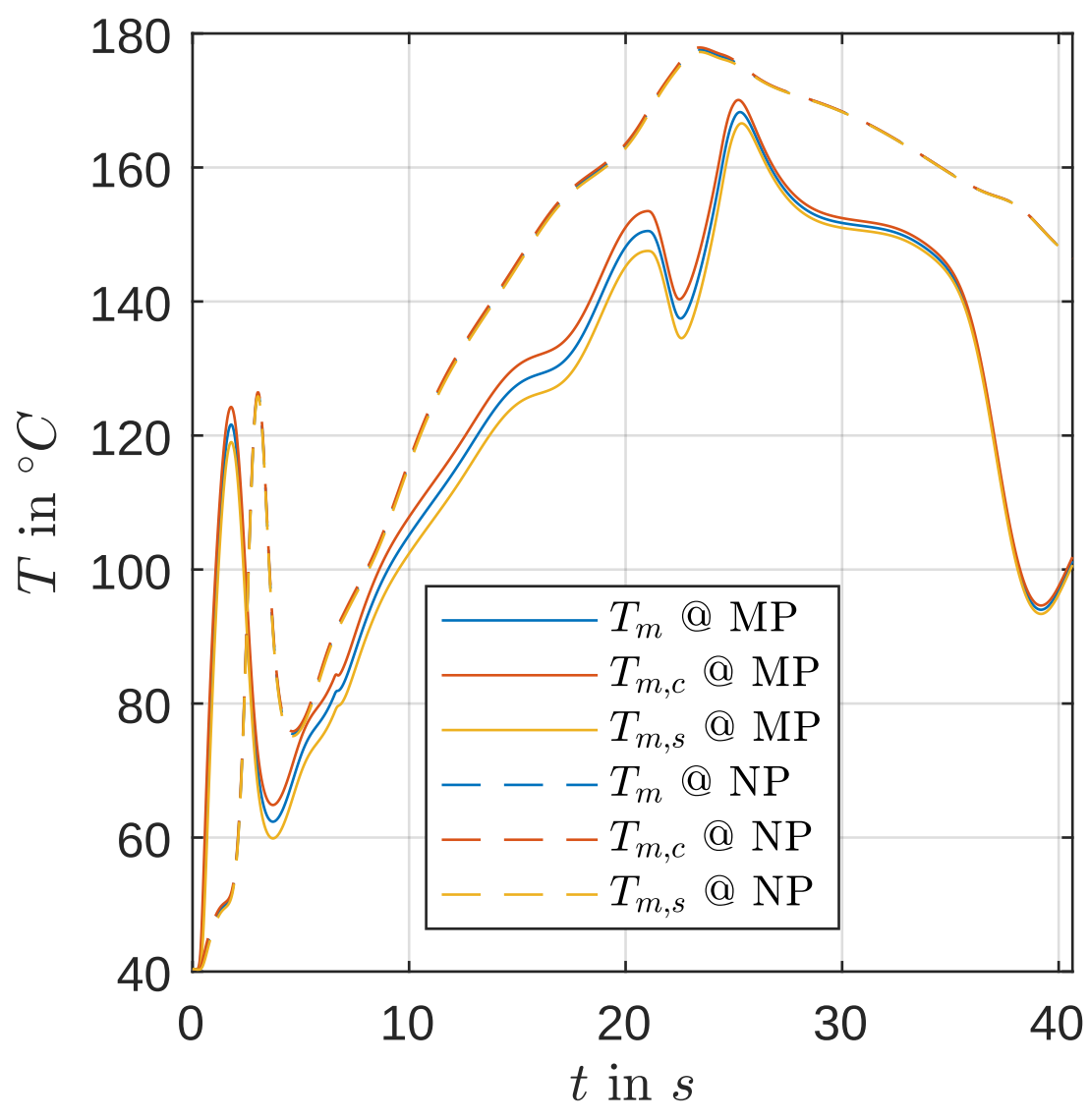


**Figure S9:** Temperature simulation at measuring point (MP) and nip point (NP) on both sides of tape during test trajectory with untuned system parameters.

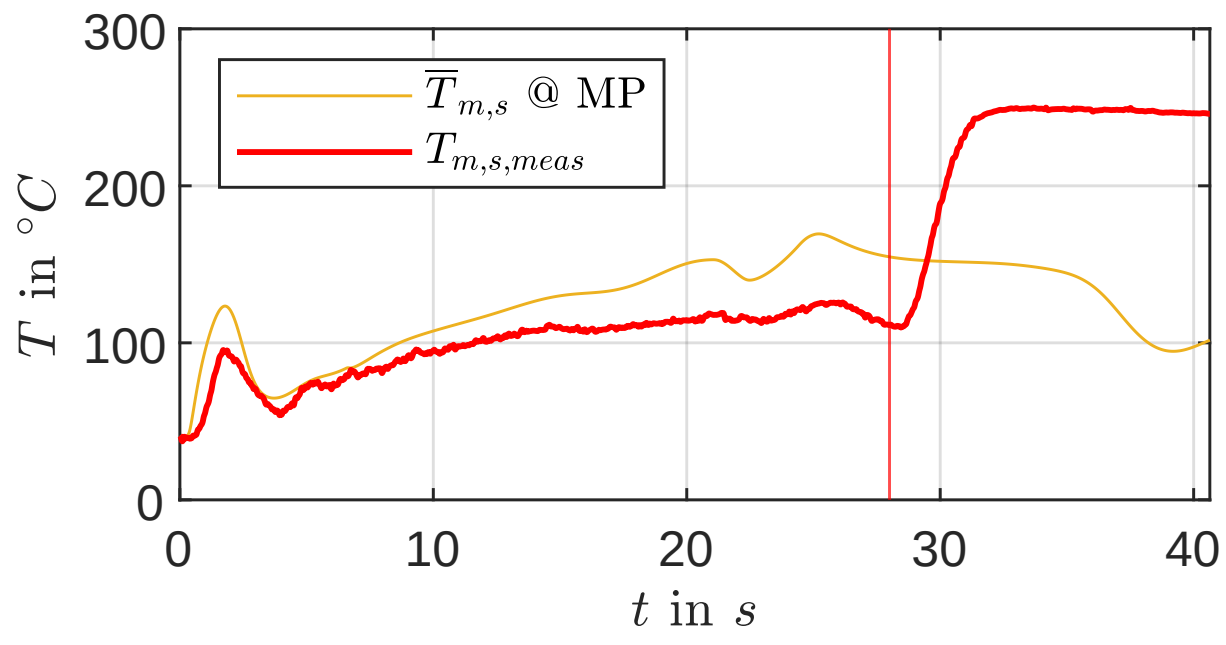


**Figure S10:** Simulated $\overline{T}_{m,s}$ and measured $T_{m,s,meas}$ temperature at measuring point on tape side facing the surroundings during test trajectory with untuned system parameters.

while Fig. S10 shows a comparison between the measured profile $T_{m,s,meas}$ and the uncalibrated simulation value $\overline{T}_{m,s}$ in the measuring point. Using the tuned parameter set (actual values in Table S3) yields a significant accuracy improvement of the model results as shown in the main manuscript. The tuning procedure systematically adjusts the uncertain parameters defined in Table S3 by iteratively modifying them within a plausible physical range until the results closely align with the experimental data. With this approach, the uncertain parameters in the third column of Table S3 are adapted until they reach the final values stated in the second column. Thereby, the qualitative temperature profile of the physical model is not changed. Instead, only the unknown values are tuned to bridge the gap between the numerical simulation and the physical reality.

# Nomenclature

*Roman symbols*

| | |
|---|---|
| $A_f$ | Surface area of filament coil [m$^2$] |
| $A_k^f$ | Closed boundary surface of filament control volume $V_k^f$ [m$^2$] |
| $A_k^q$ | Closed boundary surface of quartz control volume $V_k^q$ [m$^2$] |
| $A_q$ | Surface area of quartz glass envelope [m$^2$] |
| $A_r$ | Surface area of rectangular emitter simplification segment [m$^2$] |
| $A_t$ | Surface area of tape segment [m$^2$] |
| $A_\theta$ | Area of surface $\theta$ [m$^2$] |
| $A_\psi$ | Area of surface $\psi$ [m$^2$] |
| $a_f$ | Cross-sectional area of filament wire [m$^2$] |
| $a_g$ | Effective segment exchange area between adjacent windings [m$^2$] |
| $a_{mount}$ | Cross-sectional area of filament mounting wire [m$^2$] |
| $a_q$ | Circular area of the hollow quartz glass cylinder [m$^2$] |
| $Bi$ | Biot number [−] |
| $C''_m$ | Areal heat capacity of one tape layer in the two-node model [J m$^{-2}$ K$^{-1}$] |
| $c_0$ | Speed of light in vacuum [m s$^{-1}$] |
| $c_p$ | Specific heat capacity [J kg$^{-1}$ K$^{-1}$] |
| $c_{p,f}$ | Specific heat capacity of tungsten (filament) [J kg$^{-1}$ K$^{-1}$] |
| $c_{p,m}$ | Specific heat capacity of composite material (tape) [J kg$^{-1}$ K$^{-1}$] |
| $c_{p,q}$ | Specific heat capacity of quartz glass [J kg$^{-1}$ K$^{-1}$] |
| $d_{coil}$ | Diameter of filament coil [m] |
| $d_f$ | Diameter of filament wire [m] |
| $d_{mount}$ | Diameter of filament mounting wire [m] |
| $d_{q,e}$ | External diameter of quartz glass envelope [m] |
| $d_{q,i}$ | Internal diameter of quartz glass envelope [m] |
| $d_r$ | Internal diameter of gold reflector layer [m] |
| $E_b$ | Total hemispherical emissive power of a black body [W m$^{-2}$] |
| $E_{b,\lambda}$ | Spectral emission of a black body for a certain wave length [W m$^{-3}$] |
| $E_{b,\lambda_{1-2}}$ | Black body band emissive power in a certain spectral range [W m$^{-2}$] |
| $E_\lambda$ | Spectral emissive power of a gray body [W m$^{-3}$] |
| $E_{\lambda_{1-2}}$ | Gray body band emissive power in a certain spectral range [W m$^{-2}$] |
| $\mathbf{F}$ | View factor matrix [−] |
| $F_{\theta-\psi}$ | View factor between surface $\theta$ and surface $\psi$ [−] |

| | |
|---|---|
| $G$ | Irradiation [$\mathrm{W\,m^{-2}}$] |
| $G_{cond,gap}$ | Thermal conductance of neon in gap between adjacent windings [$\mathrm{W\,K^{-1}}$] |
| $G_f$ | Thermal conductance of filament [$\mathrm{W\,K^{-1}}$] |
| $G_{mount}$ | Thermal conductance of filament mounting wire [$\mathrm{W\,K^{-1}}$] |
| $G_q$ | Thermal conductance of quartz glass envelope [$\mathrm{W\,K^{-1}}$] |
| $Gr$ | Grashof number [−] |
| $h$ | Planck constant [J s] |
| $h_m$ | General heat transfer coefficient between tape and fluid [$\mathrm{W\,m^{-2}\,K^{-1}}$] |
| $h_{m,c}$ | Heat transfer coefficient between tape and fluid in radiation cavity [$\mathrm{W\,m^{-2}\,K^{-1}}$] |
| $h_{m,s}$ | Heat transfer coefficient between tape and surrounding fluid [$\mathrm{W\,m^{-2}\,K^{-1}}$] |
| $\overline{h}_q$ | Average heat transfer coefficient between quartz glass envelope and fluid in radiation cavity [$\mathrm{W\,m^{-2}\,K^{-1}}$] |
| $i$ | Tape segment index [−] |
| $i_e$ | Supply current through infrared heaters [A] |
| $J$ | Radiosity [$\mathrm{W\,m^{-2}}$] |
| $k$ | Heater segment index [−] |
| $k_b$ | Boltzmann constant [$\mathrm{J\,K^{-1}}$] |
| $k_{cf,\parallel}$ | Conductivity of carbon fibers parallel to fiber alignment [$\mathrm{W\,m^{-1}\,K^{-1}}$] |
| $k_{cf,\perp}$ | Conductivity of carbon fibers perpendicular to fiber alignment [$\mathrm{W\,m^{-1}\,K^{-1}}$] |
| $k_f$ | Conductivity of tungsten (filament) [$\mathrm{W\,m^{-1}\,K^{-1}}$] |
| $k_{HDPE}$ | Conductivity of HDPE matrix [$\mathrm{W\,m^{-1}\,K^{-1}}$] |
| $k_m$ | General conductivity of composite material [$\mathrm{W\,m^{-1}\,K^{-1}}$] |
| $k_{m,\parallel}$ | Conductivity of composite material parallel to fiber alignment [$\mathrm{W\,m^{-1}\,K^{-1}}$] |
| $k_{m,\perp}$ | Conductivity of composite material perpendicular to fiber alignment [$\mathrm{W\,m^{-1}\,K^{-1}}$] |
| $k_n$ | Conductivity of neon gas [$\mathrm{W\,m^{-1}\,K^{-1}}$] |
| $k_q$ | Conductivity of quartz glass [$\mathrm{W\,m^{-1}\,K^{-1}}$] |
| $L_m$ | Total tape length within considered Eulerian domain [m] |
| $l_f$ | Stretched length of filament wire [m] |
| $l_{mount}$ | Length of filament mounting wire [m] |
| $l_q$ | Length of quartz glass envelope [m] |
| $l_w$ | Length of winded filament coil [m] |
| $m_f$ | Filament mass [kg] |
| $m_q$ | Quartz glass mass [kg] |
| $N_h$ | Number of heater segments [−] |
| $N_t$ | Number of time steps [−] |

| | |
|---|---|
| $N_x$ | Number of tape segments [−] |
| $Nu_x$ | Local Nusselt number [−] |
| $P_{el}$ | Electric power [W] |
| $Pr$ | Prandtl number [−] |
| $p$ | Standard pressure [Pa] |
| $p'_{el}$ | Electric line-source power density [W m$^{-1}$] |
| $Q_{c,n}$ | Neon conduction heat [W] |
| $Q_{cond,f}$ | Conduction heat along filament wire [W] |
| $Q_{cond,gap}$ | Conduction heat of neon in the gap between adjacent filament windings [W] |
| $Q_{cond,mount}$ | Conduction heat along filament mounting wire [W] |
| $Q_{cond,q}$ | Conduction heat along quartz glass envelope [W] |
| $Q_{gap}$ | Conduction and radiation heat in the gap between adjacent filament windings [W] |
| $Q_{r,f}$ | Radiation heat of filament [W] |
| $Q_{r,q}$ | Radiation heat of quartz glass envelope [W] |
| $Q_{rad,gap}$ | Radiation heat in the gap between adjacent filament windings [W] |
| $q'''_{b,vol}$ | Effective volumetric source within the tape due to boundary convection and radiation [W m$^{-3}$] |
| $q'_{c,n}$ | Line-source conduction heat density to/from neon gas [W m$^{-1}$] |
| $q''_{cond,m}$ | Conduction heat flux across tape thickness [W m$^{-2}$] |
| $q''_{conv,c}$ | Convection heat flux acting on tape surface facing radiation cavity [W m$^{-2}$] |
| $q''_{conv,m}$ | General convection heat flux acting on tape surface [W m$^{-2}$] |
| $q''_{conv,s}$ | Convection heat flux acting on tape surface facing surroundings [W m$^{-2}$] |
| $q'_{gap}$ | Line-source heat density between adjacent filament windings [W m$^{-1}$] |
| $q'_{r,f}$ | Line-source radiation heat density of filament [W m$^{-1}$] |
| $q''_{rad}$ | Net radiation heat flux [W m$^{-2}$] |
| $q''_{rad,f}$ | Net radiation heat flux acting on filament coil surface [W m$^{-2}$] |
| $q''_{rad,m}$ | Net radiation heat flux acting on tape surface [W m$^{-2}$] |
| $q''_{rad,q}$ | Net radiation heat flux acting on quartz glass surface [W m$^{-2}$] |
| $q'''_{vol,f}$ | Volumetric heat generation rate within tungsten filament [W m$^{-3}$] |
| $q'''_{vol,m}$ | Volumetric heat generation rate within composite material [W m$^{-3}$] |
| $R$ | Electrical resistance of filament [Ω] |
| $R_s$ | Specific gas constant [J kg$^{-1}$ K$^{-1}$] |
| $Ra$ | Rayleigh number [−] |
| $Re$ | Reynolds number [−] |
| $Ri$ | Richardson number [−] |

| | |
|---|---|
| $r$ | Reflectivity [−] |
| $\mathbf{r}$ | Reflectivity matrix [−] |
| $r_t$ | Resistivity of tungsten [Ω m] |
| $T_0$ | Room temperature [K] |
| $T_{amb}$ | Ambient temperature [K] |
| $T_{app}$ | Apparent temperature detected by thermal camera [K] |
| $T_f$ | Filament temperature [K] |
| $T_{film}$ | Film temperature [K] |
| $T_m$ | Arithmetic mean temperature between surface temperatures of the two-node tape model [K] |
| $T_{m,c}$ | Tape surface temperature facing the radiation cavity [K] |
| $T_{m,s}$ | Tape surface temperature facing the surroundings [K] |
| $\overline{T}_{m,s}$ | Arithmetic mean of simulated tape segment surface temperatures captured by the sensor [K] |
| $T_{m,s,meas}$ | Tape surface temperature measurement facing the surroundings [K] |
| $T_n$ | Neon mean temperature [K] |
| $T_q$ | Quartz glass temperature [K] |
| $\overline{T}_q$ | Arithmetic mean of simulated quartz glass segment surface temperatures captured by the sensor [K] |
| $T_{q,corr}$ | Emissivity-corrected surface temperature measurement of quartz glass [K] |
| $T_{q,meas}$ | Quartz glass surface temperature measurement [K] |
| $T_r$ | Gold reflector temperature [K] |
| $T_s$ | General segment surface temperature [K] |
| $T_\infty$ | General ambient temperature [K] |
| $T_{\infty,c}$ | Ambient fluid temperature inside radiation cavity [K] |
| $T_{\infty,s}$ | Ambient fluid temperature of surroundings [K] |
| $t_h$ | Height of rectangular infrared heater simplification [m] |
| $t_m$ | Tape thickness [m] |
| $t_q$ | Thickness of quartz glass envelope [m] |
| $U$ | Characteristic velocity [m s$^{-1}$] |
| $\mathbf{u}$ | System input vector [−] |
| $V_k^f$ | Spatial filament control volume of segment $k$ [m$^3$] |
| $V_k^q$ | Spatial quartz control volume of segment $k$ [m$^3$] |
| $v_T$ | Tape velocity along feed direction $x$ [m s$^{-1}$] |
| $w_g$ | Gap between adjacent filament windings [m] |
| $w_h$ | Width of rectangular infrared heater simplification [m] |
| $w_m$ | Tape width [m] |
| $\mathbf{x}$ | Thermal state vector [K] |

| | |
|---|---|
| $x_{h,1}, x_{h,2}$ | Center positions of infrared heaters 1 and 2 in tape domain along $x$-axis [m] |
| $y_{h,1}, y_{h,2}$ | Center positions of infrared heaters 1 and 2 in tape domain along $y$-axis [m] |

*Greek symbols*

| | |
|---|---|
| $\alpha$ | Thermal diffusivity [$m^2\,s^{-1}$] |
| $\beta$ | Coefficient of thermal expansion [$K^{-1}$] |
| $\Delta x$ | Tape segment length [m] |
| $\Delta T_m$ | Temperature difference between tape surfaces [K] |
| $\Delta z$ | Heater segment length [m] |
| $\epsilon$ | Emissivity [−] |
| $\boldsymbol{\epsilon}$ | Emissivity matrix [−] |
| $\epsilon_{cam}$ | Thermal camera emissivity [−] |
| $\epsilon_{eff}$ | Effective emissivity in a certain spectral range [−] |
| $\epsilon_{gap}$ | Effective emissivity between adjacent filament coil windings [−] |
| $\eta$ | Blockage factor [−] |
| $\theta$ | Index summarizing all involved surfaces within the radiation cavities [−] |
| $\lambda$ | Wave length [m] |
| $\mu$ | Dynamic viscosity [$kg\,m^{-1}\,s^{-1}$] |
| $\nu$ | Kinematic viscosity [$m^2\,s^{-1}$] |
| $\phi_{cf}$ | Fiber volume content in composite tape [−] |
| $\rho_f$ | Density of tungsten (filament) [$kg\,m^{-3}$] |
| $\rho_m$ | Density of composite material (tape) [$kg\,m^{-3}$] |
| $\rho_q$ | Density of quartz glass [$kg\,m^{-3}$] |
| $\sigma$ | Stefan-Boltzmann constant [$W\,m^{-2}\,K^{-4}$] |
| $\tau$ | Transmissivity [−] |
| $\boldsymbol{\tau}$ | Transmissivity matrix [−] |
| $\tau_{eff}$ | Effective transmissivity in a certain spectral range [−] |

*Subscripts / Superscripts*

| | |
|---|---|
| $abs$ | Absolute value |
| $air$ | Fluid surrounding the tape |
| $m, c$ | Tape side facing the radiation cavity |
| $cf$ | Carbon fiber |
| $f$ | Filament |
| $m$ | Composite material (tape) |
| $max$ | Maximum value |

| | |
|---|---|
| $q$ | Quartz |
| $rel$ | Relative value |
| $m, s$ | Tape side facing the surroundings |

*Abbreviations*

| | |
|---|---|
| **ATL** | Automated tape laying |
| **ACF** | Active contact flange |
| **CF** | Carbon fiber |
| **DAE** | Differential-algebraic equation |
| **FEM** | Finite element method |
| **HDPE** | High-density polyethylene |
| **IR** | Infrared |
| **MP** | Measuring point |
| **MPC** | Model predictive control |
| **NP** | Nip point |
| **NRMSE** | Normalized root mean squared error |
| **ODE** | Ordinary differential equation |
| **PC** | Polycarbonate |
| **PDE** | Partial differential equation |
| **PEEK** | Polyether ether ketone |
| **RMSE** | Root mean squared error |
| **SWIR** | Short-wave infrared |
| **TCP** | Tool center point |
| **TDLD** | Temperature-dependent Lorentz-Drude |
| **UD** | Unidirectional |



\end{document}